\documentclass[letterpaper,twocolumn,fleqn]{article}
\pdfoutput=1

\usepackage{ist}
\usepackage{graphicx}
\usepackage{float}
\usepackage{url}
\usepackage{xcolor}
\usepackage{amsmath}
\usepackage{colortbl}
\usepackage{booktabs}
\usepackage[normalem]{ulem}
\usepackage{makecell}
\usepackage{multirow} 
\usepackage{amsfonts}
\usepackage{amssymb}
\usepackage{hyperref}

\usepackage{dblfloatfix}

\title{Albedo Estimation via Latent Bridge Matching}
\author{Carme Corbi, David Serrano-Lozano, Javier Vazquez-Corral, Maria Vanrell \\ Universitat Autònoma de Barcelona and Computer Vision Center}

\date{} 

\begin{document}
\maketitle 
\thispagestyle{empty} 

\begingroup
\renewcommand\thefootnote{}
\footnotetext{Paper accepted at the Color and Imaging Conference (CIC 2026), hosted by the Society for Imaging Science and Technology (IS\&T).}
\endgroup

\begin{abstract}
\noindent Recent advances in Intrinsic Image Decomposition (IID) have increasingly relied on generative models. However, progress remains limited by three key challenges: (a) insufficient physical consistency, (b) high computational cost at inference time, and (c) limited generalization capabilities. In this work, we show that latent bridge matching (LBM) effectively addresses these limitations for albedo estimation. We introduce a novel LBM-based architecture that enforces physical consistency through a pixel reconstruction loss, benefits from the inherent efficiency of LBM low-cost inference, and improves generalization across diverse datasets by incorporating a shading conditioning. In this extended version, we additionally show that conditioning the shading estimator itself on the predicted albedo further improves reconstruction fidelity, and we benchmark our best model against state-of-the-art IID methods across five real and synthetic datasets.
\href{https://github.com/CVC-Color/albedoLBM}{https://github.com/CVC-Color/albedoLBM}
\end{abstract}

\section{Introduction}
\label{sec:intro}
\noindent Understanding the interaction between light and surfaces of the scene from a single image is a fundamental problem in computer vision. Although humans can unconsciously discount illumination changes and infer material properties, replicating this perceptual ability in computational systems remains a highly challenging task.

\noindent Intrinsic Image Decomposition (IID) aims to separate an image into its underlying physical components. Given a single RGB image $I$, IID seeks to recover two factorized images: reflectance or albedo, $A$, which represents the inherent color properties of surfaces; and shading $S$, which captures the effects of the interaction between light and scene geometry. The most basic formulation, as originally proposed by Barrow and Tenenbaum~\cite{Barrow1978Intrinsic} is given by
\begin{equation}
I(x, y) = A(x, y) \cdot S(x, y)
\label{eq:image_formation}
\end{equation}
\noindent where $(x, y)$ are pixel coordinates and $\cdot$ denotes pixel-wise product. It is based on a simplified Lambertian assumption that is frequently violated; as a consequence, most real-world images cannot be accurately explained using only these two components. Several works \cite{serra2014,shi2017learning} have proposed extending the basic model explicitly to account for non-Lambertian effects:
\begin{equation}
I(x, y) = A(x, y) \cdot S(x, y) + R(x,y)
\label{eq:extimage_formation}
\end{equation}
\noindent where $R(x,y)$ captures residual effects including others global illumination phenomena and non-diffuse material components. In summary, IID is a severely ill-posed problem, as infinite pairs $(A,S)$ or triplets $(A,S,R)$ can explain the same observed image. 

The estimation of intrinsic components has been addressed using a variety of techniques over the years. Recently, the primary focus has shifted toward generative frameworks. Despite their success, these approaches exhibit several notable drawbacks. First, they often lack sufficient physical consistency, likely due to the difficulty of effectively incorporating pixel-level physical constraints during training. Second, they tend to produce residual noise as a consequence of the generative process. Third, they incur a relatively high computational cost at inference time. Finally, all IID methods frequently demonstrate limited generalization capabilities, which may stem from their dependency on large-scale synthetic datasets required for training.

In this work, we hypothesize that LBM, a recent generative framework, can help address some of these limitations and provide a promising new approach for IID. Based on this hypothesis, and after presenting a brief overview of existing generative frameworks, we propose a pipeline that leverages the ability of LBM to be trained with a pixel-wise reconstruction loss while naturally incorporating constraints from the input image and additional cues. We then design a set of experiments to evaluate the proposed approach, leading to a novel method. We summarize the following contributions:
\begin{itemize}
\item Demonstrating that LBM constitutes an effective generative framework for IID, achieving improvements in both computational efficiency (up to $75\%$ reduction in inference time) and accuracy (from 3 to 6 PSNR units depending on dataset) compared to other generative approaches.
\item Exploiting the pixel-level nature of LBM to introduce a reconstruction loss that encourages physical consistency, promoting adherence of the estimated albedo to the underlying image formation model (reconstruction error is reduced to $50\%$) 
\item Investigating different conditioning mechanisms to enrich the generation process with additional contextual information, showing that shading outperforms surface normal-based alternatives.
\item Training a complementary shading estimation model as an efficient conditioning module to guide albedo estimation, resulting in a physically grounded two-stage intrinsic decomposition pipeline.
\end{itemize}

\section{Related works}
\label{sec:related}

\paragraph{Classical approaches} to IID predate the current generative wave and largely fall into three families. The first builds on Retinex theory~\cite{Land1971Retinex}, which assumes that reflectance is piece-wise constant while shading varies smoothly, attributing large image gradients to reflectance changes and small ones to illumination~\cite{Barrow1978Intrinsic}; later variants extended this idea to color by exploiting shading-invariant chromaticity cues~\cite{Funt1992Shading}, amongst others. The second family casts decomposition as an energy-minimization problem, imposing hand-crafted priors on reflectance and shading (e.g., sparsity of reflectance, smoothness of shading, and local color constancy) and solving it through optimization~\cite{Tappen2005Intrinsic,Grosse2009Intrinsic,Bell2014Intrinsic}. A key advantage of these optimization-based methods is that they require no labeled training data, but they rely on assumptions that are frequently violated in real scenes and tend to generalize poorly to complex, non-Lambertian content—limitations that motivated the shift toward learning-based and, more recently, generative formulations. Therefore, the third family is the learning-based that comes in parallel with the advent of large-scale datasets and deep learning techniques that implicitly learn intrinsic regularities from example. They are based on an encoder mapping the input image into a compact yet expressive latent representation, which is then decoded into one or more output images, here we highlight latest ones, such as, PIE-Net~\cite{Das2022PIENet}, Zhu et-al~\cite{zhu2022learning} and Careaga et-al~\cite{careaga2023ordinal}. The IID ill-posed nature makes that classical physical assumptions remain valuable through tailored loss functions or explicitly via regularization terms \cite{Li2018CGIntrinsics,Ma2018NoIntrinsic}. Despite these advances, learning-based approaches remain sensitive to dataset bias, which limit their ability to robustly generalize to real-world scenes.

\vspace{-5mm}
\paragraph{Generative approaches.}~Recently, generative models have taken over most low-level vision tasks, including IID. GAN-based methods cast the decomposition as a layer-separation problem, leveraging adversarial losses together with cycle-consistency and self-supervision constraints to recover albedo and shading without paired ground truth~\cite{lettry2018darn, yang2025relighting}. More recently, diffusion models have become the dominant paradigm: by repurposing the strong priors of large-scale pretrained text-to-image generators,  such as RGB$\leftrightarrow$X~\cite{zeng2024rgbx}, IntrinsicDiffusion~\cite{luo2024intrinsicdiffusion},  Marigold~\cite{ke2025marigold}, PRISM~\cite{dirik2026prism}, and ReasonX~\cite{dirik2026reasonx} which achieve high-quality, generalizable decompositions, often framing IID as a conditional generation task. However, these approaches inherit the high inference cost and residual stochasticity of the generative sampling process.

\vspace{-5mm}
\paragraph{IID datasets.}~A central obstacle in IID is the difficulty of obtaining ground-truth decompositions, which has shaped the available benchmarks. Existing datasets fall broadly into three categories, reflecting different trade-offs between realism, scale, and annotation quality. \emph{Real-world controlled} datasets capture scenes under laboratory conditions with known illuminations; the MIT Intrinsic Images dataset~\cite{grosse2009ground} is the primary example, containing 220 images of 20 objects under 11 directional lights, with ground-truth shading obtained by spray-painting the objects gray. Its controlled setting makes decomposition tractable but limits generalization due to its small, object-centric scope. \emph{Synthetic} datasets instead use physically based rendering to obtain dense, pixel-accurate annotations at scale: InteriorVerse~\cite{zhu2022learning} provides over 50K indoor renderings with reflectance, shading, and surface normal maps, Hypersim~\cite{roberts2021hypersim} offers more than 77K images across 461 cluttered indoor scenes with an explicit residual term for non-Lambertian effects (Eq.~\ref{eq:extimage_formation}), and ARAP~\cite{bonneel2017intrinsic} contributes 149 photorealistic images rendered with LuxRender. All synthetic sources, however, share a sim-to-real gap that can hinder generalization to real images. Finally, \emph{real-world in-the-wild} datasets trade dense supervision for scene diversity: Intrinsic Images in the Wild (IIW)~\cite{Bell2014Intrinsic} addresses this through nearly a million crowdsourced \emph{relative} reflectance judgments over 5,230 indoor images, providing sparse supervision via the WHDR metric rather than dense ground truth. As a result, current methods do not agree on a common protocol: they are trained on different combinations of these datasets and evaluated with disparate, dataset-specific metrics, making fair comparison between approaches difficult -- a point we return to in Section~\ref{sec:limitations}.

\section{Background}
\label{sec:background}
Diffusion models are generative models that learn to reverse a gradual noising process: starting from pure Gaussian noise, they iteratively denoise a sample by predicting, at each step, the noise added to a clean image~\cite{ho2020denoising}. Usually this iterative process is performed within a learned latent space, reducing dimensionality. While powerful, their many-step sampling makes inference slow. Flow Matching (FM)~\cite{lipman2023flow} reframes generation as learning a continuous velocity field $v_\theta$ that transports samples from a simple prior to the data distribution. Given the linear interpolant $x_t = (1-t)\,x_0 + t\,x_1$ between a prior sample $x_0 \sim \mathcal{N}(0, I)$ and a data sample $x_1$, the field is trained by regressing onto the constant target velocity:
\begin{equation}
\mathcal{L}_{\text{FM}} = \mathbb{E}_{t, x_0, x_1} \left[\,\lVert v_\theta(x_t, t) - (x_1 - x_0) \rVert^2\,\right].
\label{eq:fm}\end{equation}

Sampling then integrates $v_\theta$ along straighter paths, so inference is not tied to a fixed noise schedule and requires fewer iterations~\cite{lipman2023flow}.

For completeness, we note that the diffusion formulation above corresponds to a discrete-time Markov chain in which Gaussian noise is progressively added to a clean latent $z_0$ over timesteps $t \in \{0,1,\dots,T\}$, giving $z_t = \sqrt{\bar{\alpha}_t}\,z_0 + \sqrt{1-\bar{\alpha}_t}\,\epsilon$, where $\bar{\alpha}_t$ is the cumulative product of a predefined noise schedule. A network $\epsilon_\theta(z_t,t)$ is trained to predict the injected noise, and generation proceeds by iteratively reversing this chain from $z_T \sim \mathcal{N}(0,I)$ down to $z_0$. Flow Matching replaces this stochastic, many-step chain with the deterministic ODE trajectory of Eq.~\ref{eq:fm}, which is what enables inference in far fewer steps.

However, a key limitation of both formulations for image-to-image tasks is that the trajectory always \emph{originates from a fixed Gaussian prior}. The source image can only enter as a conditioning signal rather than as the starting point of the transport. This has two consequences. First, the output is never anchored to the observed pixels, making it hard to impose physical constraints on the generated result. Second, the stochastic starting point introduces residual noise and variability in the prediction.

Recently, LBM removes this constraint by building a stochastic interpolant directly between the source distribution $\pi_0$ and the target distribution $\pi_1$, learning a transport map between two arbitrary distributions instead of from noise~\cite{chadebec2025lbm}. In our case, $\pi_0$ corresponds to the distribution of natural images and $\pi_1$ to the distribution of albedo maps. Given a paired sample $(x_0, x_1) \sim \pi_0 \times \pi_1$, where it corresponds to an image and its albedo, the bridge is defined as
\begin{equation}
x_t = (1-t)\,x_0 + t\,x_1 + \sigma\sqrt{t(1-t)}\,\epsilon, \qquad \epsilon \sim \mathcal{N}(0, I),
\label{eq:bridge}
\end{equation}
and a network is trained to predict the corresponding drift, recovering $x_1$ from $x_0$ in as little as a single step. LBM applies this idea in the latent space of a pretrained autoencoder, achieving fast, even single-step, image-to-image translation across tasks~\cite{chadebec2025lbm, serrano2026synclight, li2026syncfix}. This makes LBM particularly well suited to IID: because the trajectory \emph{begins at the source image itself}, the prediction remains anchored to the observed pixels. As a result, we can decode the estimated target back to pixel space and impose an explicit reconstruction loss enforcing the image formation model, directly grounding the decomposition in physical constraints, which is precisely the property that diffusion and FM lack.

It is worth noting that the three formulations above form a hierarchy rather than three unrelated frameworks. Setting $\sigma = 0$ and replacing $x_0$ with a Gaussian sample reduces the bridge of Eq.~\ref{eq:bridge} to the deterministic FM interpolant, and further replacing the deterministic transport with the discrete noising chain recovers standard diffusion. In our experiments (Section~\ref{sec:experiments}) we exploit this relationship directly: the same UNet backbone and training pipeline are reused across all three variants, isolating the effect of the transport formulation itself, independent of architecture or capacity, on albedo estimation quality.

\section{Method}
\label{sec:method}
\begin{figure*}
    \centering
    \includegraphics[width=\linewidth]{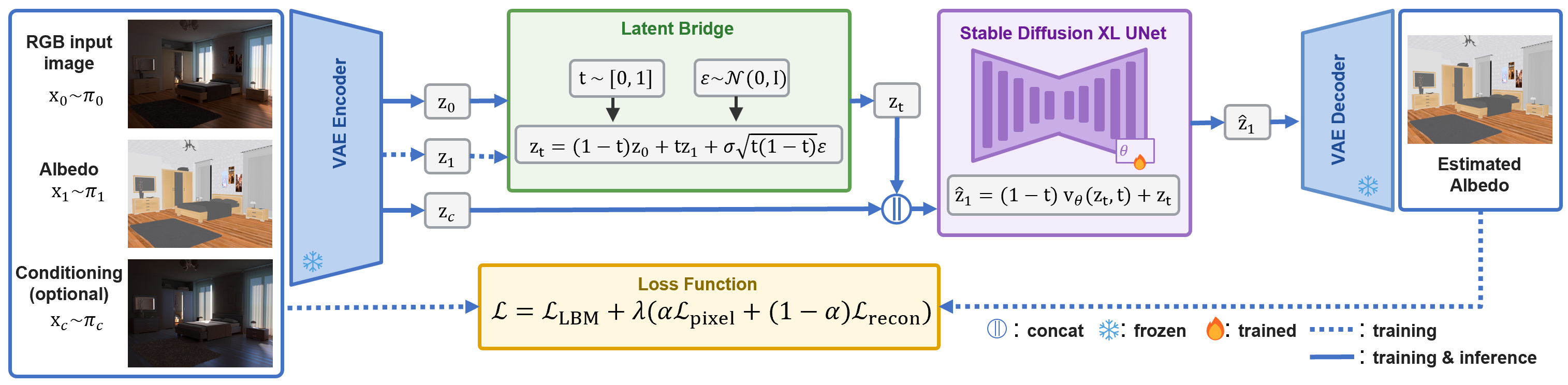}
    \caption{Method Overview. Input image $x_0$, ground-truth albedo $x_1$, and optional conditioning $x_c$ are mapped to the latent space through a frozen VAE encoder. The latent $z_0$ and $z_1$ are combined via a stochastic interpolant to produce $z_t$, which is concatenated with the conditioning $z_c$ and fed to a Stable Diffusion XL UNet. The UNet predicts the velocity field $v_\theta(z_t, t)$, from which the estimated target latent $\hat{z}_1$ is recovered and decoded by a frozen VAE decoder into the predicted albedo $\hat{x}_1$. Training combines the LBM objective $\mathcal{L}_{\text{LBM}}$ on the velocity field with a pixel-space term $\mathcal{L}_{\text{pixel}}$ and a reconstruction term $\mathcal{L}_{\text{recon}}$ enforcing the image formation model. }
    \label{fig:method}
\end{figure*}

Our method addresses albedo estimation through a generative formulation built upon Latent Bridge Matching (LBM). We first describe how LBM is adapted to the albedo estimation task, then detail the latent bridge formulation that defines the transport process, the reconstruction loss that grounds the decomposition in the image formation model, and finally the shading-conditioned variant that further disentangles illumination from reflectance. In Figure~\ref{fig:method} a global outline of the method is given.

\paragraph{Albedo estimation.}~We cast albedo estimation as an image-to-image translation between two domains: the observed RGB image and the corresponding ground-truth albedo, which is by definition independent of illumination effects such as shading, shadows, and specularities (Eq.~\ref{eq:image_formation}). As illustrated in Fig.~\ref{fig:method}, the source image $x_0 \sim \pi_0$ is the observed RGB input and the target $x_1 \sim \pi_1$ is the ground-truth albedo. Both are mapped into a compressed latent space through the frozen encoder $\mathcal{E}$ of a pretrained variational autoencoder (VAE), producing the latent representations $z_0 = \mathcal{E}(x_0)$ and $z_1 = \mathcal{E}(x_1)$. The transport between $z_0$ and $z_1$ is modeled by a trainable drift network, and the predicted target latent $\hat{z}_1$ is decoded back to pixel space through the frozen VAE decoder $\mathcal{D}$ to obtain the estimated albedo $\hat{x}_1 = \mathcal{D}(\hat{z}_1)$. Crucially, because the trajectory starts from the input image rather than from Gaussian noise, the prediction remains anchored to the observed pixels throughout the process.

\vspace{-5mm}
\paragraph{Latent Bridge Matching formulation.}
With the source and target latents $z_0$ and $z_1$ defined above, we instantiate the bridge of Eq.~\ref{eq:bridge} directly in latent space,
where $\sigma \ge 0$ controls the variance of the transport path and $t \sim \pi(t)$.
In contrast to the continuous uniform schedule used in diffusion training, we sample
$t$ from a small set of equally spaced timesteps. This aligns the timesteps seen at
training and inference and caps sampling at four steps, which is the primary source of
LBM's efficiency over conventional diffusion models.

The interpolated latent $z_t$ is passed to a drift network $v_\theta(z_t, t)$ that
predicts the velocity pointing toward the clean target latent, from which the target
is recovered in closed form,
\begin{equation}
    \hat{z}_1 = (1-t)\,v_\theta(z_t, t) + z_t .
    \label{eq:recover}
\end{equation}
The network is trained with the bridge-matching regression objective
\begin{equation}
    \mathcal{L}_{\mathrm{LBM}}
    = \mathbb{E}_{t, z_t}\!\left[\,
        \lVert v_\theta(z_t, t) - v(z_t, t) \rVert^{2}
    \,\right],
    \label{eq:lbm}
\end{equation}
where $v(z_t, t)$ is the target drift induced by Eq.~\ref{eq:recover}. We take
$\mathcal{E}$ and $\mathcal{D}$ to be the frozen VAE of Stable Diffusion XL and
implement $v_\theta$ as a UNet initialized from the same pretrained backbone.

\vspace{-5mm}
\paragraph{Reconstruction Loss.}\label{par:recons_loss} The latent objective in Eq.~\ref{eq:lbm} supervises the transport in latent space, but does not guarantee that the decoded albedo is consistent with the input image under the intrinsic image formation model. Because LBM operates on a pixel-aligned trajectory, we can decode the predicted latent and impose supervision directly in pixel space. We therefore augment the objective with two pixel-level terms: a supervised term $\mathcal{L}_{\text{pixel}}$ comparing the decoded prediction $\hat{x}_1$ against the ground-truth albedo $x_1$, and a reconstruction term $\mathcal{L}_{\text{recon}}$ that enforces physical consistency with the image formation model. The full training objective is
\begin{equation}
\mathcal{L} = \mathcal{L}_{\text{LBM}} + \lambda\left(\alpha\,\mathcal{L}_{\text{pixel}} + (1-\alpha)\,\mathcal{L}_{\text{recon}}\right),
\label{eq:full_loss}
\end{equation}
where $\lambda$ weights the pixel-space supervision against the latent objective and $\alpha \in [0,1]$ balances the supervised and reconstruction terms. For $\mathcal{L}_{\text{pixel}}$ we consider the perceptual LPIPS loss. The reconstruction term is defined as the distance between the input image $I$ and the image re-synthesized from the estimated albedo,
\begin{equation}
\mathcal{L}_{\text{recon}} = \mathcal{L}\!\left(I,\ \hat{A} \cdot S\right) \quad\text{or}\quad \mathcal{L}\!\left(I,\ \hat{A} \cdot S + R\right),
\label{eq:recon}
\end{equation}
where $\hat{A}$ is the estimated albedo (i.e. $\hat{x}_1$), $S$ the shading, and $R$ a residual term capturing non-Lambertian effects. The choice between the two forms depends on the dataset: the Lambertian product model (Eq.~\ref{eq:image_formation}) or its residual extension. This ability to attach a physically motivated, pixel-level supervision signal is precisely what makes LBM well suited to IID, where the image formation model operates directly at the pixel level.

\vspace{-5mm}
\paragraph{Conditioning Albedo Estimation}
The LBM framework naturally admits a conditional extension. In addition to the paired samples $(x_0, x_1)$, an auxiliary conditioning image $x_c$ is introduced to guide the transport. As shown in Fig.~\ref{fig:method}, $x_c$ is encoded with the frozen VAE into a latent $z_c$ and concatenated with the interpolated latent $z_t$ along the channel dimension, yielding the joint input $[z_t, z_c]$ to the drift network. This lets $v_\theta$ predict the drift while explicitly accounting for the additional cue, improving controllability and reducing ambiguity in the estimated albedo.

To condition the albedo estimation process, we explore several alternatives based on complementary intrinsic components, namely surface normals---proven useful for related problems such as lighting ambient normalization \cite{promptnorm2025}---and shading. As a first approach, we employ a foundation model for normal estimation based on Marigold~\cite{ke2025marigold} . Alternatively, we train two LBM-based models following the architecture shown in Figure~\ref{fig:method}: one for surface normal estimation and another for shading estimation. In Experiment~3, we evaluate these alternatives; our results indicate that shading-based conditioning provides the most effective guidance for albedo estimation among the alternatives considered. Providing the network with shading supplies an explicit signal indicating which intensity variations in the input are attributable to illumination, allowing the model to focus its capacity on recovering the residual reflectance. This improves the disentanglement between illumination and surface color and reduces the tendency to bake soft shadows and inter-reflections into the predicted albedo. To make this conditioning usable at inference, when ground-truth shading is unavailable, we pair it with a complementary shading estimation model that acts as an efficient conditioner, yielding a physically grounded two-stage decomposition pipeline. We note that this design introduces a two-stage dependency: the final albedo quality is inherently bounded by the accuracy of the shading estimator itself, since any error in the predicted shading is propagated as conditioning noise to the albedo model. We discuss this dependency, and a partial mitigation, further in Experiment~3.

\section{Experiments and Results}
\label{sec:experiments}
In this section we propose a set of experiments to evaluate the initial hypothesis and the proposed methods.

\vspace{-5mm}
\subsection{Experimental setup}
Models are trained on InteriorVerse ($\approx 44K$ images) and Hypersim  ($\approx 59K$ images) datasets for 50 epochs. InteriorVerse provides ground-truth albedo and surface normal maps, shading is computed using image formation model defined in equations \ref{eq:image_formation} and \ref{eq:full_loss} Hypersim provides ground-truth albedo, shading, surface normals, and an additional residual term capturing non-Lambertian effects.

The encoder $\mathcal{E}$ and decoder $\mathcal{D}$ are taken from the pre-trained variational autoencoder (VAE) of Stable Diffusion XL \cite{podell2023sdxl} and are kept frozen during training to preserve the learned latent representation. The parametrized drift function $v_0$ is implemented as a U-Net, initialized from the pre-trained text-to-image Stable Diffusion XL model. The model is optimized using the AdamW optimizer with a learning rate of $4 \times 10^{-5}$.

To ensure compatibility with the pre-trained VAE, all input images and intrinsic properties are normalized to the range $[-1,1]$. During training, we apply random cropping to $256 \times 256$ patches without padding. At inference time, all images are processed at their original 
resolution without any cropping or resizing, allowing the model to 
generalize to arbitrary input dimensions.

\subsection{Evaluation datasets and metrics}
\label{sec:metrics}
We evaluate on five benchmarks spanning real controlled, synthetic, and real in-the-wild scenarios: MIT Intrinsic Images (220 images), IIW (1{,}046 test images), ARAP (149 images), and the InteriorVerse (2{,}672 test images) and Hypersim (7{,}690 test images) held-out splits. Where prior work provides evaluation masks we follow the established protocol for each dataset; Hypersim, which has no mask annotations, is evaluated on the full image. Because no single dataset provides every ground-truth component alongside a consistent metric set, we report the following measures, matching the protocol used in prior IID literature for each benchmark.

\emph{MSE} is the mean squared error between the predicted albedo $\hat{A}$ and the ground truth $A$ over the $N$ evaluated pixels, $\text{MSE} = \frac{1}{N}\sum_i (\hat{A}_i - A_i)^2$. \emph{LMSE} (Local MSE) computes this same quantity within overlapping local windows $\Omega_k$ before averaging, which reduces sensitivity to global scale ambiguities inherent to albedo estimation. \emph{PSNR} is derived from MSE as $\text{PSNR} = 10\log_{10}(\text{MAX}^2/\text{MSE})$. \emph{DSSIM} is the structural dissimilarity, $\text{DSSIM} = (1-\text{SSIM})/2$, where SSIM compares luminance, contrast, and structure between the two images. \emph{LPIPS} measures perceptual similarity using weighted differences between deep features extracted from a pretrained network across layers $l$ and spatial locations. \emph{WHDR} (Weighted Human Disagreement Rate), used for IIW, is the weighted fraction of pairwise human judgments about relative reflectance that the predicted albedo contradicts, thresholded at $\delta$; we report it at the two thresholds standard in the literature, $10\%$ and $20\%$. For all error-based metrics ($\downarrow$) lower is better, while PSNR ($\uparrow$) is better when higher.

\subsection{Experiment 1. Baseline}
\begin{figure}[tbp]
\centering

\begin{minipage}{0.19\linewidth}
    \centering
    \includegraphics[width=\linewidth]{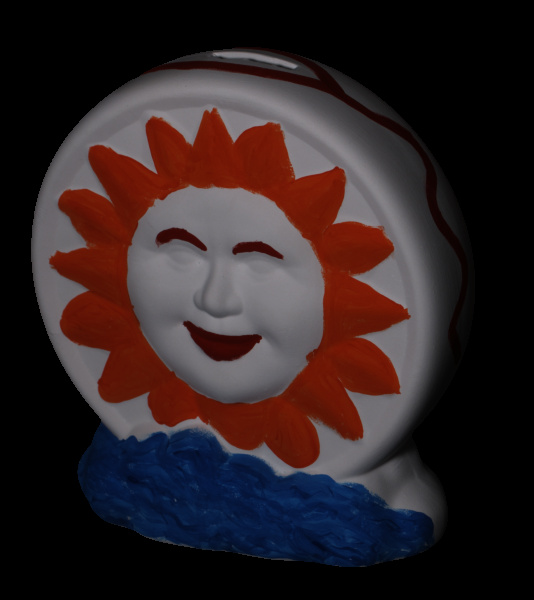}
\end{minipage}
\begin{minipage}{0.19\linewidth}
    \centering
    \includegraphics[width=\linewidth]{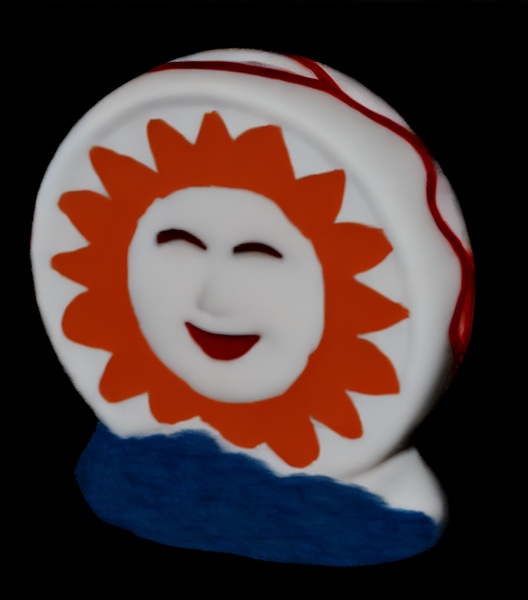}
\end{minipage}
\begin{minipage}{0.19\linewidth}
    \centering
    \includegraphics[width=\linewidth]{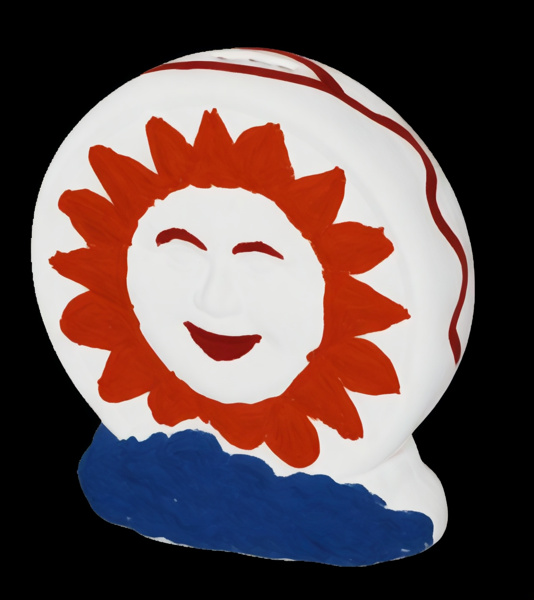}
\end{minipage}
\begin{minipage}{0.19\linewidth}
    \centering
    \includegraphics[width=\linewidth]{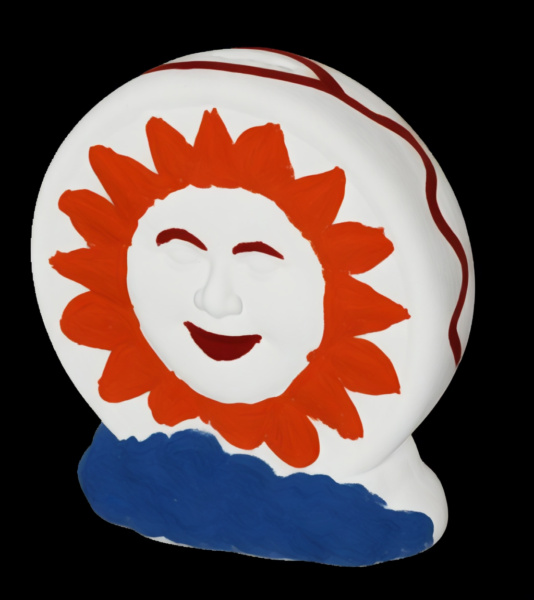}
\end{minipage}
\begin{minipage}{0.19\linewidth}
    \centering
    \includegraphics[width=\linewidth]{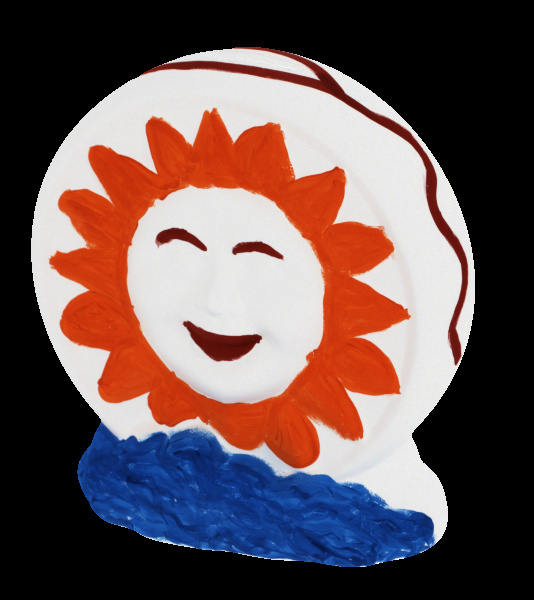}
\end{minipage}

\vspace{0.2cm}

\begin{minipage}{0.19\linewidth}
    \centering
    \includegraphics[width=\linewidth]{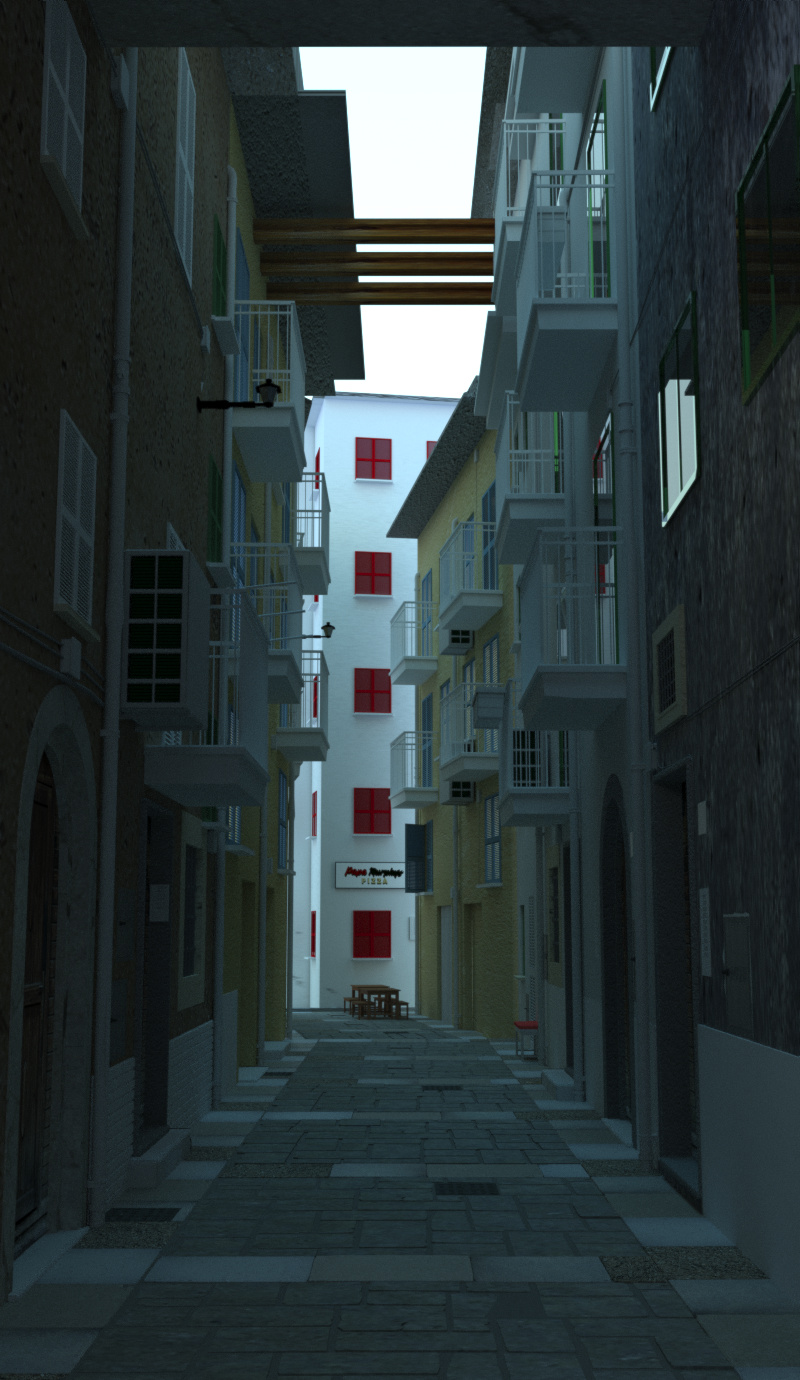}
\end{minipage}
\begin{minipage}{0.19\linewidth}
    \centering
    \includegraphics[width=\linewidth]{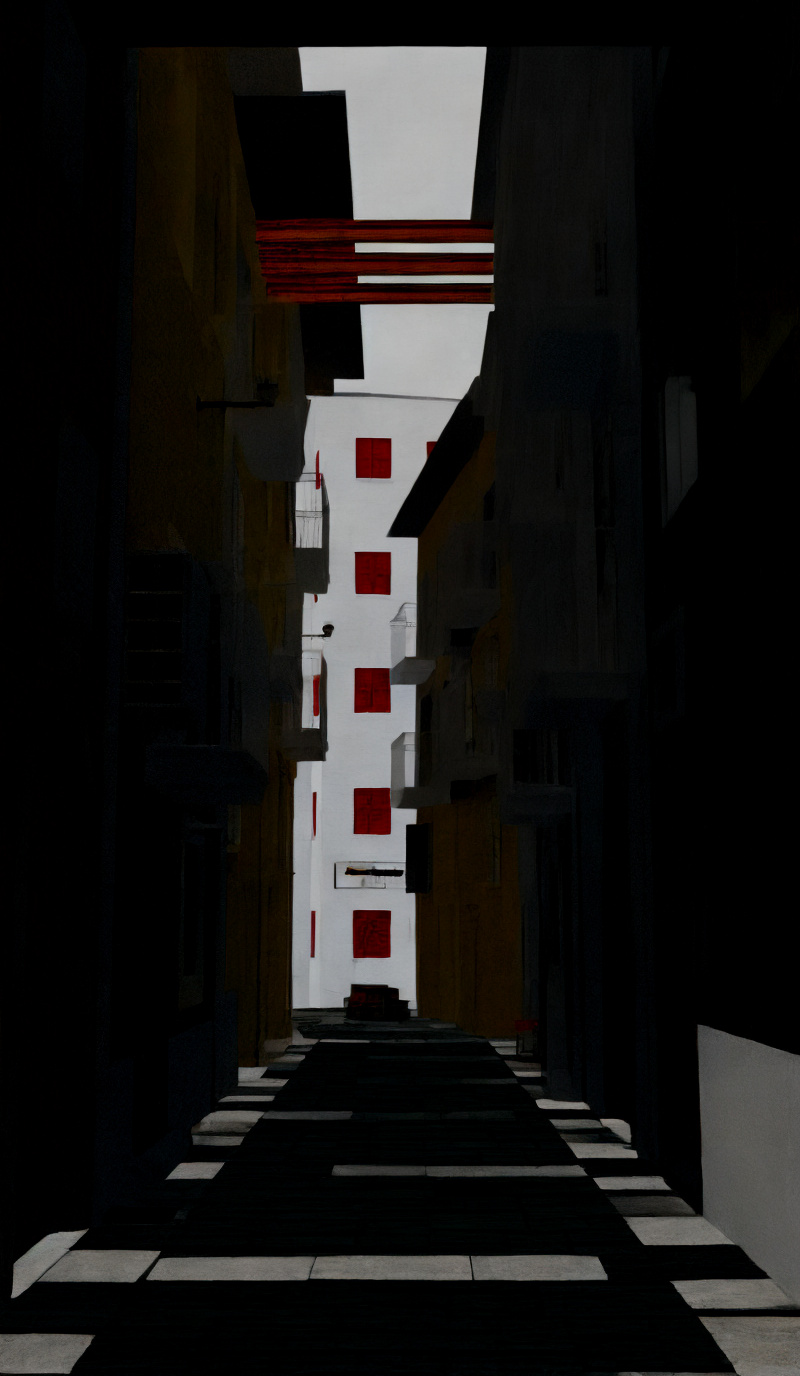}
\end{minipage}
\begin{minipage}{0.19\linewidth}
    \centering
    \includegraphics[width=\linewidth]{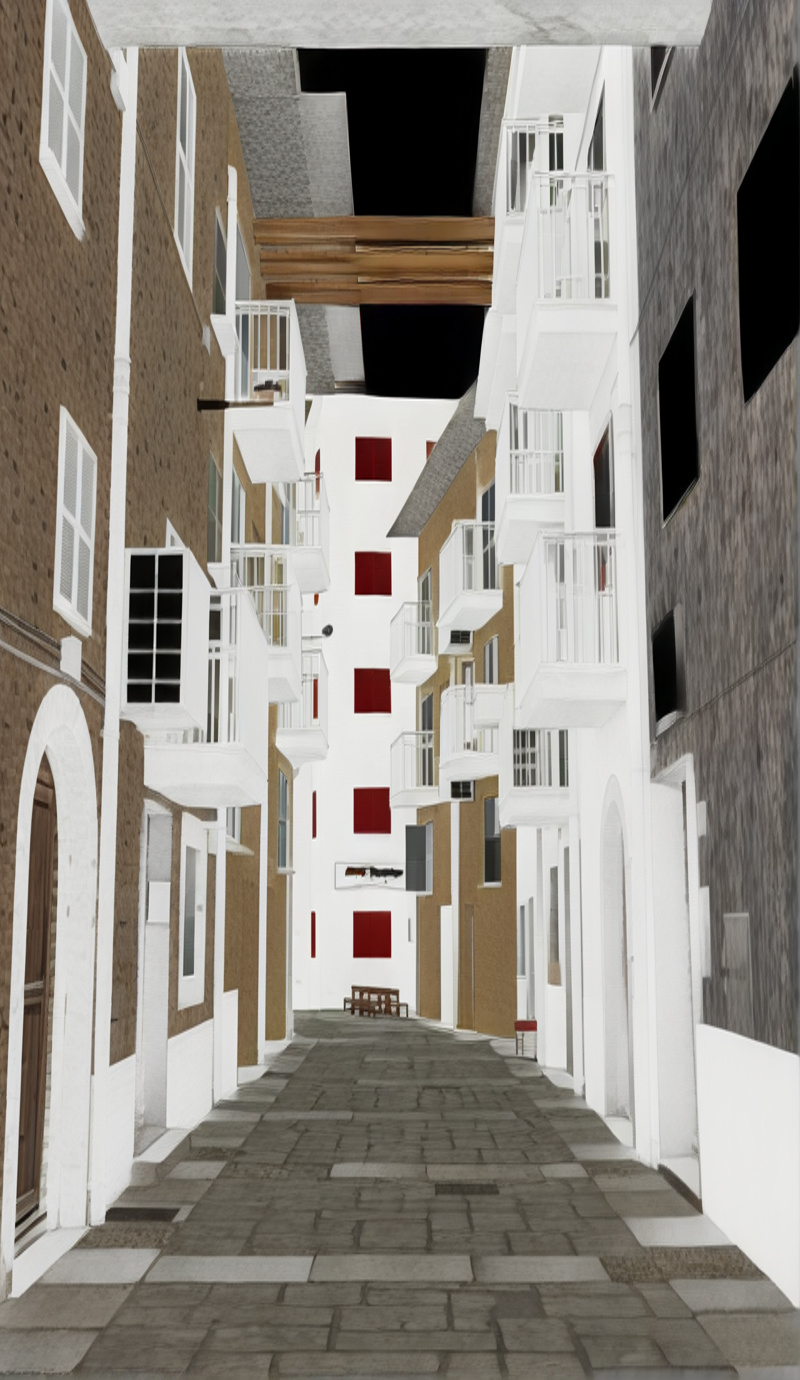}
\end{minipage}
\begin{minipage}{0.19\linewidth}
    \centering
    \includegraphics[width=\linewidth]{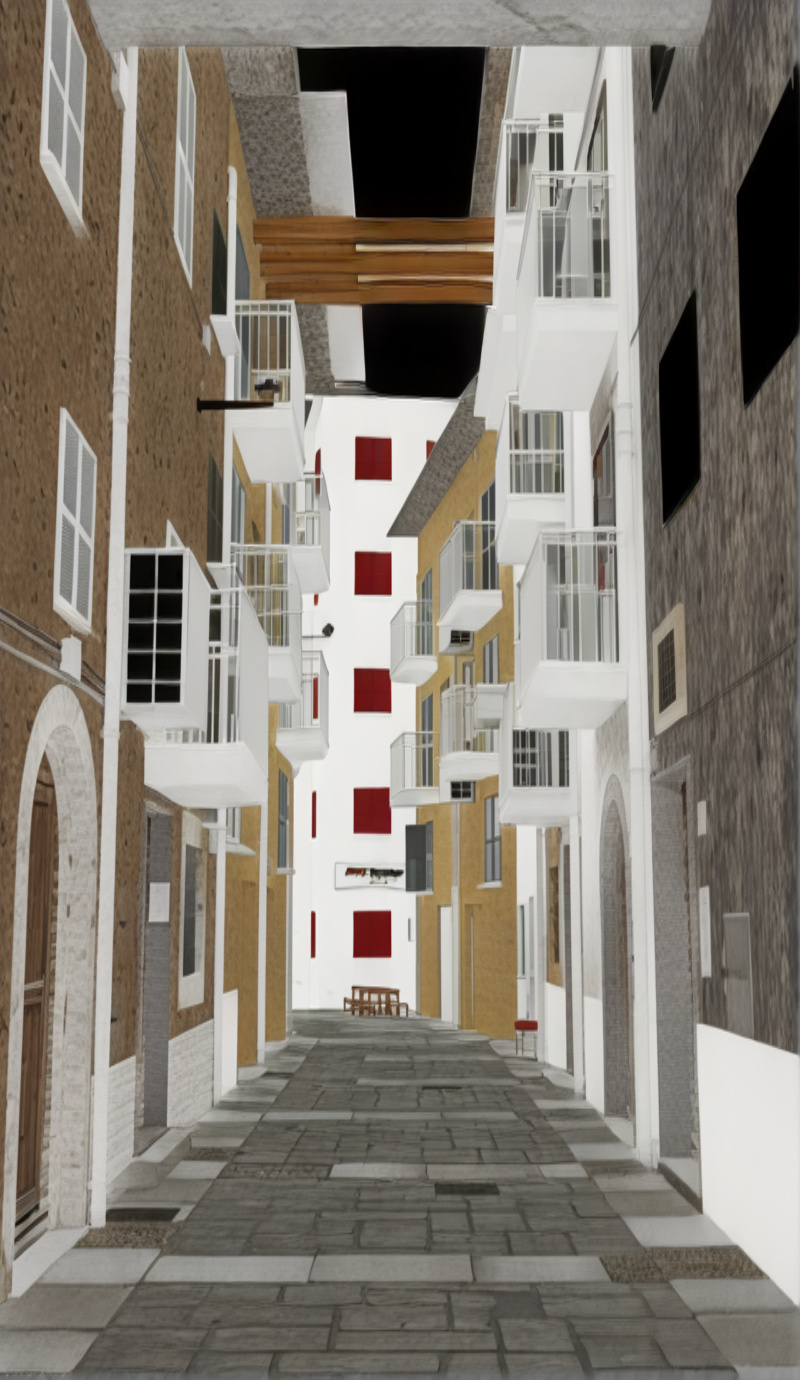}
\end{minipage}
\begin{minipage}{0.19\linewidth}
    \centering
    \includegraphics[width=\linewidth]{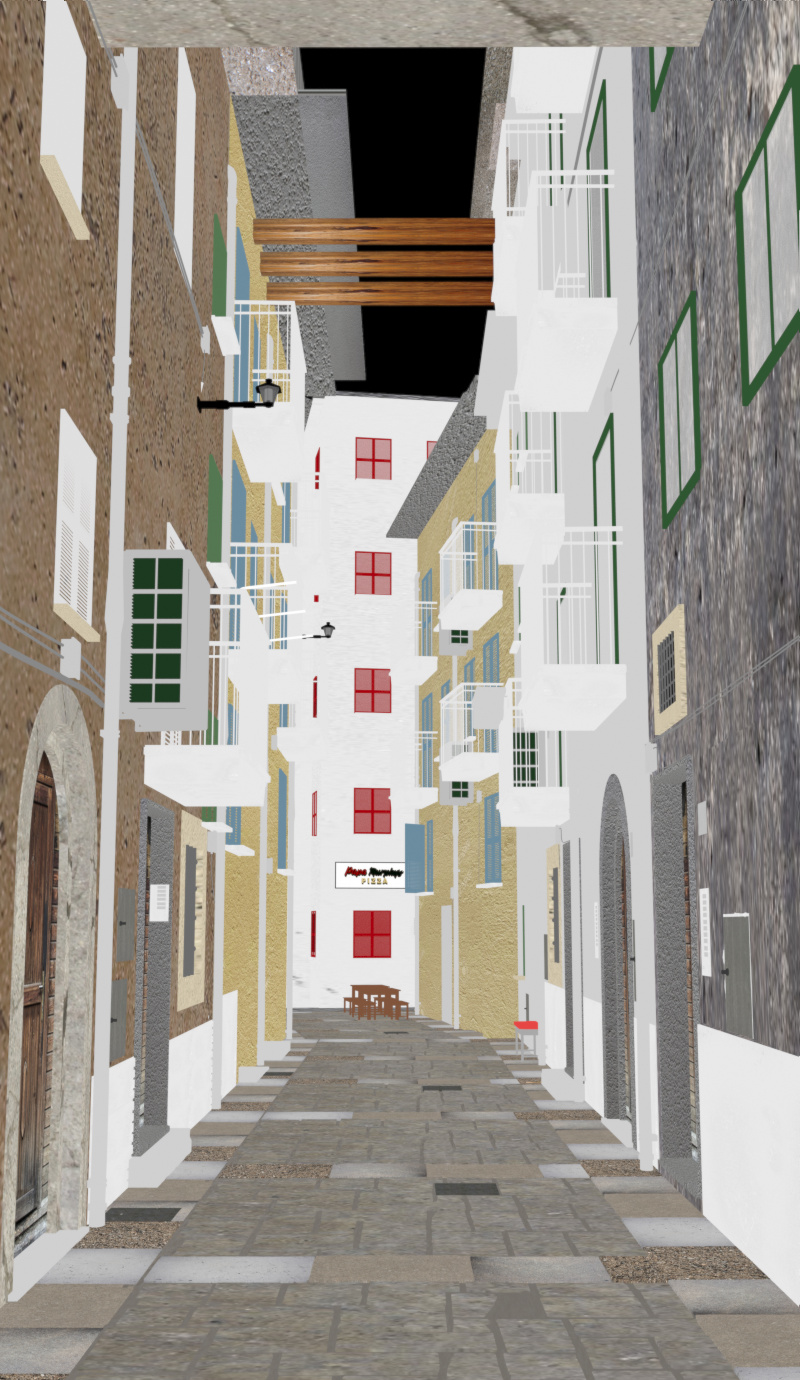}
\end{minipage}

\vspace{0.2cm}

\begin{minipage}{0.19\linewidth}
    \centering
    \includegraphics[width=\linewidth]{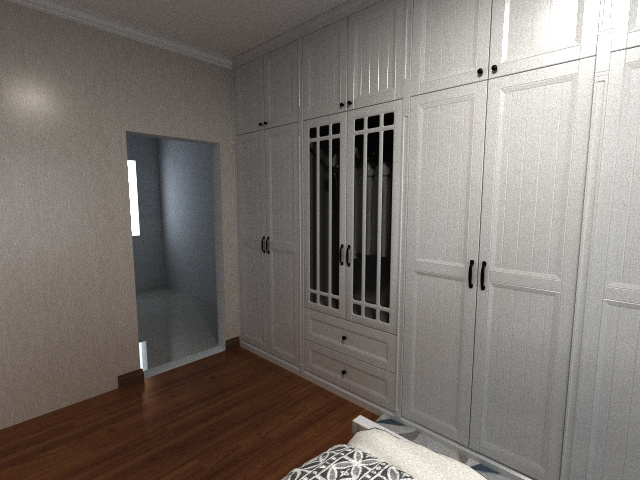}
\end{minipage}
\begin{minipage}{0.19\linewidth}
    \centering
    \includegraphics[width=\linewidth]{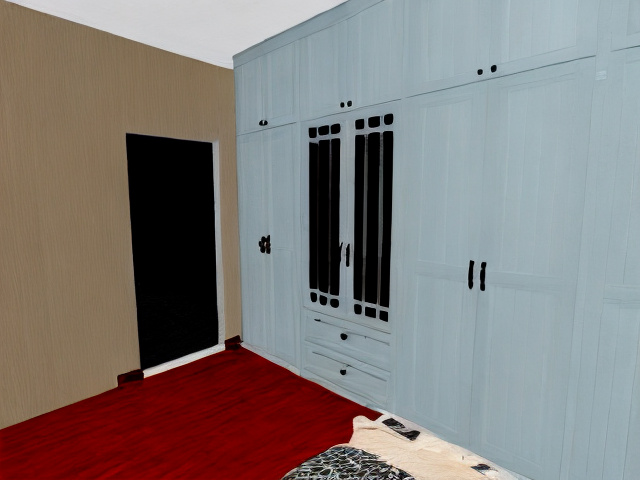}
\end{minipage}
\begin{minipage}{0.19\linewidth}
    \centering
    \includegraphics[width=\linewidth]{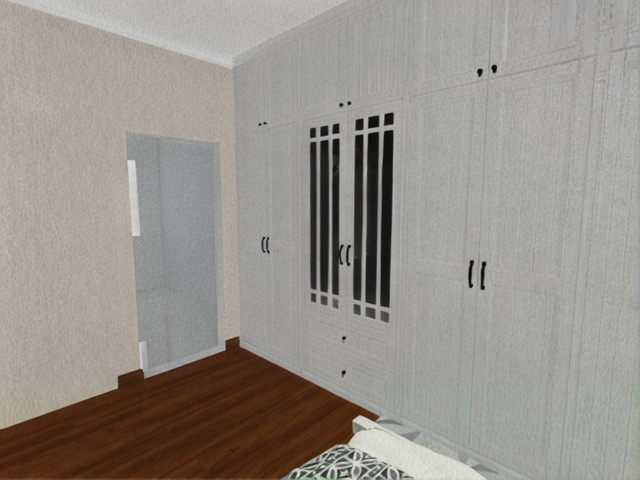}
\end{minipage}
\begin{minipage}{0.19\linewidth}
    \centering
    \includegraphics[width=\linewidth]{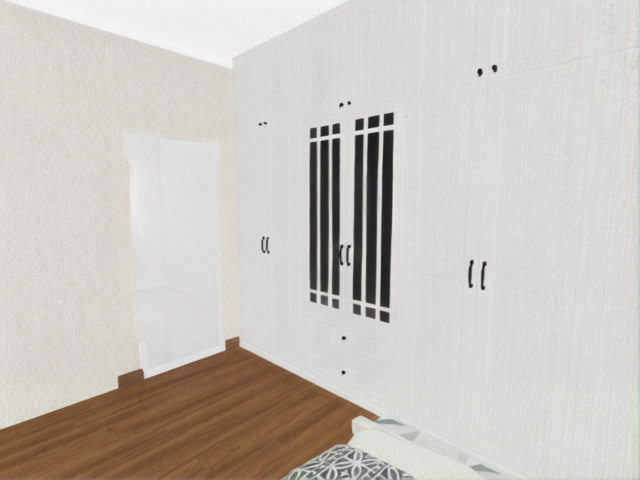}
\end{minipage}
\begin{minipage}{0.19\linewidth}
    \centering
    \includegraphics[width=\linewidth]{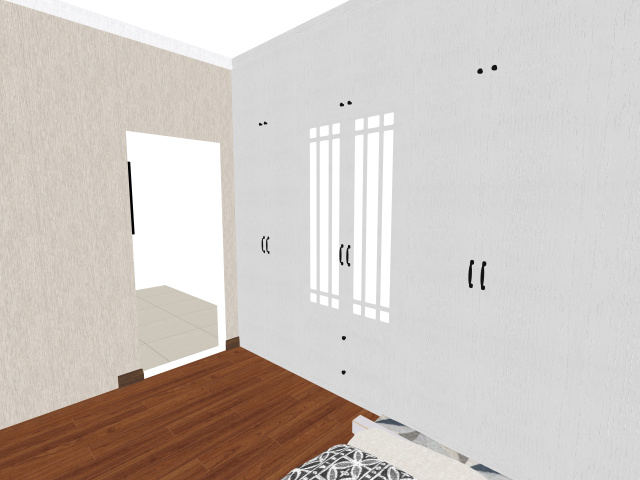}
\end{minipage}

\vspace{0.2cm}

\begin{minipage}{0.19\linewidth}
    \centering
    \includegraphics[width=\linewidth]{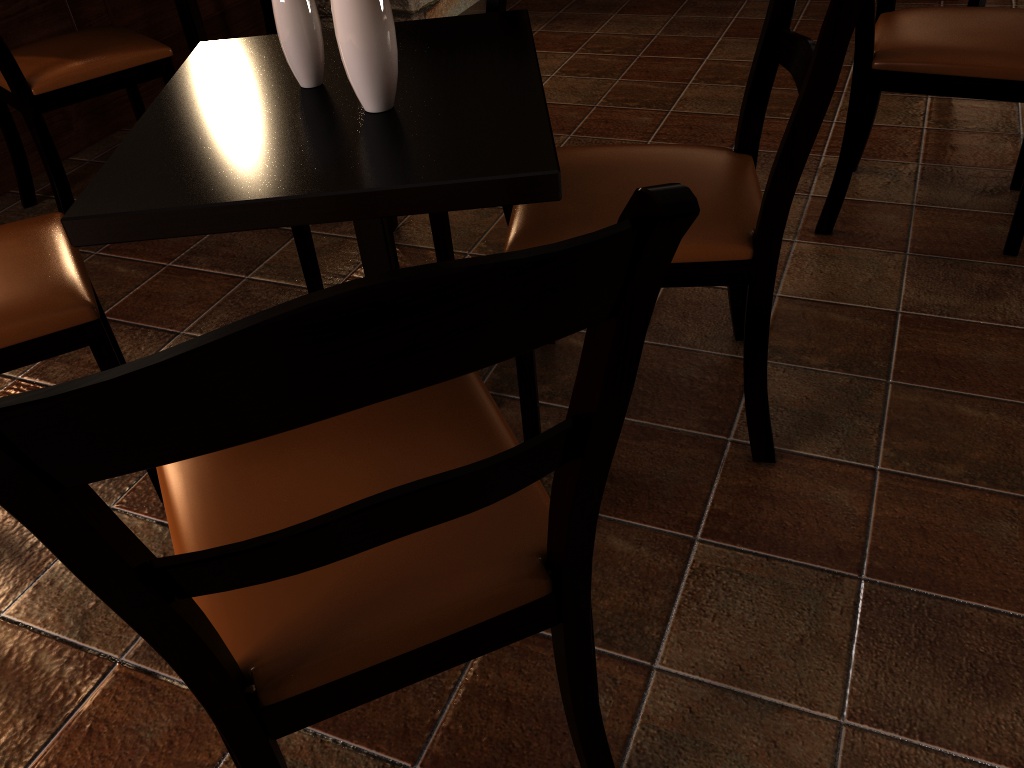}
\end{minipage}
\begin{minipage}{0.19\linewidth}
    \centering
    \includegraphics[width=\linewidth]{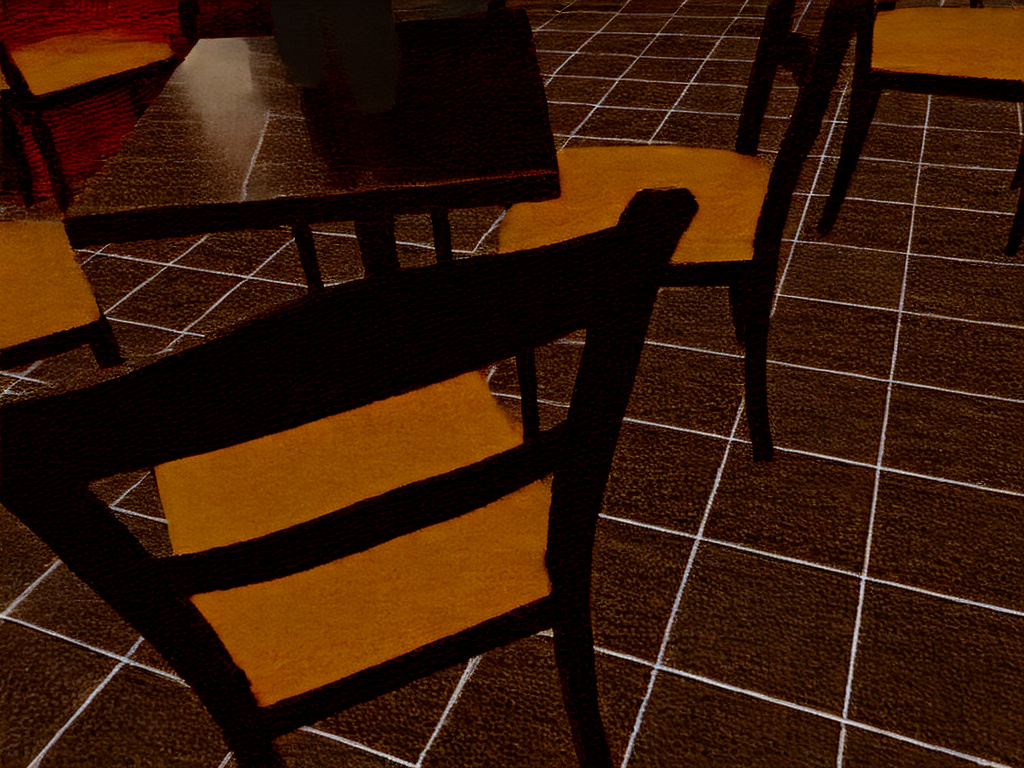}
\end{minipage}
\begin{minipage}{0.19\linewidth}
    \centering
    \includegraphics[width=\linewidth]{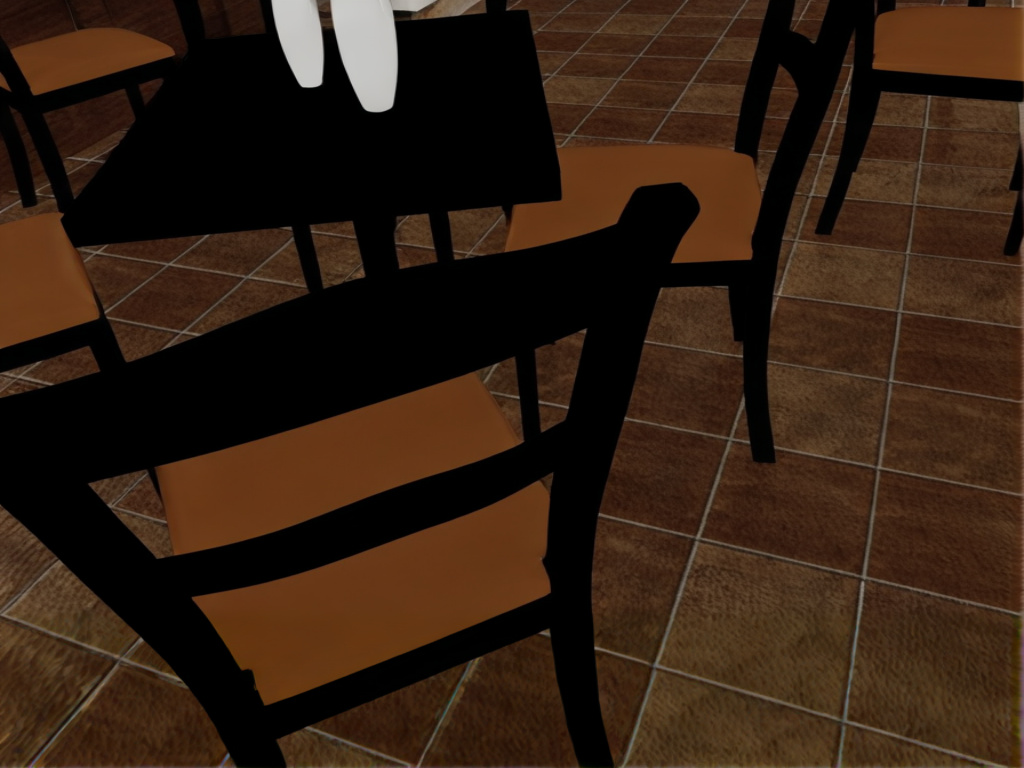}
\end{minipage}
\begin{minipage}{0.19\linewidth}
    \centering
    \includegraphics[width=\linewidth]{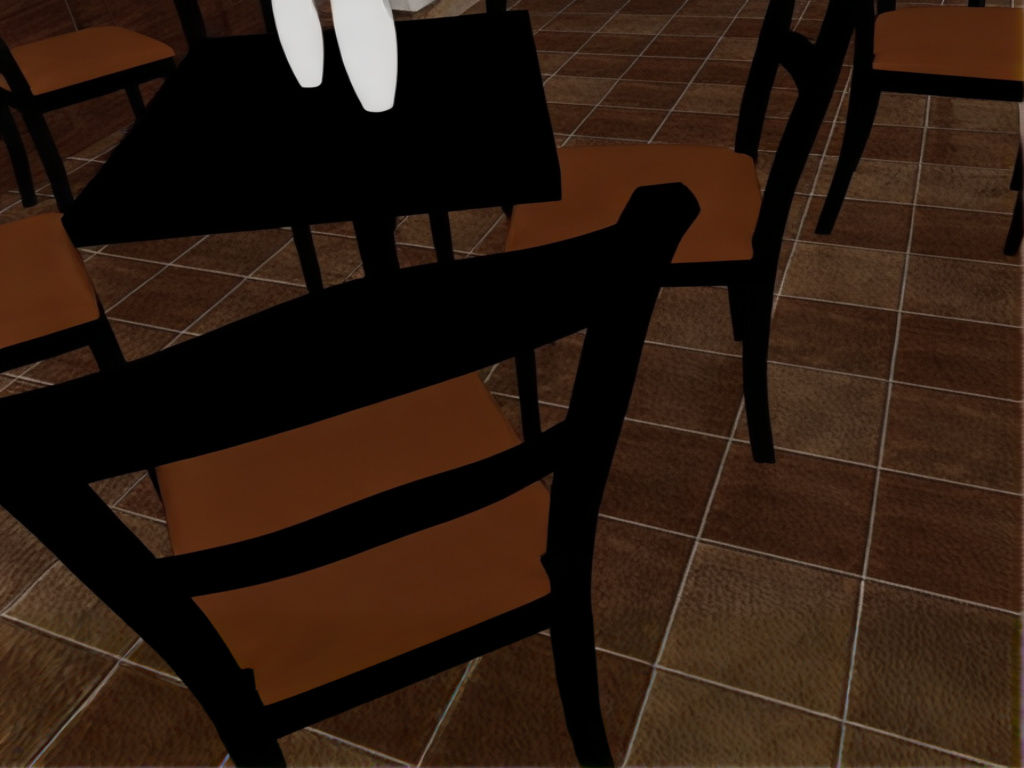}
\end{minipage}
\begin{minipage}{0.19\linewidth}
    \centering
    \includegraphics[width=\linewidth]{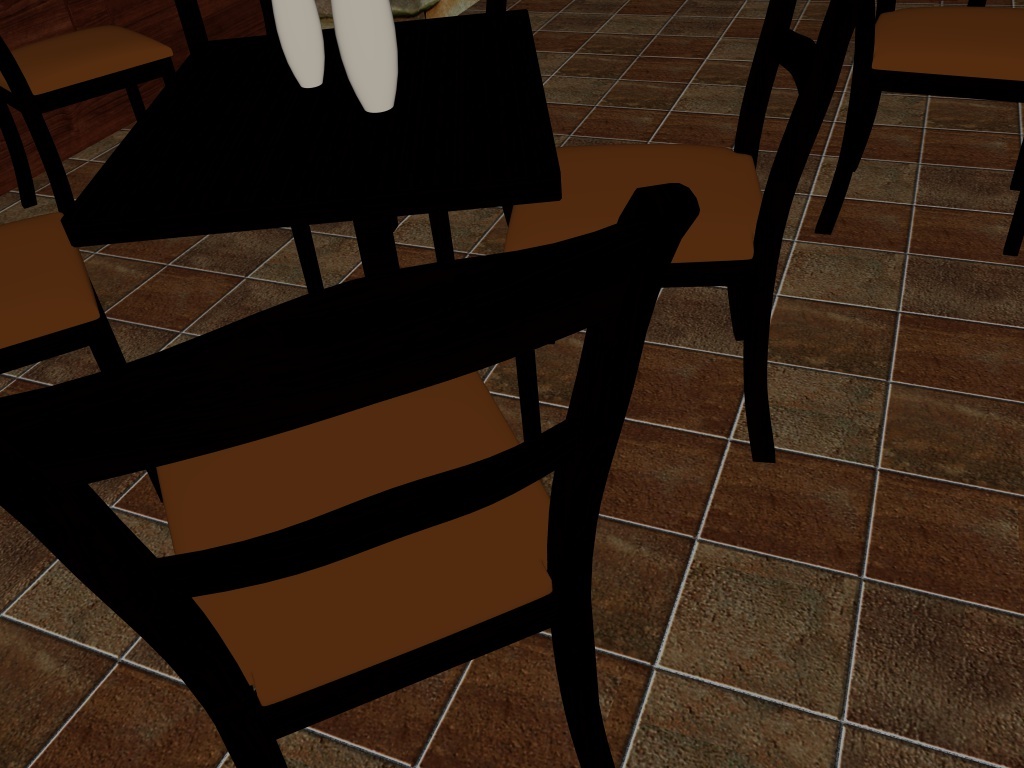}
\end{minipage}

\vspace{0.15cm}

\begin{minipage}{0.19\linewidth}
    \centering Input
\end{minipage}
\begin{minipage}{0.19\linewidth}
    \centering SD-AID
\end{minipage}
\begin{minipage}{0.19\linewidth}
    \centering FD-AID
\end{minipage}
\begin{minipage}{0.19\linewidth}
    \centering LBM-AID
\end{minipage}
\begin{minipage}{0.19\linewidth}
    \centering GT Albedo
\end{minipage}

\caption{Qualitative comparison across MIT, ARAP, InteriorVerse and Hypersim datasets. From left to right: input image, estimated albedo for SD-AID, FD-AID and LBM-AID and ground-truth albedo.}
\label{fig:qualitative_comparison}
\end{figure}

In this first experiment, we evaluated our initial hypothesis claiming LBM as a suitable framework to improve inference time efficiency and albedo estimation accuracy compared to previous generative formulations. In Table~\ref{tab:generative_models_comparison} we show the results for three different generative architectures trained for albedo estimation. These are SD-AID (Stable Diffusion for Albedo Intrinsic Decomposition), likewise FM-AID and LBM-AID denote FM and LBM in the generative process. All three methods share a similar backbone architecture and are trained under identical conditions as described above. We observe that the LBM-AID model achieves the strongest performance for albedo estimation in most settings and the inference time is equally reduced by LBM and FM, compared to the inefficiency of the SD approach. Some albedo estimation results are qualitatively shown in Figure \ref{fig:qualitative_comparison} for different datasets.

\begin{table*}[tbp]
\centering
\resizebox{\textwidth}{!}{%
\begin{tabular}{lccccccccc}
\toprule
& \multicolumn{3}{c}{MIT}
& \multicolumn{2}{c}{IIW}
& \multicolumn{3}{c}{Hypersim}
& \multirow{2}{*}{\shortstack{Inf.Time\\(s/img)}}\\
\cmidrule(lr){2-4}
\cmidrule(lr){5-6}
\cmidrule(lr){7-9}
\multicolumn{1}{c}{Framework}
& PSNR$\uparrow$ & MSE$\downarrow$ & Recon. MSE$\downarrow$
& WHDR\small{(10\%)}$\downarrow$ & WHDR\small{(20\%)}$\downarrow$
& PSNR$\uparrow$ & LPIPS$\downarrow$ & Recon. MSE$\downarrow$
& \\
\midrule
SD-AID \small{\textit{(10)}}
& 13.96 & 0.05 & 0.0215
& 27.12 & 21.90
& 11.55 & 0.50 & 0.1346
&0.77 \\

SD-AID \small{\textit{(50)}}
& 14.18 & 0.05 & 0.0204
& 28.21 & 23.66
& 12.11 & 0.49 & 0.1266
& 2.80 \\

SD-AID-FM \small{\textit{(10)}}
& 13.00 & 0.07 & 0.0256
& 30.37 & 25.12
& 11.08 & 0.58 & 0.1214
& 0.77 \\

FM-AID
& 19.92 & 0.02 & 0.0129
& 32.54 & 27.83
& 15.40 & 0.32 & 0.0341
& 0.77 \\

LBM-AID
& 21.51 & 0.01 & 0.0111
& 26.13 & 22.49
& 15.51 & 0.30 & 0.0362
& 0.77 \\
\bottomrule
\end{tabular}}
\caption{\textit{Table 1}. Performance of different generative models on MIT, IIW, and Hypersim datasets. Reconstruction error is computed between the input image and its estimated version $\hat{I} = \hat{A}\cdot\hat{S}$. SD-AID models were trained on Stable Diffusion 2.1 with a denoising-diffusion objective (10 and 50 inference steps respectively); SD-AID-FM shares the same SD2.1 backbone but is trained with a flow-matching objective instead, isolating the effect of the training objective from that of the backbone. FM-AID and LBM-AID were trained using Stable Diffusion XL as velocity estimator.}
\vspace{2mm}
\label{tab:generative_models_comparison}
\end{table*}

\subsection{Experiment 2. Setting LBM parameters}

The second experiment is devoted to analysing the impact of the reconstruction loss term introduced in Equation \ref{eq:recon}. We examine the influence of the weighting coefficient $\alpha$, which balances the contribution between the supervised pixel loss and the reconstruction-based consistency term. Table~\ref{tab:parsetting} makes us decide $\alpha=0.9$ as the best tradeoff for all datasets and will be used in the subsequent experiments. 

\begin{table}[tbp]
\centering\resizebox{\linewidth}{!}{\begin{tabular}{ccccccc}
\toprule
& \multicolumn{2}{c}{MIT} 
& \multicolumn{2}{c}{ARAP} 
& \multicolumn{2}{c}{Hypersim} \\
\cmidrule(lr){2-3}
\cmidrule(lr){4-5}
\cmidrule(lr){6-7}
$\boldsymbol{\alpha}$ 
& DSSIM$\downarrow$ & MSE$\downarrow$
& DSSIM$\downarrow$ & MSE$\downarrow$ 
& PSNR$\uparrow$ & LPIPS$\downarrow$ \\
\midrule
1.0 & 0.0490 & 0.0100 & 0.1911 & 0.0371 & 15.51 & 0.30 \\
0.9 & 0.0481 & 0.0097 & 0.1904 & 0.0670 & 16.09 & 0.30 \\
0.8 & 0.0637 & 0.0239 & 0.1911 & 0.0708 & 16.18 & 0.30 \\
0.7 & 0.0554 & 0.0140 & 0.1520 & 0.0284 & 14.07 & 0.32 \\
0.5 & 0.0732 & 0.0291 & 0.1568 & 0.0324 & 14.39 & 0.32 \\
0.0 & 0.0911 & 0.0411 & 0.1772 & 0.0368 & 9.72 & 0.51 \\
\bottomrule
\end{tabular}}
\caption{\textit{Table 2}. Parameter setting  of the reconstruction loss weight ($\alpha$, Equation \ref{eq:full_loss}) on MIT, ARAP, and Hypersim datasets. $\lambda = 10$ is fixed throughout.}\label{tab:parsetting}
\end{table}

\subsection{Experiment 3. Model Conditioning}
As discussed earlier, an important advantage of LBM is the ease with which additional image-based information can be incorporated to condition the albedo estimation process. In this experiment, we explore two conditioning strategies based on the estimation of complementary intrinsic scene components, namely surface normals and image shading.

For surface normal estimation, we consider two alternatives: (a) using the foundation model proposed by Marigold et al.~\cite{ke2025marigold}, and (b) training an LBM-based model for normal estimation following the same framework as in LBM-AID (we denote as LBM-NID for Normal Intrinsic Decomposition). In addition, we employ this same framework to train a shading estimation model (we denote it as LBM-SID for Shading Intrinsic Decomposition), which serves as a third conditioning option. 

Results are summarized in Table~\ref{tab:conditioning_ablation}, from which two main conclusions can be drawn. First, conditioning does not always guarantee improved performance. In particular, surface normals appear to provide no benefit for albedo estimation, whereas shading proves to be a more effective conditioning term. Second, the best performance among the conditioning strategies tested is achieved when conditioning is based on shading. As shown in Table~\ref{tab:shading_arap}, the most effective shading conditioning arises from the best shading estimates, which are obtained when the LBM-SID model is trained without reconstruction loss and without normal-based conditioning. This is the opposite trend to what we observe for albedo estimation (Table~\ref{tab:generative_models_comparison}), where the reconstruction loss consistently improves performance. A plausible explanation is that shading is a smooth, low-frequency signal, whereas the reconstruction term enforces pixel-level fidelity to the output image $I$, a criterion naturally aligned with recovering high-frequency reflectance detail. When applied to shading, this same pixel-wise constraint is comparatively under-determined: many locally smooth shading fields can satisfy $I = A \cdot S$ equally well once $A$ is fixed, so the reconstruction signal contributes noisier, less informative gradients than the direct supervised loss.

This two-stage dependency is visible in our own results: the shading quality differences across LBM-SID variants (Table~\ref{tab:shading_arap}) propagate directly to the downstream albedo estimates (Table~\ref{tab:conditioning_ablation}), where only the better shading estimators yield a net improvement over unconditioned albedo estimation. We partially mitigate this coupling in Section~\ref{par:recons_loss} by conditioning the shading estimator itself on a first-pass albedo prediction, which improves the auxiliary shading estimate and, in turn, the final albedo (Table~\ref{tab:image_reconstruction}); however, this remains a two-pass approximation rather than a joint resolution of the circularity, since any error in the first-pass albedo can still degrade the retrained shading estimator and, consequently, the final prediction. Fully decoupling this dependency, for instance through joint end-to-end training of both models or an uncertainty-aware conditioning mechanism, remains a direction for future work.

Because shading and albedo are two views of the same decomposition, we further ask whether the dependency can be made bidirectional: instead of conditioning albedo estimation on a shading estimate alone, we also condition the shading estimator LBM-SID on the albedo predicted by LBM-AID. As Table~\ref{tab:shading_arap} shows, this albedo-conditioned shading estimator improves over every previous LBM-SID variant on ARAP, and feeding its output back as the conditioning signal for albedo estimation (Table~\ref{tab:image_reconstruction}) yields the lowest reconstruction error we observe on all three datasets (MIT: 0.0104, ARAP: 0.0139, Hypersim: 0.0250). This confirms that albedo and shading are mutually informative signals under the LBM framework: each component provides a physically meaningful conditioning cue for estimating the other. We use this configuration -- LBM-AID conditioned on the albedo-conditioned LBM-SID -- as our best model in the comparison against state-of-the-art methods in Section~\ref{sec:sota}.

Finally, we revisit our initial hypothesis concerning adherence to the image formation model defined in Equation~\ref{eq:image_formation}, which was previously evaluated for the baseline methods in Table~\ref{tab:generative_models_comparison}. We now observe that our proposed approach improves physical consistency, both through the inclusion of the reconstruction loss and the incorporation of shading-based conditioning. These improvements are demonstrated in Table~\ref{tab:image_reconstruction}.

\begin{table*}[tbp]
\centering
\resizebox{\textwidth}{!}{%
\begin{tabular}{lcccccccc}
\toprule
& \multicolumn{2}{c}{MIT}
& \multicolumn{2}{c}{IIW}
& \multicolumn{2}{c}{ARAP}
& \multicolumn{2}{c}{InteriorVerse} \\
\cmidrule(lr){2-3}
\cmidrule(lr){4-5}
\cmidrule(lr){6-7}
\cmidrule(lr){8-9}
Conditioning
& DSSIM$\downarrow$ & MSE$\downarrow$
& WHDR (10\%)$\downarrow$ & WHDR (20\%)$\downarrow$
& DSSIM$\downarrow$ & MSE$\downarrow$
& PSNR$\uparrow$ & LPIPS$\downarrow$ \\
\midrule
None & 0.0481 & 0.0097 & 24.41 & 20.72 & 0.1904 & 0.0670 & 15.50 & 0.31 \\
with LBM-NID & 0.0561 & 0.0147  & 33.44 & 30.69 & 0.2058 & 0.0844 & 11.64 & 0.47 \\
with Normal \cite{ke2025marigold} & 0.0554 & 0.0141 & 33.65 & 30.90 & 0.2033 & 0.0818 & 12.16 & 0.45 \\
with LBM-SID & 0.0472 & 0.0093 & 18.36 & 15.68 & 0.1536 & 0.0307 & 16.18 & 0.35 \\
\bottomrule
\end{tabular}}\caption{\textit{Table 3.} Ablation study of different conditioning inputs across MIT, IIW, ARAP, and InteriorVerse datasets.}
\label{tab:conditioning_ablation}
\end{table*}

\begin{table}[tbp]
\centering
\resizebox{\linewidth}{!}{%
\begin{tabular}{lccc}
\toprule
 &  \multicolumn{3}{c}{ARAP} \\
\cmidrule(lr){2-4}
 LBM Shading estimation 
& DSSIM$\downarrow$
& LMSE$\downarrow$
& MSE$\downarrow$ \\
\midrule
LBM-SID without Rec.loss & 0.1116 & 0.0102 & 0.0200 \\
LBM-SID  & 0.1383 & 0.0164 & 0.0262 \\
LBM-SID with Norm. cond. &  0.1391 & 0.0157 & 0.0256 \\
LBM-SID with Albedo cond. &  0.1106 & 0.0090 & 0.0179 \\
\bottomrule
\end{tabular}}
\caption{\textit{Table 4}. Evaluation of shading prediction on the ARAP dataset. \textit{Extended:} the last row conditions the shading estimator on the albedo predicted by LBM-AID (with reconstruction loss), rather than on ground-truth normals, and improves on every metric over the unconditioned model.}
\vspace{3mm}
\label{tab:shading_arap}
\end{table}

\begin{table}[tbp]
\centering
\resizebox{\linewidth}{!}{%
\begin{tabular}{lccc}
\toprule
& \multicolumn{3}{c}{MSE Reconstruction$\downarrow$}
\\
\cmidrule(lr){2-4}
Method & \multicolumn{1}{c}{MIT}
& \multicolumn{1}{c}{ARAP} 
& \multicolumn{1}{c}{Hypersim} \\
\midrule
SD-AID (10) & 0.0215 & 0.0261 & 0.1346 \\
SD-AID (50) & 0.0204 & 0.0216 & 0.1266 \\
FM-AID & 0.0129 & 0.0194  & 0.0341 \\ 
LBM-AID (wout Rec.Loss) & 0.0111 & 0.0207 & 0.0362 \\ 
\midrule
\multicolumn{4}{l}{LBM-AID (with Rec.Loss \& Conditionings) } \\ 
\hspace{0.2cm} No conditioning & 0.0111 & 0.0186 & 0.0305 \\
\hspace{0.2cm} LBM-NID & 0.0120 & 0.0266 & 0.0326\\
\hspace{0.2cm} Normal with \cite{ke2025marigold} & 0.0117 & 0.0230 & 0.0325 \\
\hspace{0.2cm} LBM-SID & 0.0111 & 0.0147 & 0.0280 \\
\hspace{0.2cm} LBM-SID with Albedo cond. & 0.0104 & 0.0139 & 0.0250 \\
\bottomrule
\end{tabular}}
\caption{\textit{Table 5.} Image reconstruction error for all the evaluated methods. Computed between input image and its estimated version $\hat{I} = \hat{A}\cdot S$ from the estimated albedo. Rows 1 to 4 reproduce some values from Table~\ref{tab:generative_models_comparison}. Rows 6 to 10 are for all versions of LBM-AID with reconstruction loss and different conditioning. \textit{Extended:} the last row uses the albedo-conditioned shading estimator introduced in Table~\ref{tab:shading_arap}.}
\label{tab:image_reconstruction}
\end{table}

\subsection{Comparison with state-of-the-art methods}
\label{sec:sota}
We compare our best model -- LBM-AID with the reconstruction loss, conditioned on the albedo-conditioned LBM-SID (Section~\ref{par:recons_loss}) -- against recent state-of-the-art IID methods on all five benchmarks. Quantitative figures for competing methods are taken from their original publications; where prior work does not report a given dataset or metric, the corresponding cell is omitted from the tables below.

\begin{figure}[tbp]
\centering
\setlength{\tabcolsep}{1pt}
\renewcommand{\arraystretch}{1}
\begin{tabular}{ccc}
\includegraphics[width=0.32\linewidth]{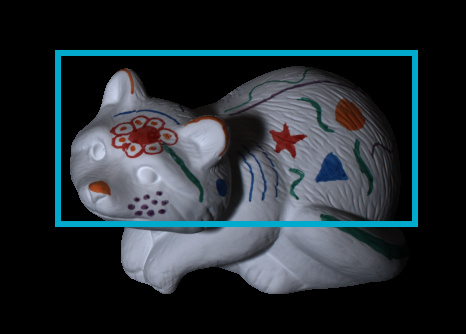} &
\includegraphics[width=0.32\linewidth]{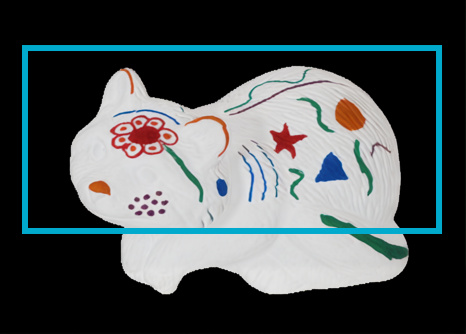} &
\includegraphics[width=0.32\linewidth]{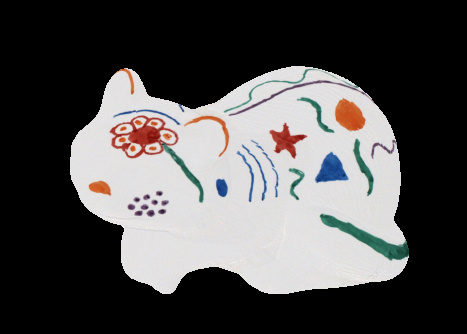} \\
\includegraphics[width=0.32\linewidth]{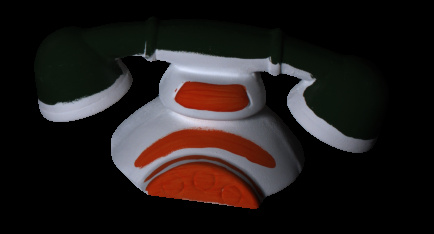} &
\includegraphics[width=0.32\linewidth]{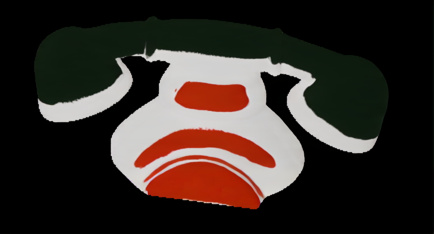} &
\includegraphics[width=0.32\linewidth]{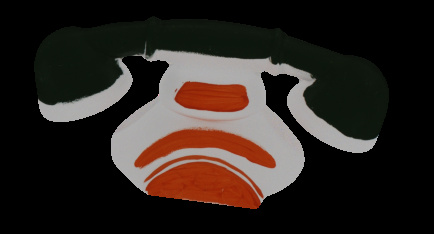} \\
\includegraphics[width=0.32\linewidth]{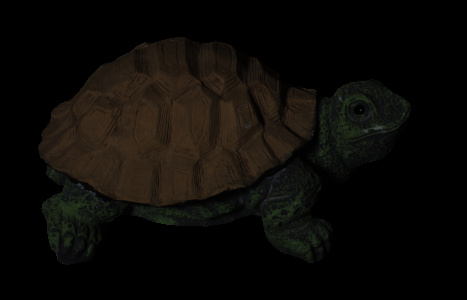} &
\includegraphics[width=0.32\linewidth]{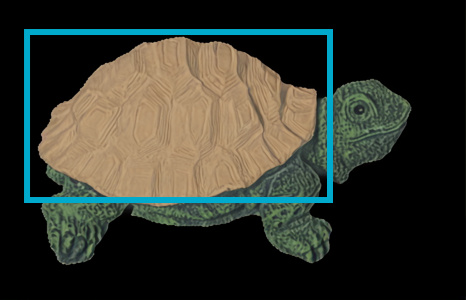} &
\includegraphics[width=0.32\linewidth]{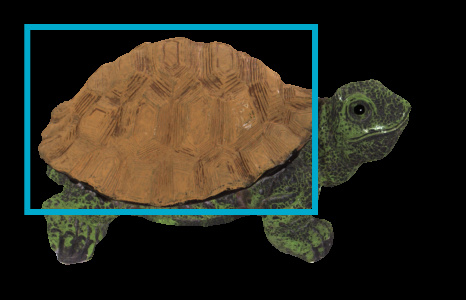} \\
\includegraphics[width=0.32\linewidth]{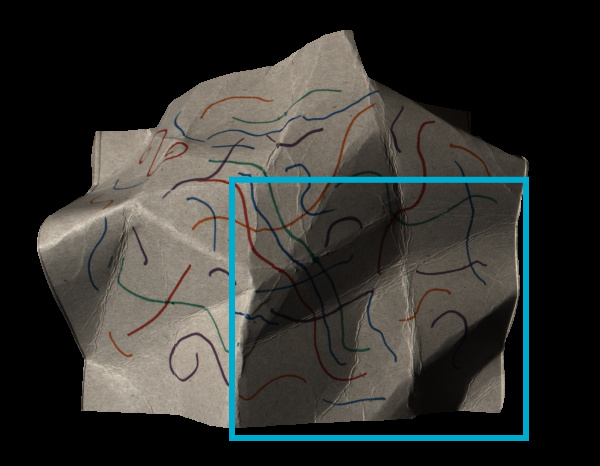} &
\includegraphics[width=0.32\linewidth]{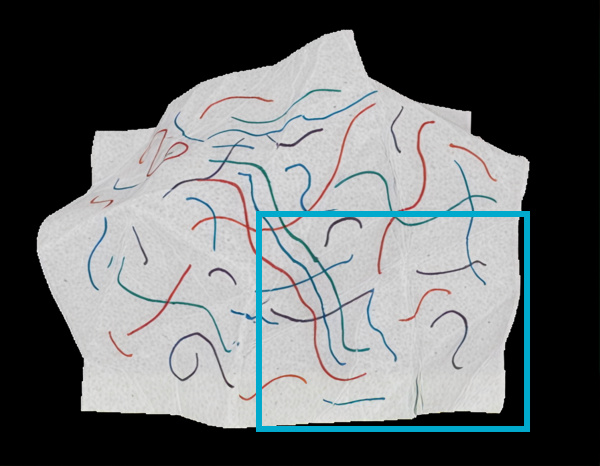} &
\includegraphics[width=0.32\linewidth]{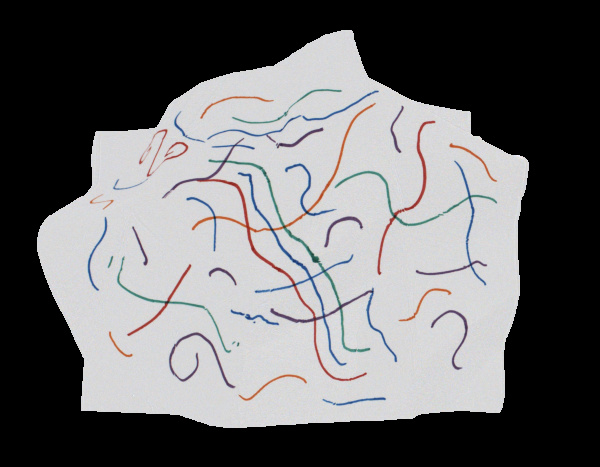} \\
\includegraphics[width=0.32\linewidth]{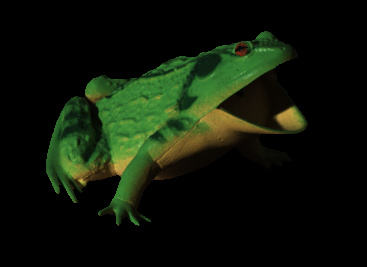} &
\includegraphics[width=0.32\linewidth]{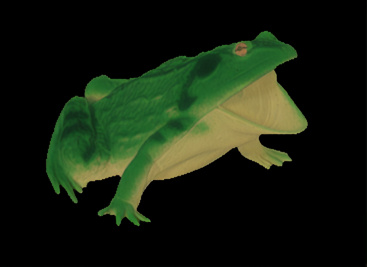} &
\includegraphics[width=0.32\linewidth]{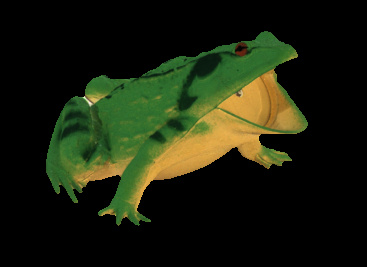} \\
\includegraphics[width=0.32\linewidth]{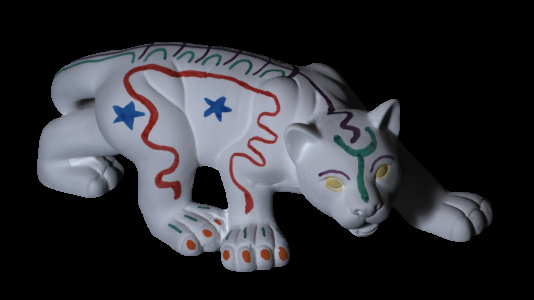} &
\includegraphics[width=0.32\linewidth]{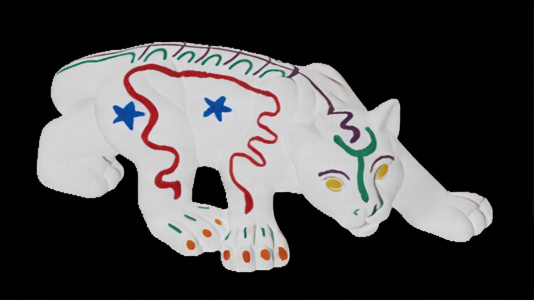} &
\includegraphics[width=0.32\linewidth]{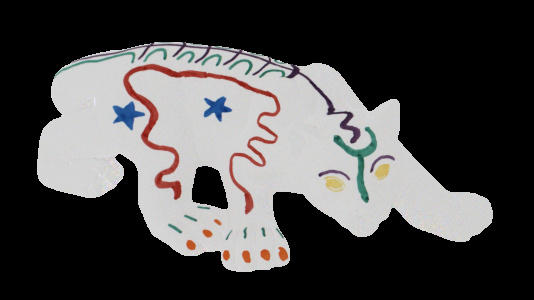} \\
\vspace{-1mm}
\footnotesize Input RGB & \footnotesize Ours & \footnotesize GT Reflectance \\
\end{tabular}
\caption{Qualitative comparison of our method against the ground-truth reflectance on the MIT dataset. Cyan rectangles highlight shadow suppression (rows 1 and 4) and color misalignment (row 3).}
\vspace{-1mm}
\label{fig:sota_mit}
\end{figure}

On MIT (Table~\ref{tab:sota_mit}), our method obtains the best LMSE among the compared methods, though at a higher DSSIM and MSE than PIE-Net~\cite{Das2022PIENet} and FlowIID~\cite{singla2026flowiid}. Qualitatively (Figure~\ref{fig:sota_mit}), our predictions closely follow the ground truth, correctly suppressing cast shadows and specular highlights, though some color misalignment remains on a small number of scenes.

\begin{table}[tbp]
\centering
\begin{tabular}{lccc}
\toprule
Method & DSSIM$\downarrow$ & LMSE$\downarrow$ & MSE$\downarrow$ \\
\midrule
PIE-Net~\cite{Das2022PIENet} & 0.0340 & 0.0136 & 0.0028 \\
FlowIID~\cite{singla2026flowiid} & 0.0435 & 0.0043 & 0.0040 \\
Ours & 0.0443 & 0.0030 & 0.0076 \\
\bottomrule
\end{tabular}
\caption{\textit{Table 6.} Comparison with SOTA IID methods on MIT.}
\label{tab:sota_mit}
\end{table}

\begin{figure*}[tbp]
\centering
\setlength{\tabcolsep}{1pt}
\renewcommand{\arraystretch}{1}
\begin{tabular}{ccccc}
\footnotesize Input RGB & \footnotesize Zhu et al. & \footnotesize IntrinsicDiffusion & \footnotesize Ours & \footnotesize GT Reflectance \\
\includegraphics[width=0.17\linewidth]{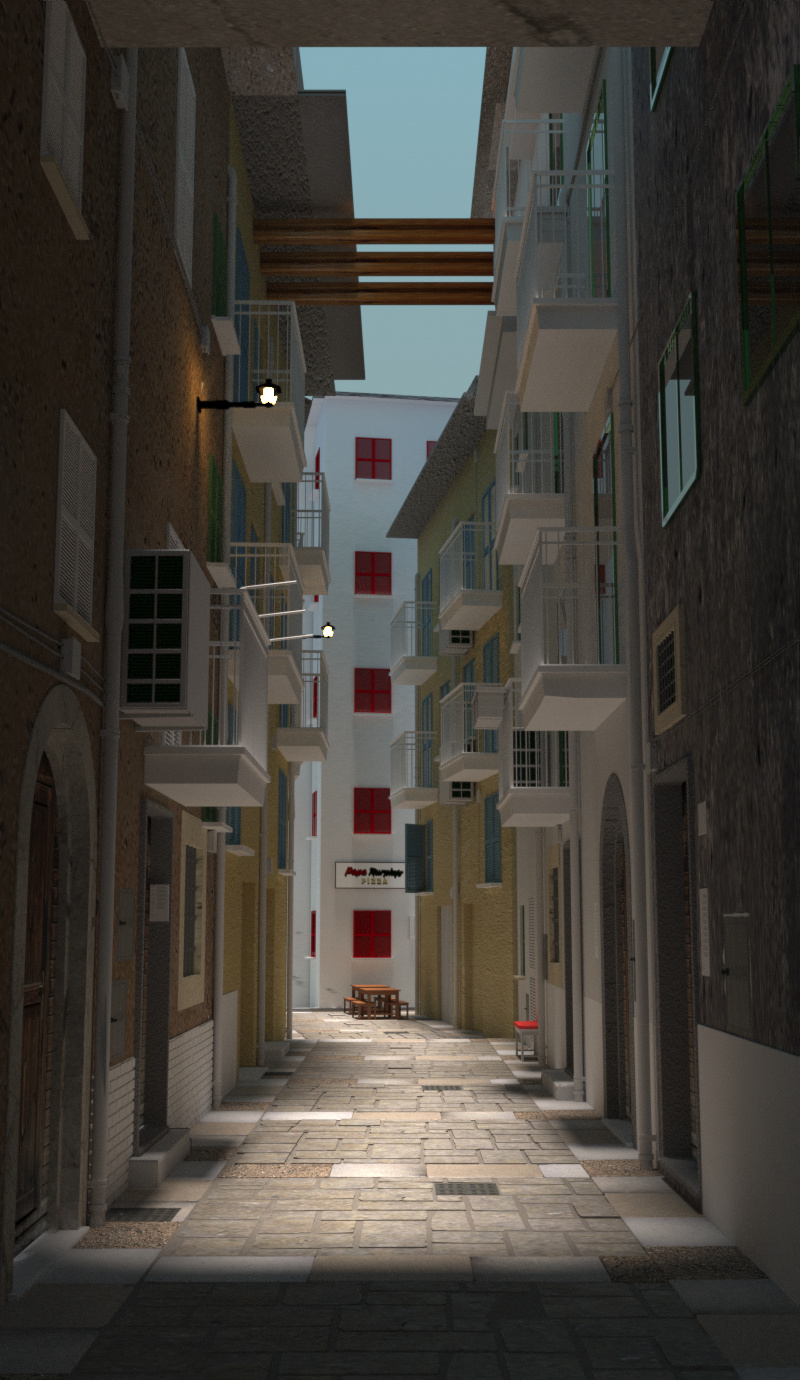} &
\includegraphics[width=0.17\linewidth]{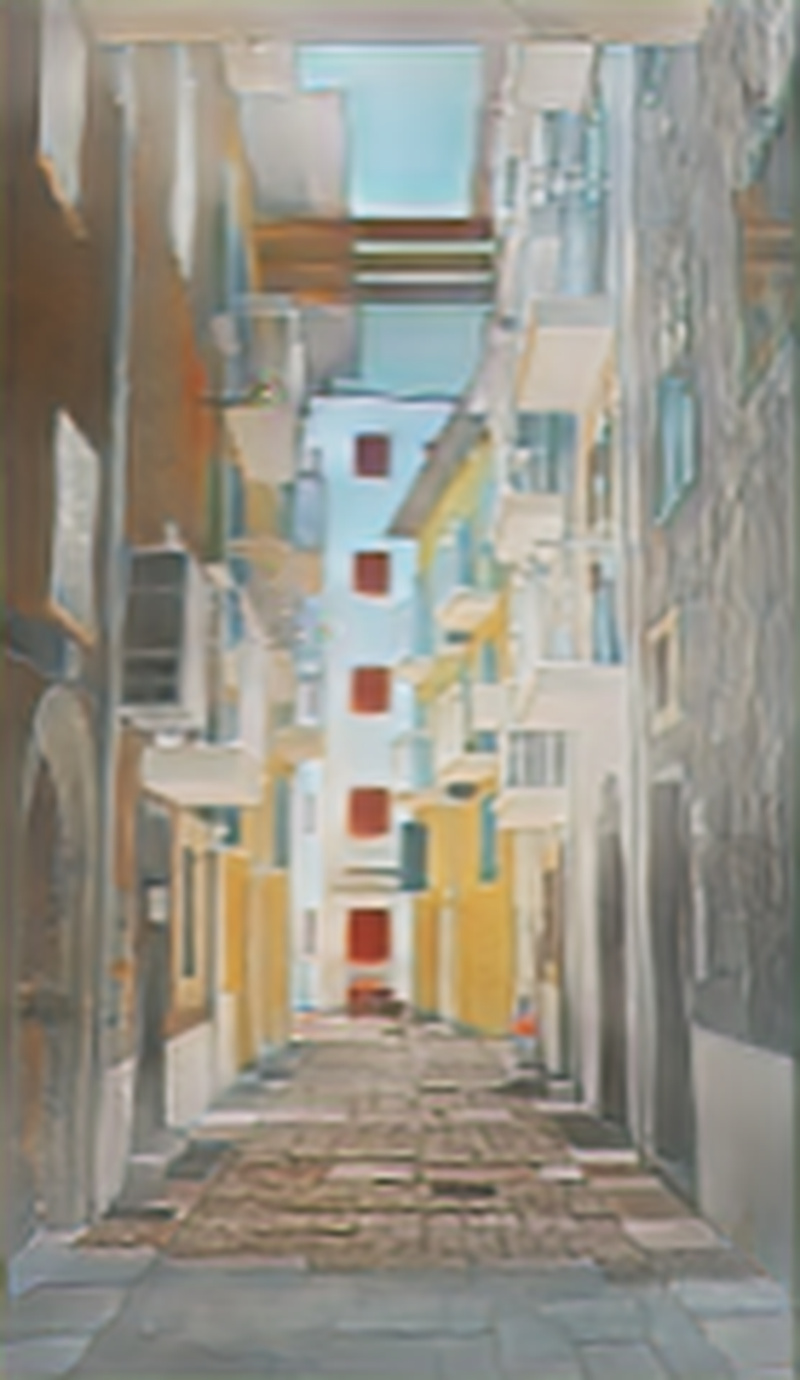} &
\includegraphics[width=0.17\linewidth]{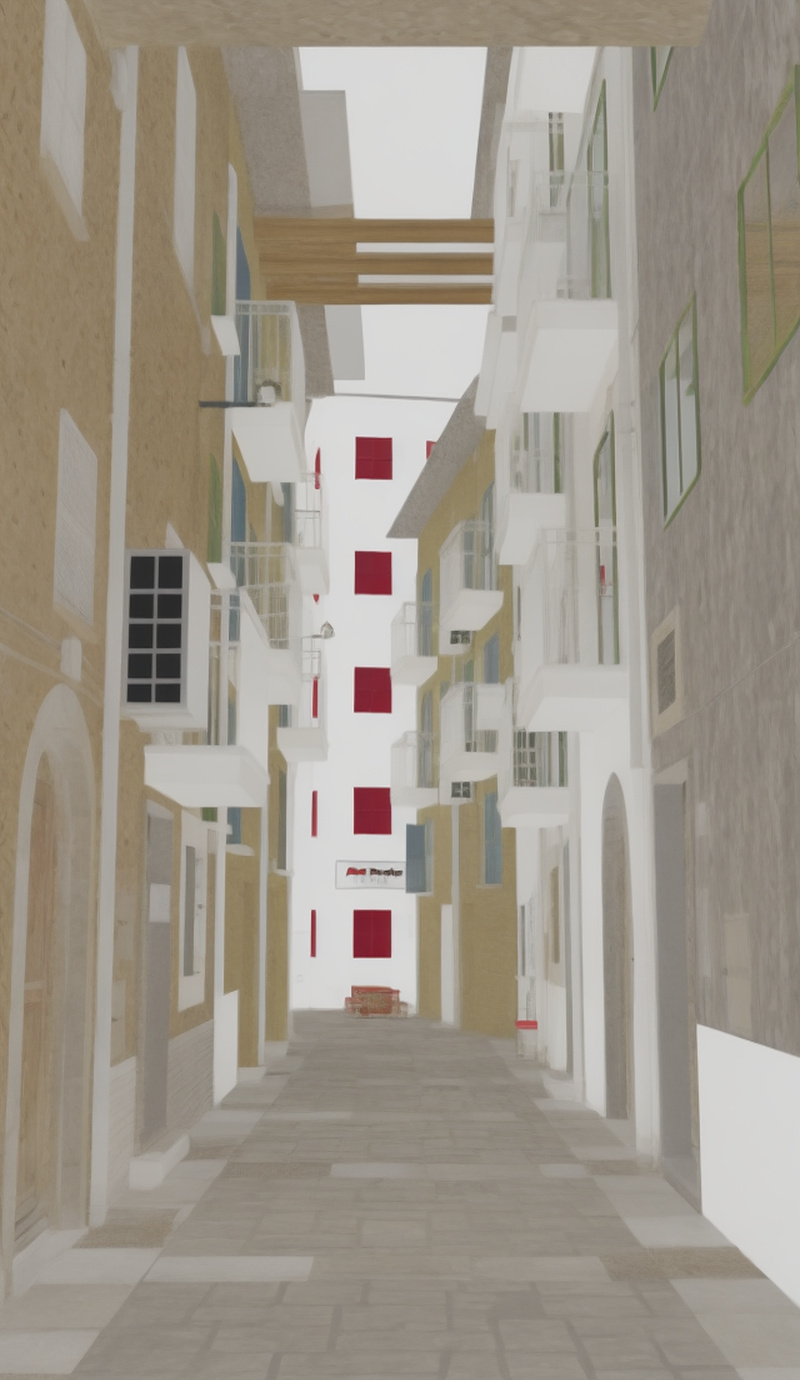} &
\includegraphics[width=0.17\linewidth]{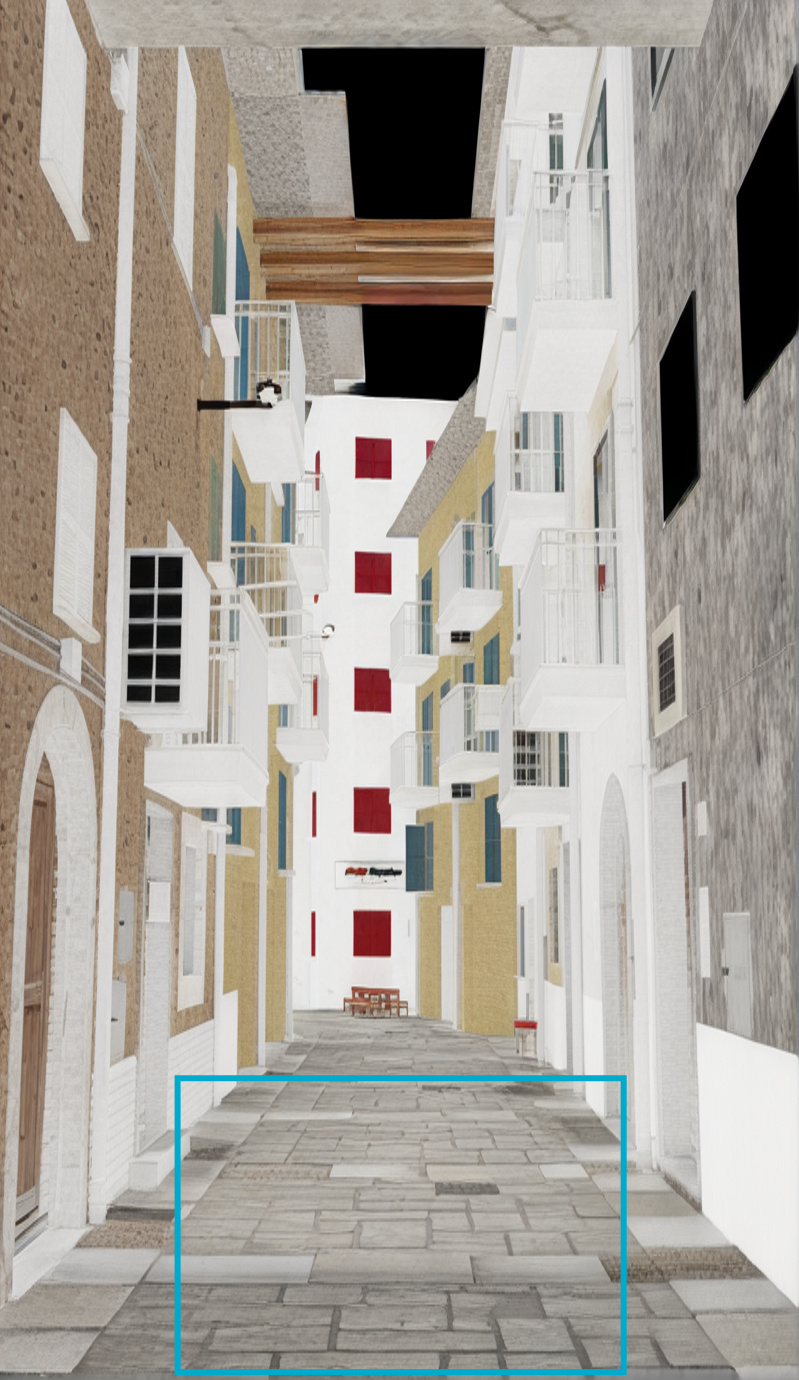} &
\includegraphics[width=0.17\linewidth]{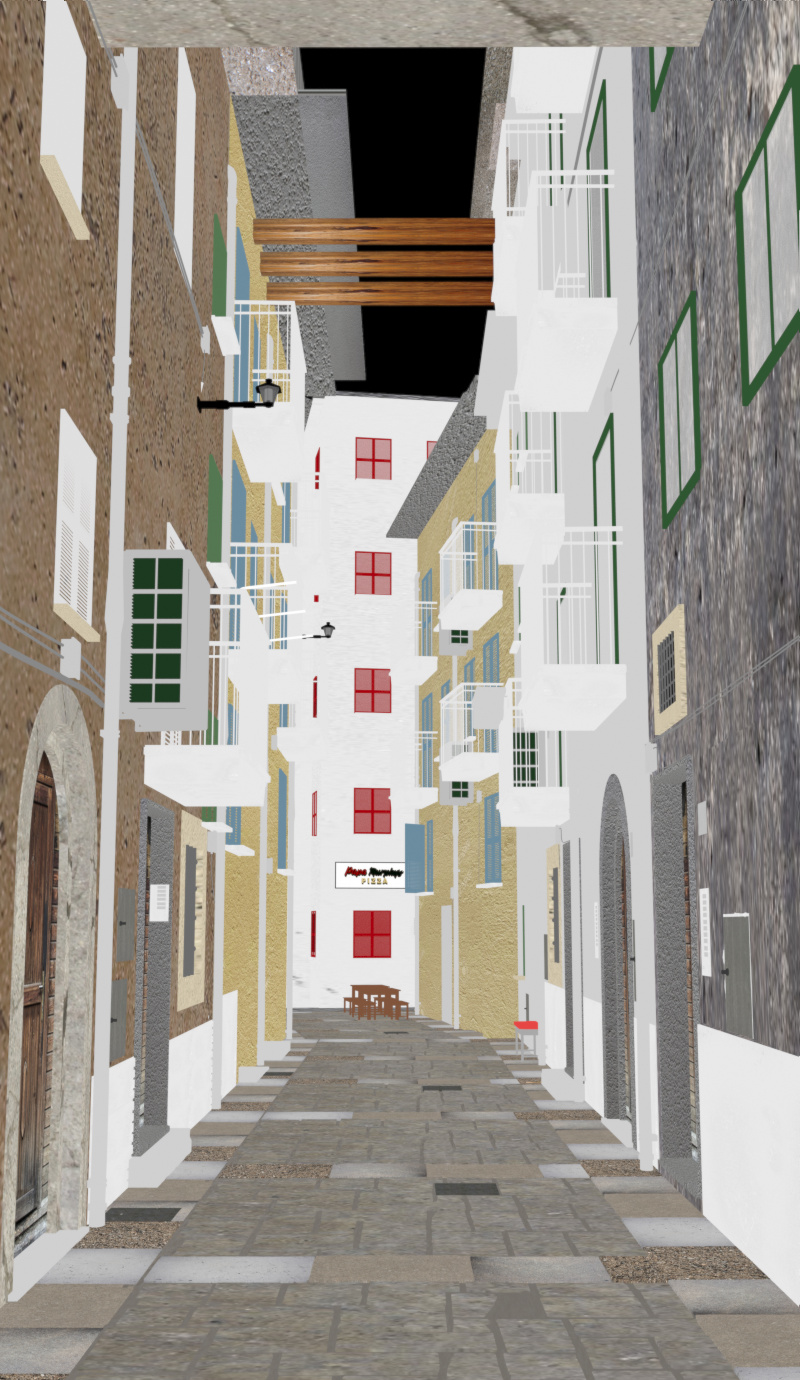} \\
\includegraphics[width=0.17\linewidth]{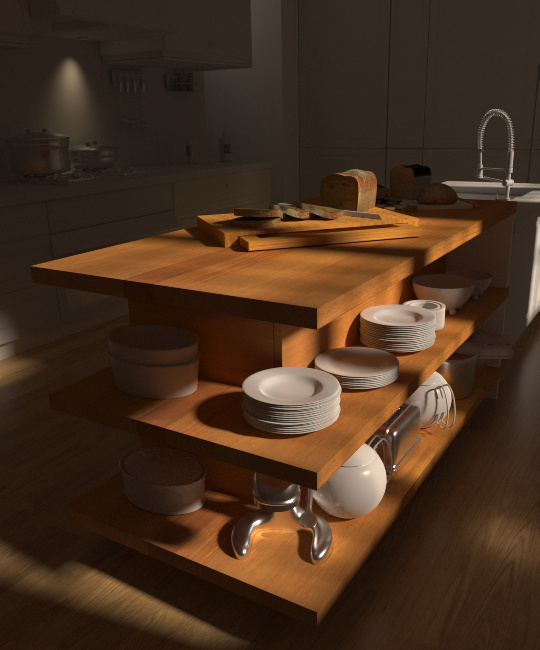} &
\includegraphics[width=0.17\linewidth]{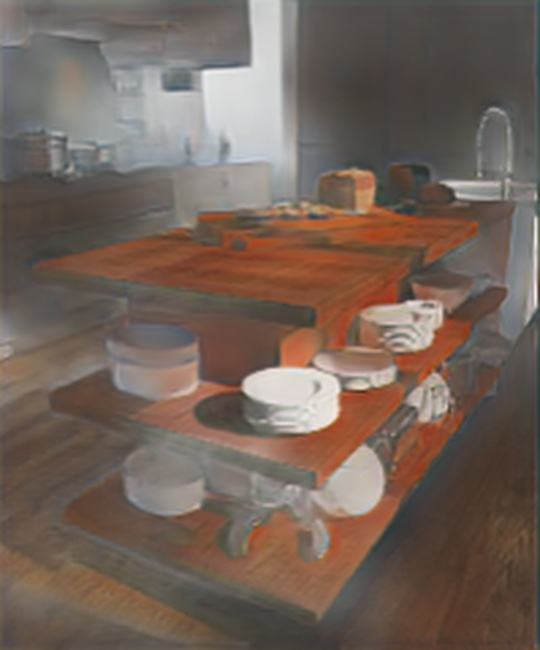} &
\includegraphics[width=0.17\linewidth]{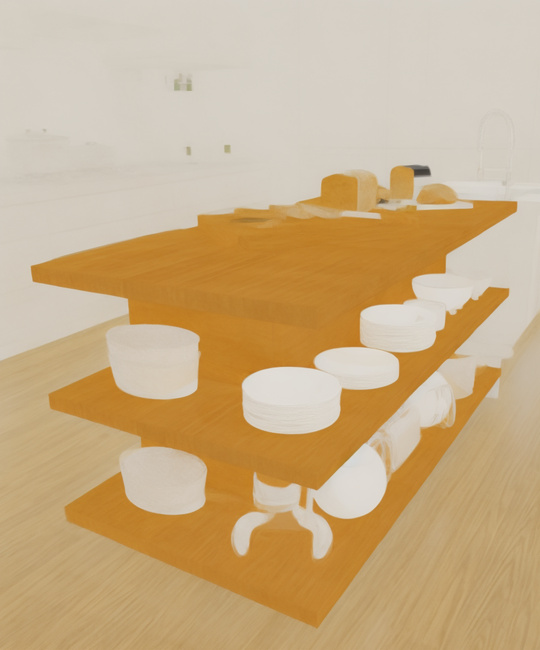} &
\includegraphics[width=0.17\linewidth]{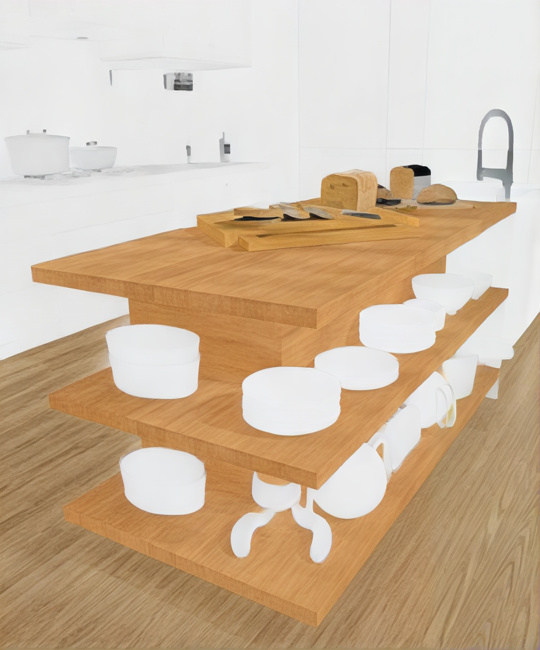} &
\includegraphics[width=0.17\linewidth]{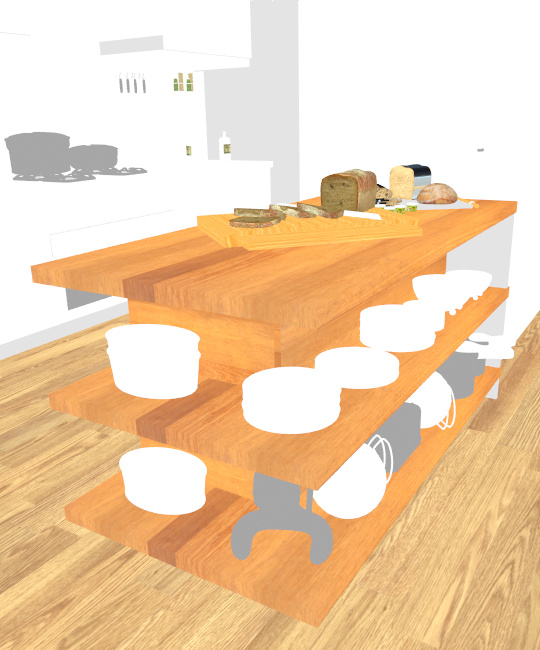} \\
\includegraphics[width=0.17\linewidth]{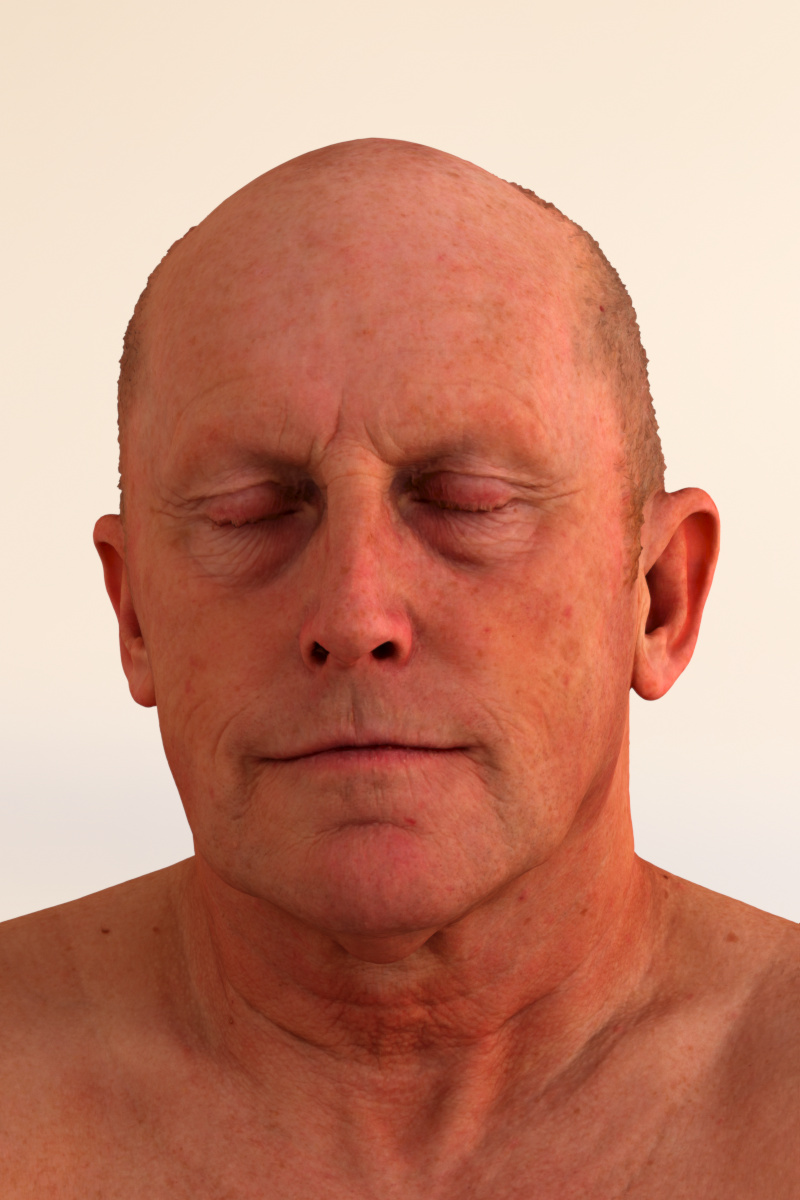} &
\includegraphics[width=0.17\linewidth]{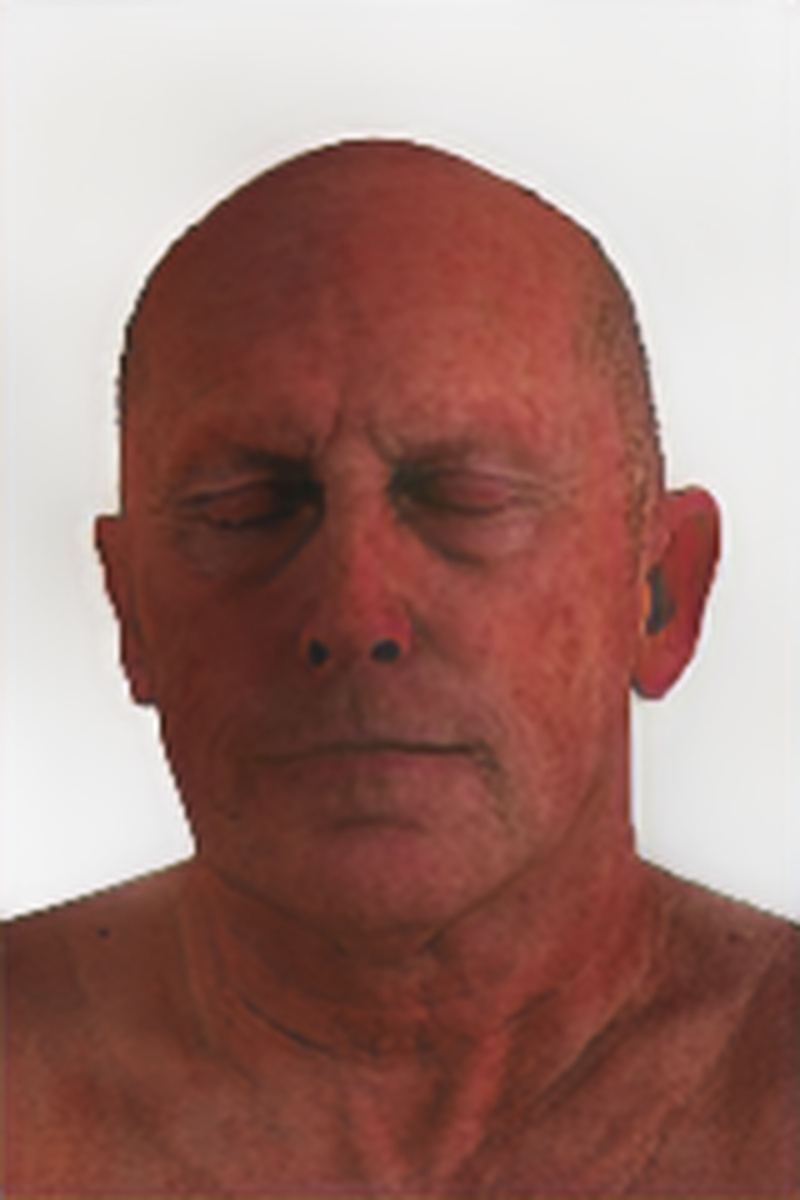} &
\includegraphics[width=0.17\linewidth]{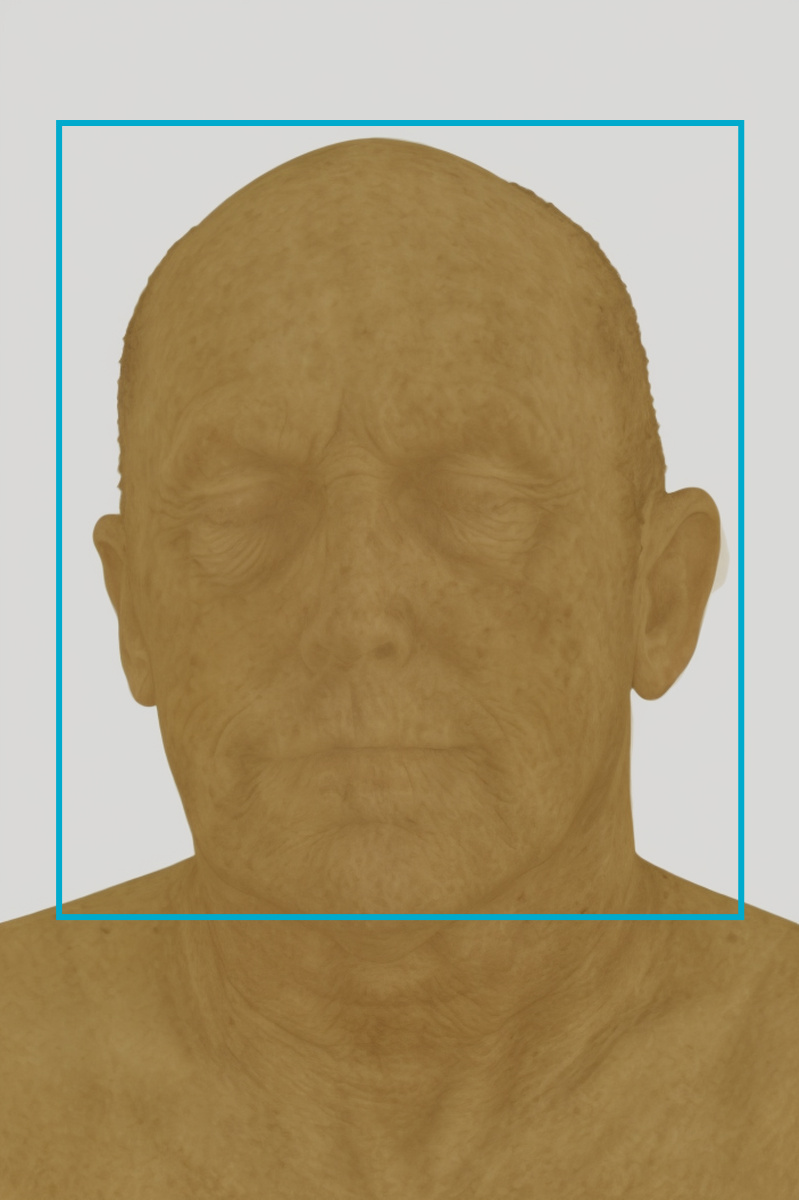} &
\includegraphics[width=0.17\linewidth]{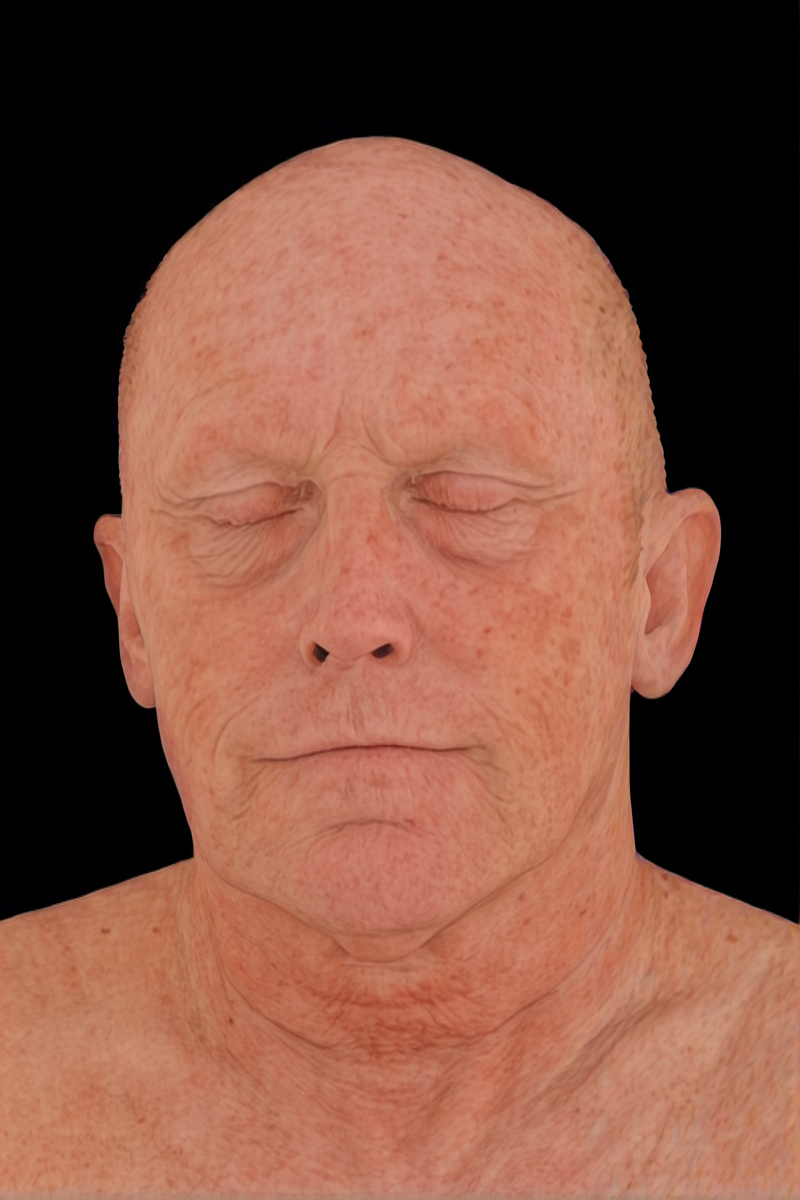} &
\includegraphics[width=0.17\linewidth]{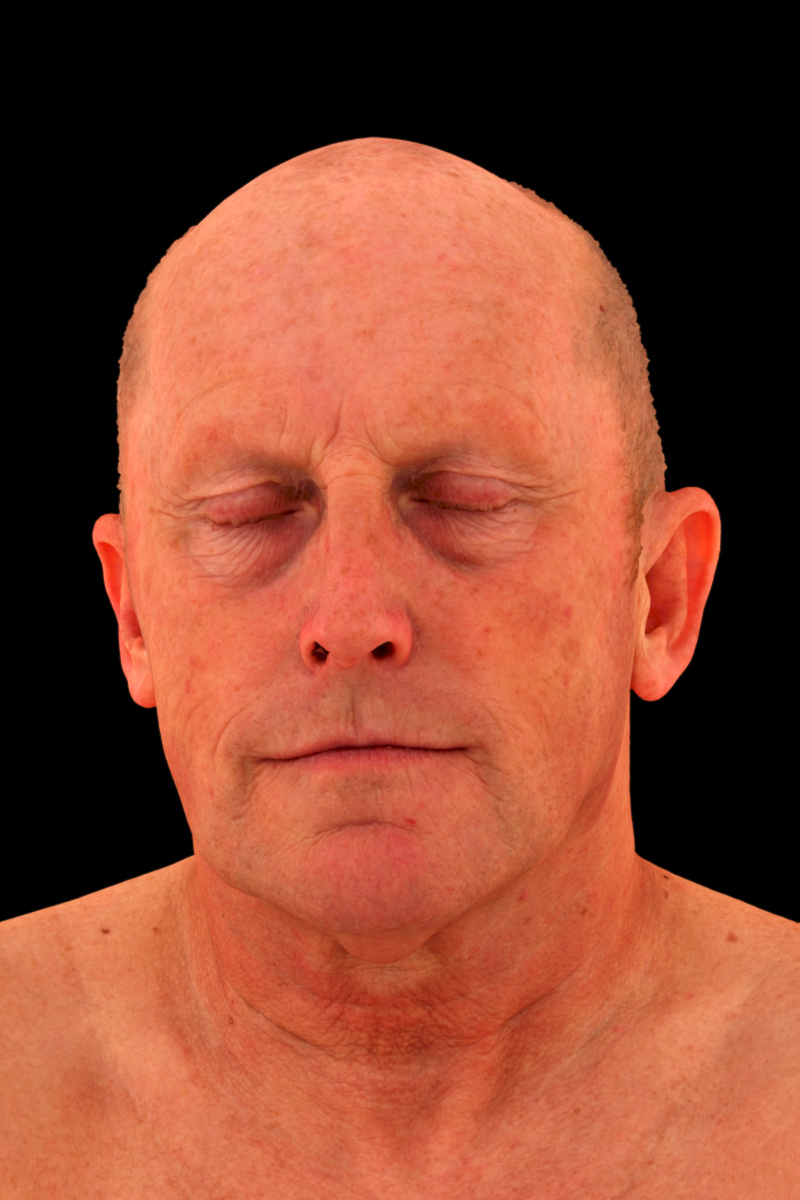} \\
\includegraphics[width=0.17\linewidth]{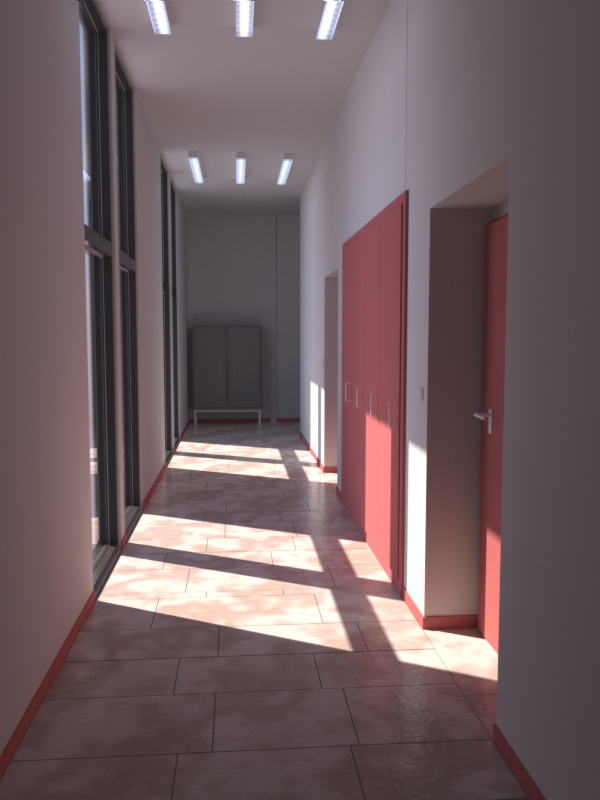} &
\includegraphics[width=0.17\linewidth]{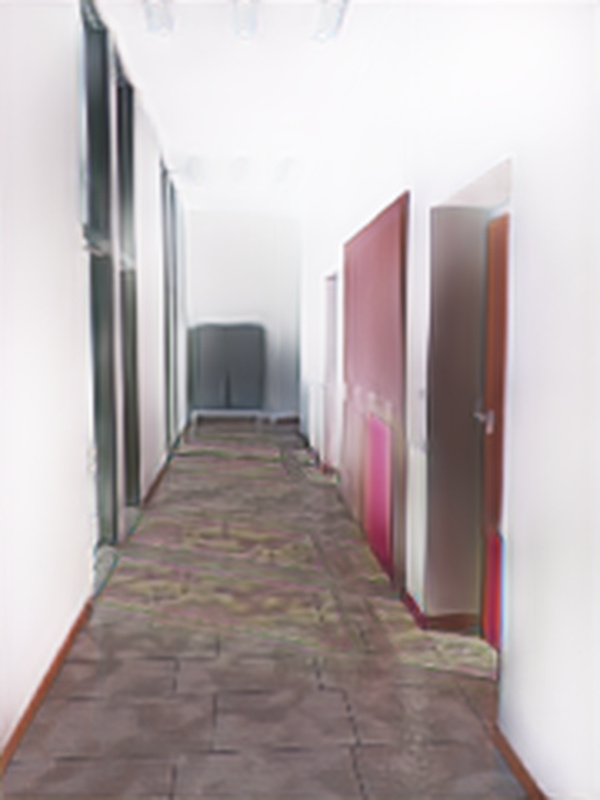} &
\includegraphics[width=0.17\linewidth]{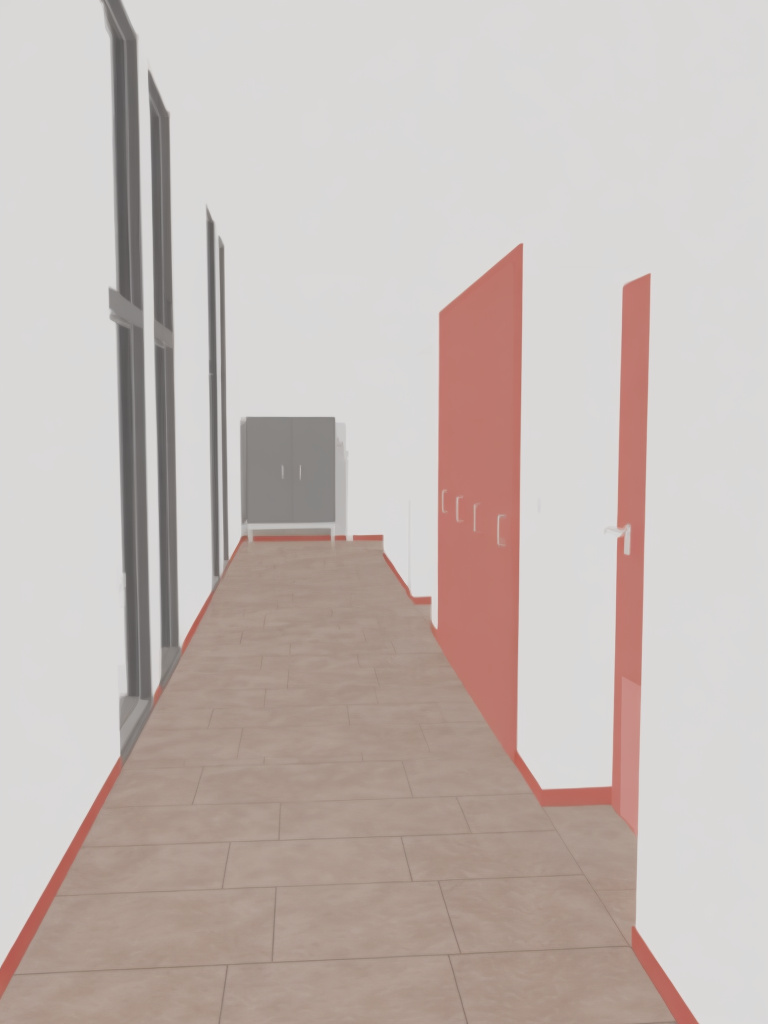} &
\includegraphics[width=0.17\linewidth]{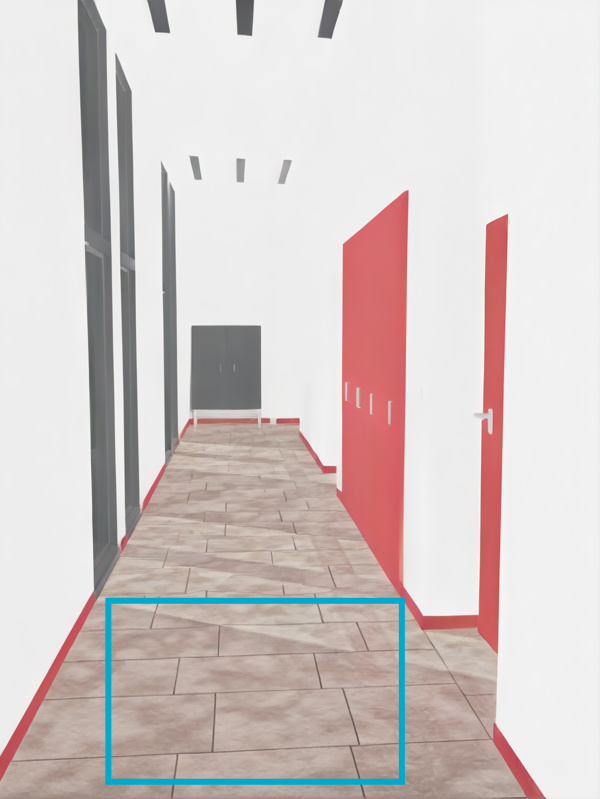} &
\includegraphics[width=0.17\linewidth]{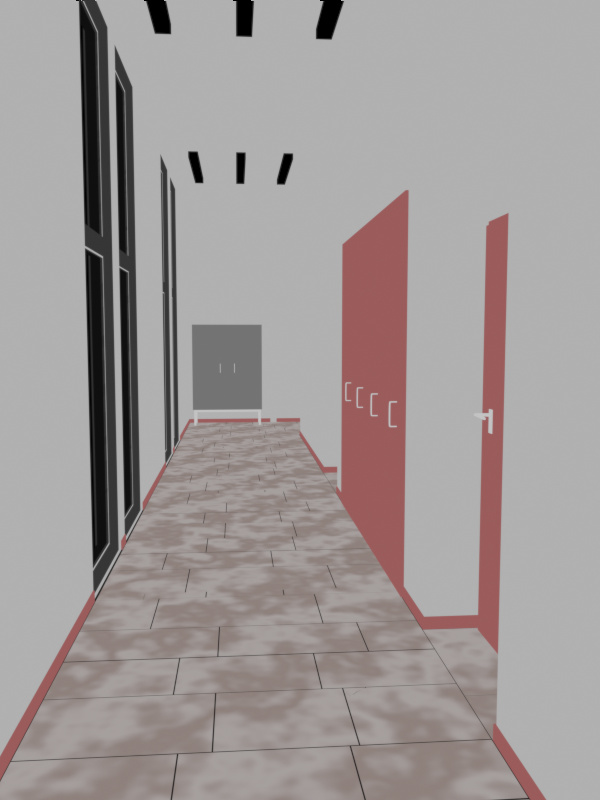} \\
\includegraphics[width=0.17\linewidth]{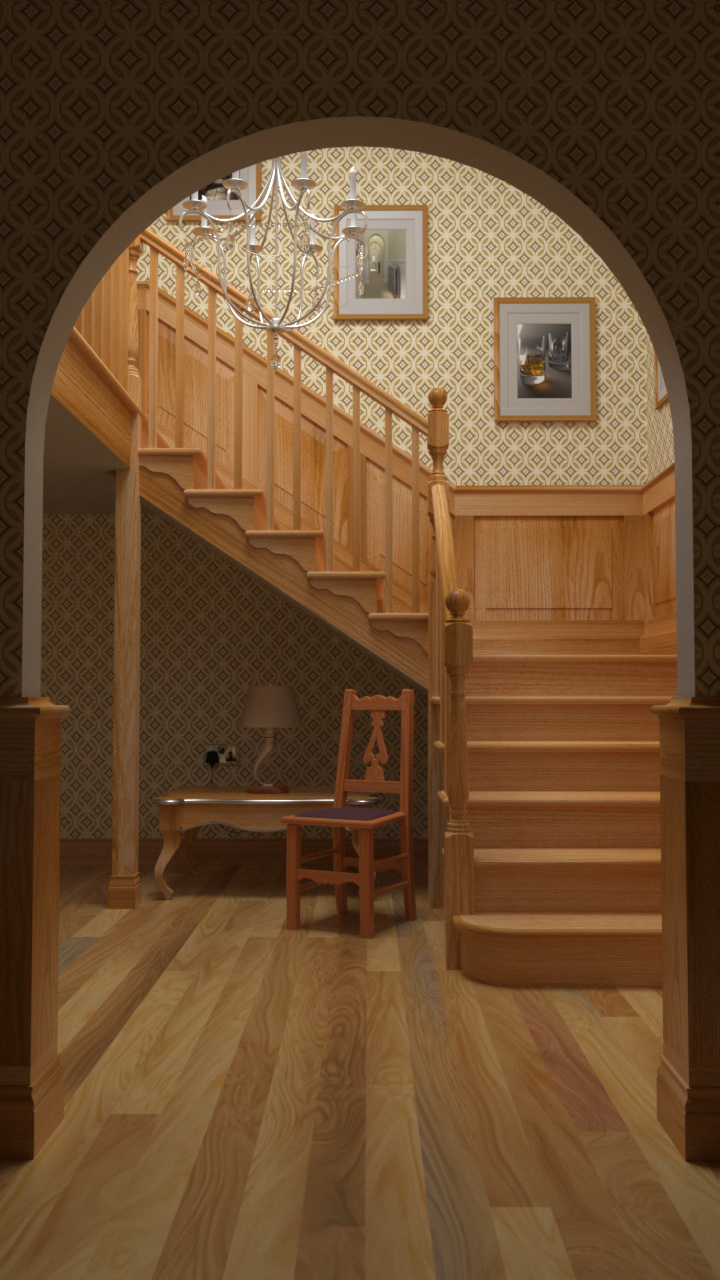} &
\includegraphics[width=0.17\linewidth]{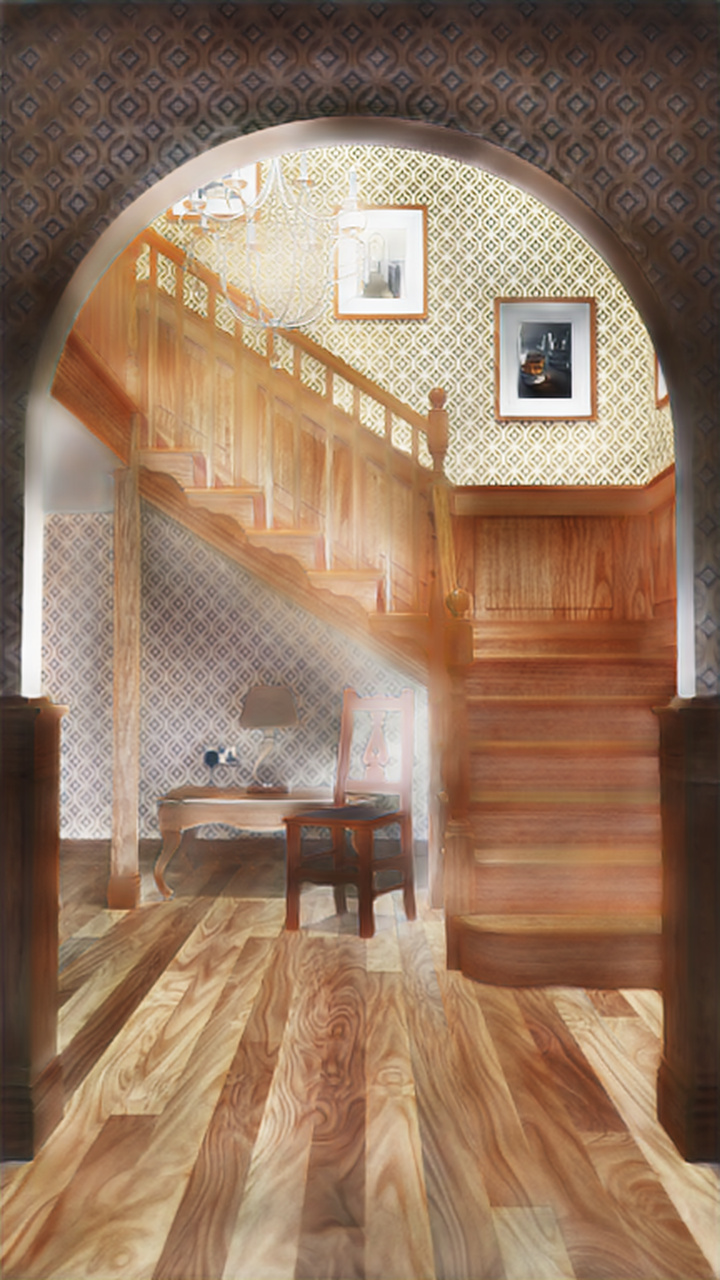} &
\includegraphics[width=0.17\linewidth]{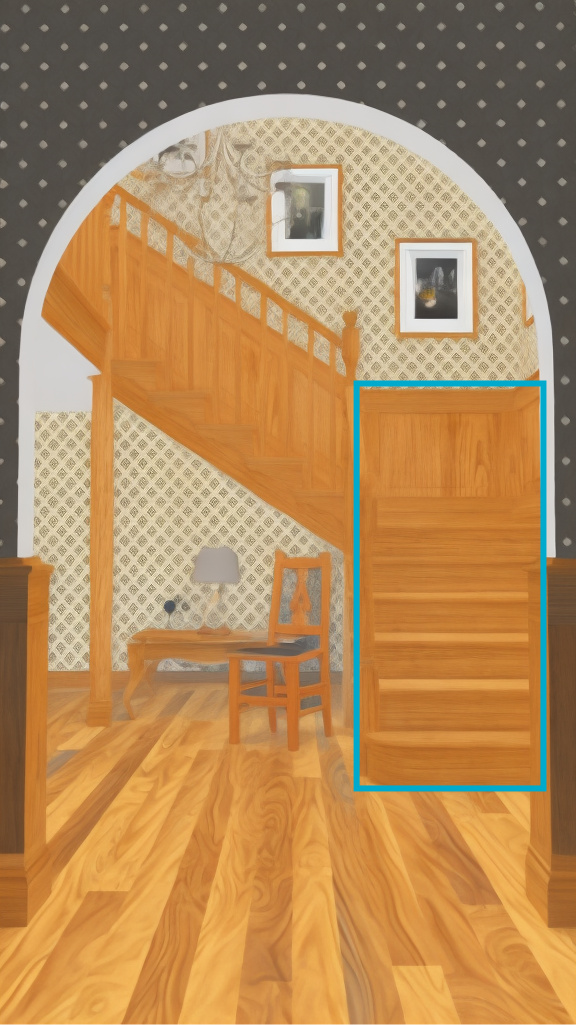} &
\includegraphics[width=0.17\linewidth]{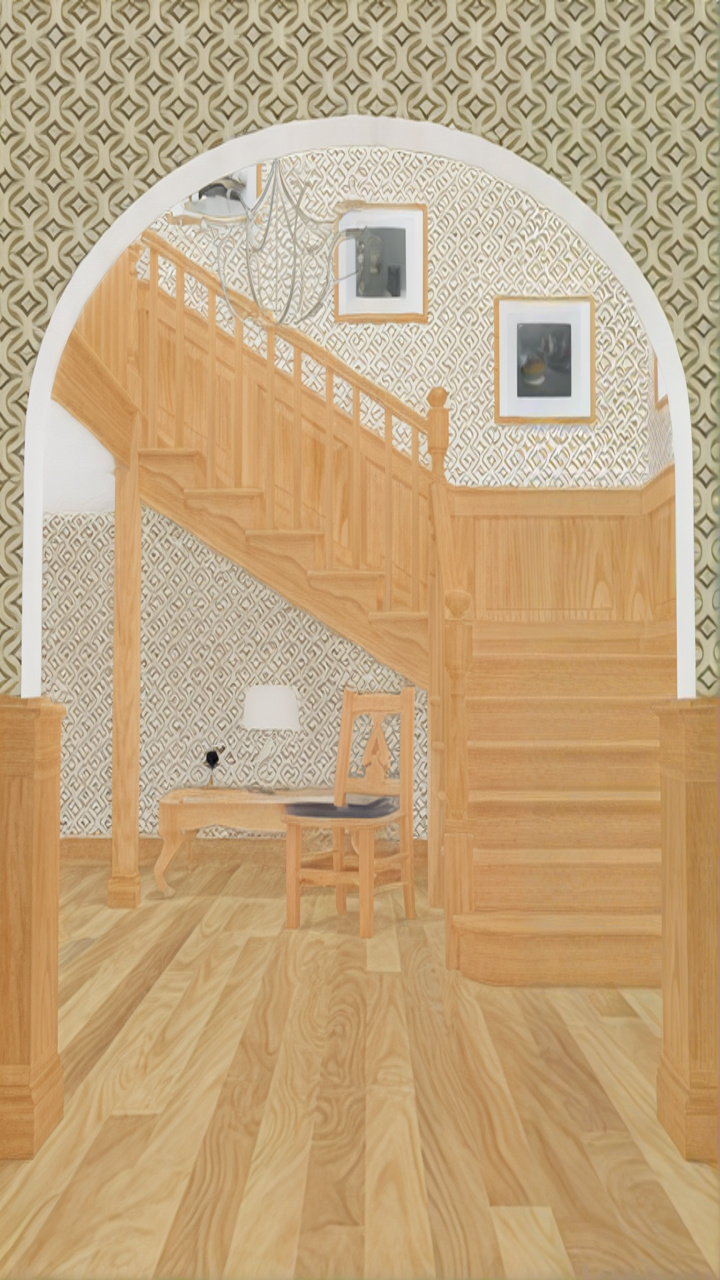} &
\includegraphics[width=0.17\linewidth]{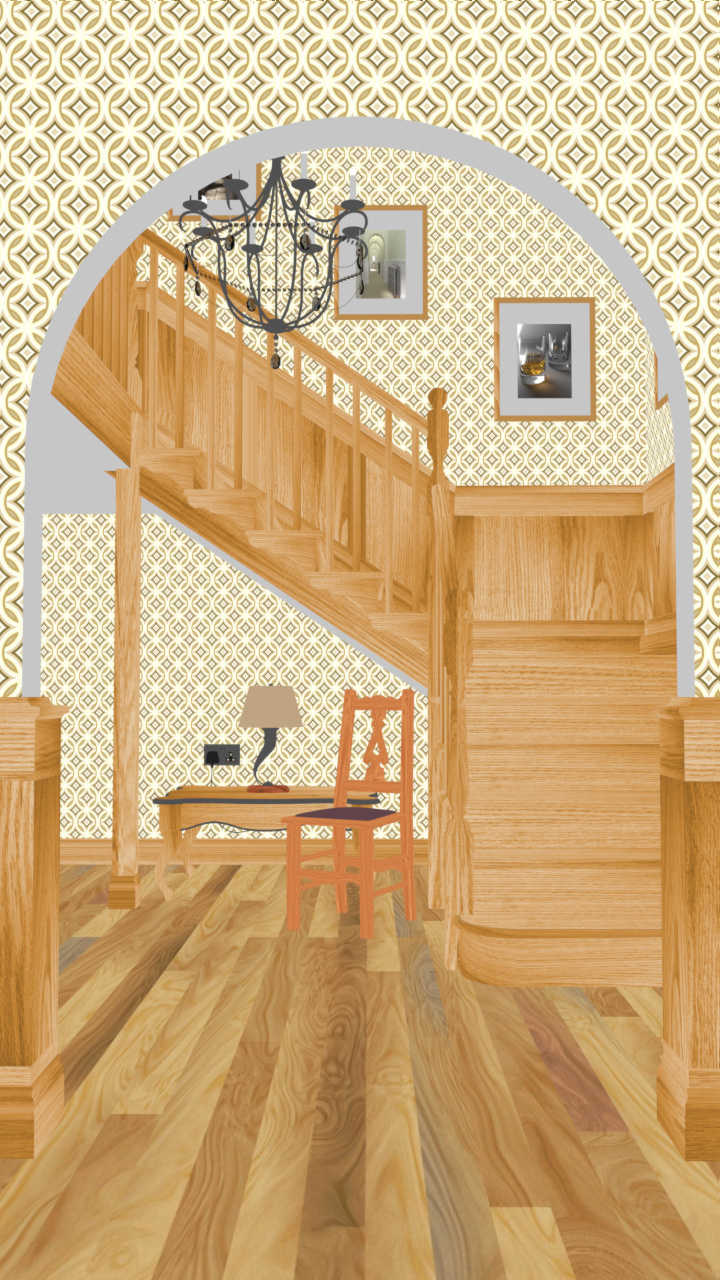} \\
\end{tabular}
\caption{Qualitative results on the ARAP dataset. Cyan rectangles highlight the orange/yellowish color shift introduced by IntrinsicDiffusion on the face and wooden surfaces, and the texture preservation of our method on the floor.}
\label{fig:sota_arap}
\end{figure*}

On ARAP (Table~\ref{tab:sota_arap}), our method surpasses the deep-learning baselines but trails the diffusion-based IntrinsicDiffusion~\cite{luo2024intrinsicdiffusion}. As Figure~\ref{fig:sota_arap} shows, IntrinsicDiffusion introduces a consistent orange/yellowish color shift on wooden surfaces and faces that is not reflected in the aggregate metrics, whereas our predictions preserve fine surface texture well but tend to underestimate overall brightness.

\begin{table}[tbp]
\centering
\begin{tabular}{lccc}
\toprule
Method & DSSIM$\downarrow$ & LMSE$\downarrow$ & MSE$\downarrow$ \\
\midrule
PIE-Net~\cite{Das2022PIENet} & 0.1598 & 0.0280 & 0.0428 \\
Careaga et al.~\cite{careaga2023ordinal} & 0.1555 & 0.0239 & 0.0366 \\
Zhu et al.~\cite{zhu2022learning} & 0.1363 & 0.0165 & 0.0276 \\
IntrinsicDiffusion~\cite{luo2024intrinsicdiffusion} & 0.1399 & 0.0141 & 0.0235 \\
Ours & 0.1460 & 0.0204 & 0.0266 \\
\bottomrule
\end{tabular}
\caption{\textit{Table 7.} Comparison with SOTA IID methods on ARAP.}
\label{tab:sota_arap}
\end{table}

Table~\ref{tab:sota_combined} reports the remaining three benchmarks together, since they share the same set of compared methods. For InteriorVerse and Hypersim we additionally report recent reinforcement-learning-guided variants (Marigold-X, PRISM-X), which fine-tune existing generative decomposition models with perceptual feedback~\cite{dirik2026reasonx}. Our single-step model is competitive with, but does not match, these methods or DNF-Intrinsic~\cite{zheng2025dnf}, which achieves the strongest results on InteriorVerse; our method still clearly outperforms the earlier deep-learning baselines and Kocsis et al.~\cite{kocsis2024intrinsic} on PSNR. On IIW our method outperforms all deep-learning and most diffusion-based baselines, but again falls short of the RL-guided variants.

\begin{table}[tbp]
\centering
\setlength{\tabcolsep}{2.2pt}
\begin{tabular}{lcccccc}
\toprule
& \multicolumn{2}{c}{InteriorVerse}
& \multicolumn{2}{c}{Hypersim}
& \multicolumn{2}{c}{IIW (WHDR)} \\
\cmidrule(lr){2-3}
\cmidrule(lr){4-5}
\cmidrule(lr){6-7}
Method
& PSNR & LPIPS
& PSNR & LPIPS
& 10\% & 20\% \\
\midrule
Zhu et al.~\cite{zhu2022learning}                  & 13.6 & 0.24 & 11.7 & 0.54 & 34.7 & 24.1 \\
PIE-Net~\cite{Das2022PIENet}                       & --   & --   & --   & --   & 33.3 & 23.5 \\
Careaga et al.~\cite{careaga2023ordinal}           & 17.4 & 0.20 & 13.5 & 0.34 & 24.8 & 19.2 \\
Kocsis et al.~\cite{kocsis2024intrinsic}           & 12.2 & 0.30 & 12.1 & 0.41 & 26.1 & 20.7 \\
RGB$\leftrightarrow$X~\cite{zeng2024rgbx}          & 16.6 & 0.17 & 17.4 & 0.16 & 23.6 & 21.1 \\
IntrinsicDiff.~\cite{luo2024intrinsicdiffusion}& --   & --   & --   & --   & 17.9 & 13.3 \\
Marigold~\cite{ke2025marigold}                     & 19.5 & 0.19 & 18.2 & 0.22 & 16.7 & 14.8 \\
PRISM~\cite{dirik2026prism}                        & 19.9 & 0.14 & 19.3 & 0.18 & 17.2 & 15.9 \\
DNF-Intrinsic~\cite{zheng2025dnf}         & 21.9 & 0.12 & --   & --   & --   & --   \\
Marigold-X~\cite{dirik2026reasonx}                 & 19.8 & 0.16 & 19.0 & 0.20 & 15.2 & 14.0 \\
PRISM-X~\cite{dirik2026reasonx}                    & 20.7 & 0.12 & 19.9 & 0.17 & 12.9 & 11.9 \\
\midrule
Ours                                               & 18.3 & 0.31 & 16.7 & 0.26 & 17.9 & 15.1 \\
\bottomrule
\end{tabular}
\caption{\textit{Table 8.} Comparison with SOTA IID methods on the InteriorVerse, Hypersim, and IIW test sets, grouped since they share the same set of compared methods. IIW is evaluated with WHDR at the $10\%$ and $20\%$ thresholds. Dashes (--) denote results not reported in the corresponding original publication.}
\label{tab:sota_combined}
\end{table}

\begin{figure*}[tbp]
\centering
\setlength{\tabcolsep}{1pt}
\renewcommand{\arraystretch}{1}
\begin{tabular}{cccccc}
\footnotesize Input RGB & \footnotesize Kocsis et al. & \footnotesize RGB$\leftrightarrow$X & \footnotesize DNF-Intrinsic & \footnotesize Ours & \footnotesize GT Reflectance \\
\includegraphics[width=0.15\linewidth]{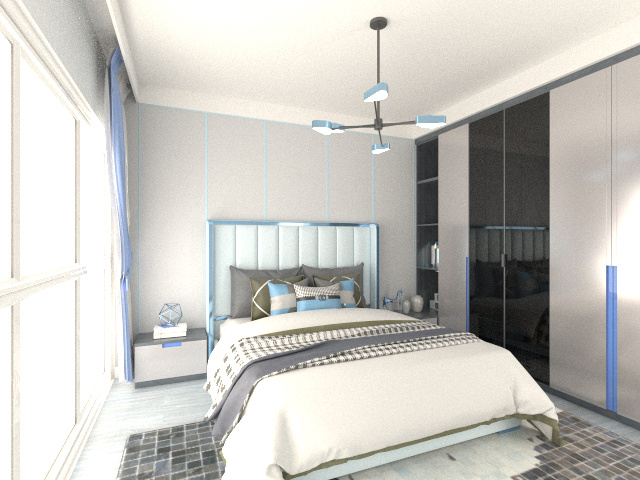} &
\includegraphics[width=0.15\linewidth]{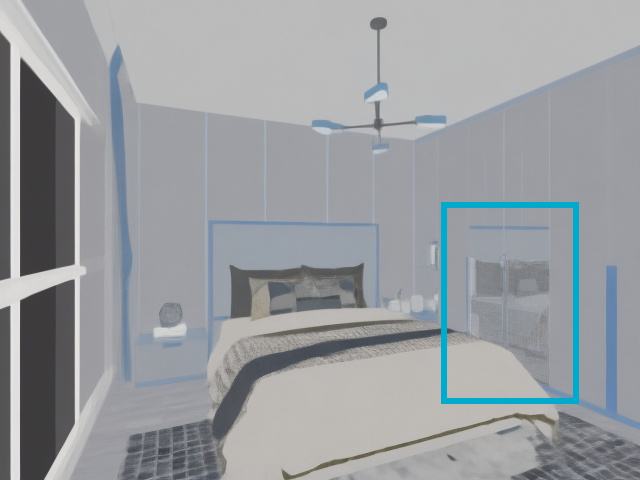} &
\includegraphics[width=0.15\linewidth]{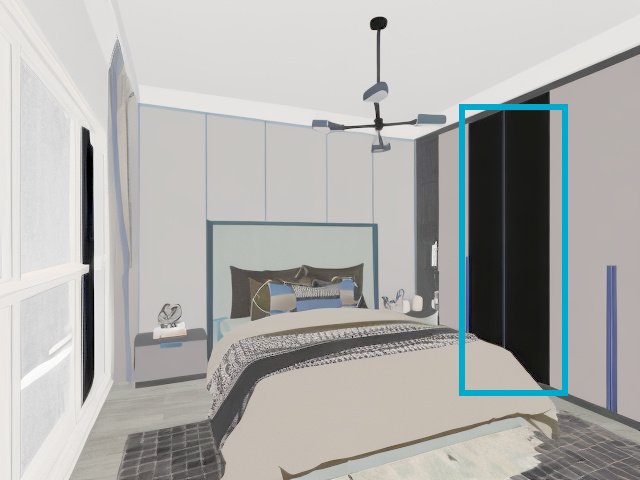} &
\includegraphics[width=0.15\linewidth]{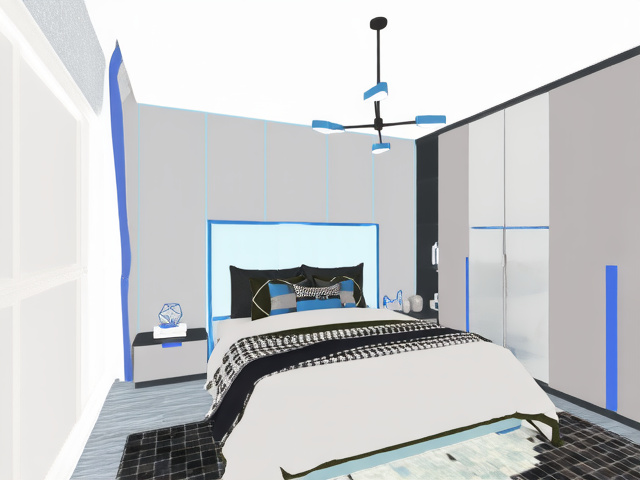} &
\includegraphics[width=0.15\linewidth]{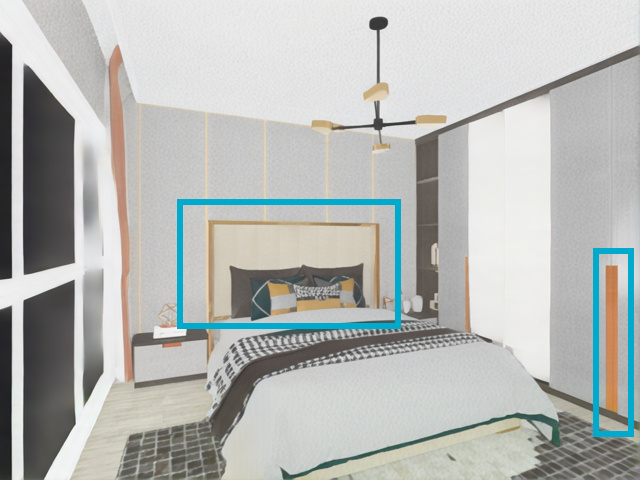} &
\includegraphics[width=0.15\linewidth]{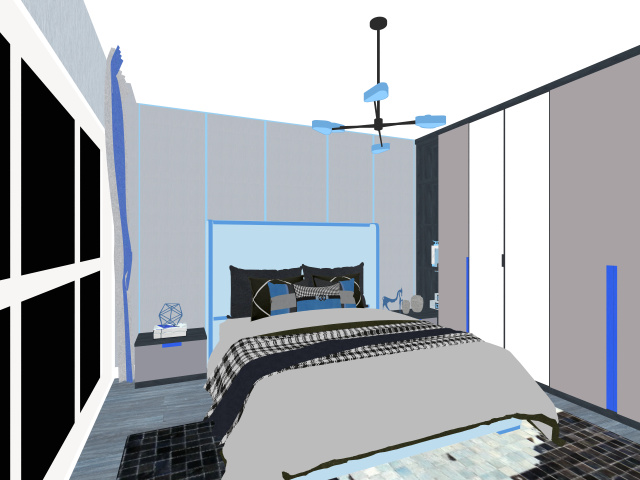} \\
\includegraphics[width=0.15\linewidth]{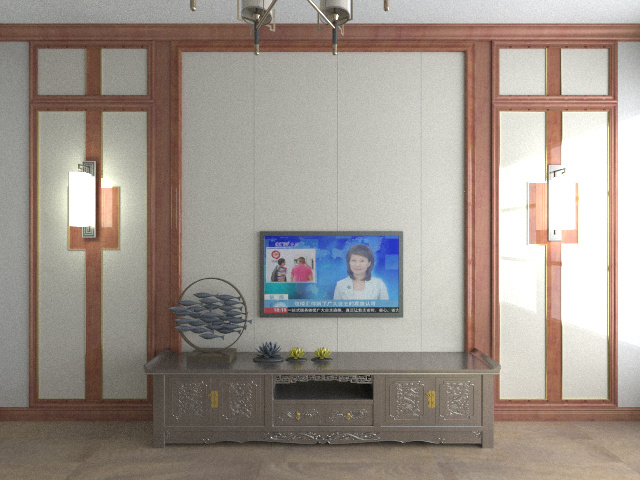} &
\includegraphics[width=0.15\linewidth]{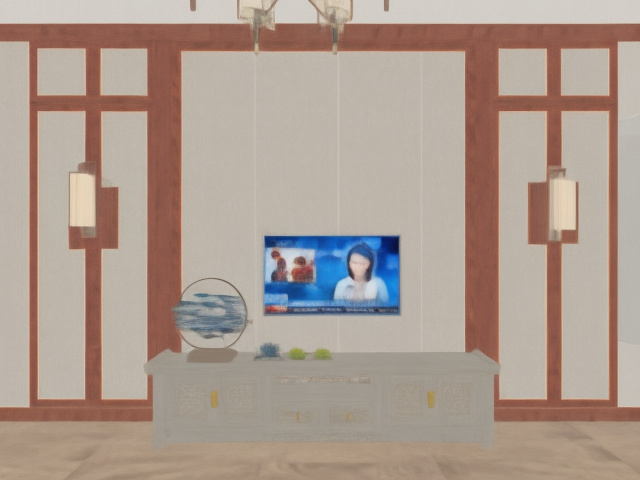} &
\includegraphics[width=0.15\linewidth]{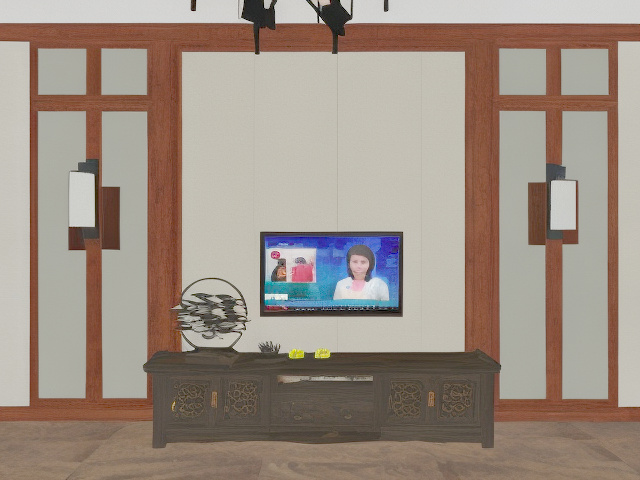} &
\includegraphics[width=0.15\linewidth]{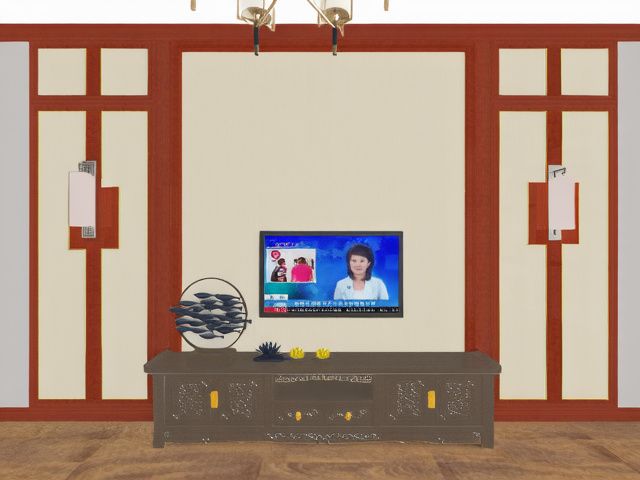} &
\includegraphics[width=0.15\linewidth]{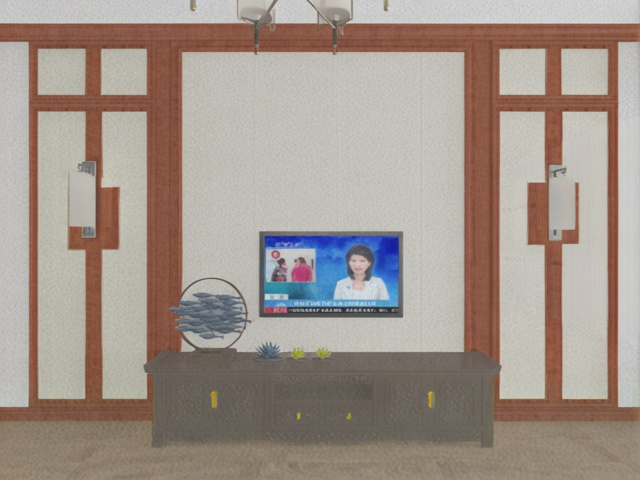} &
\includegraphics[width=0.15\linewidth]{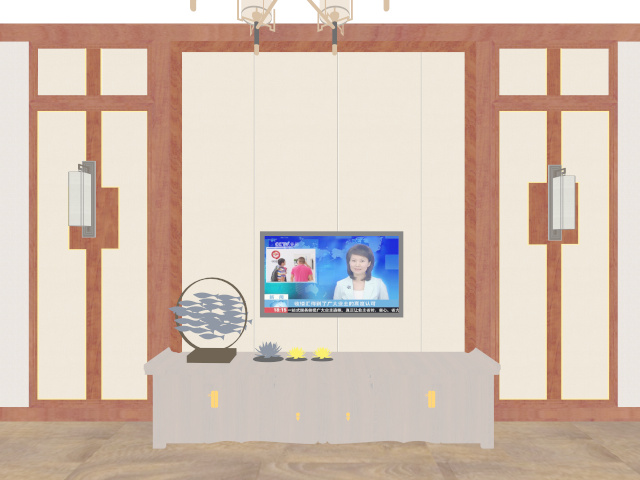} \\
\includegraphics[width=0.15\linewidth]{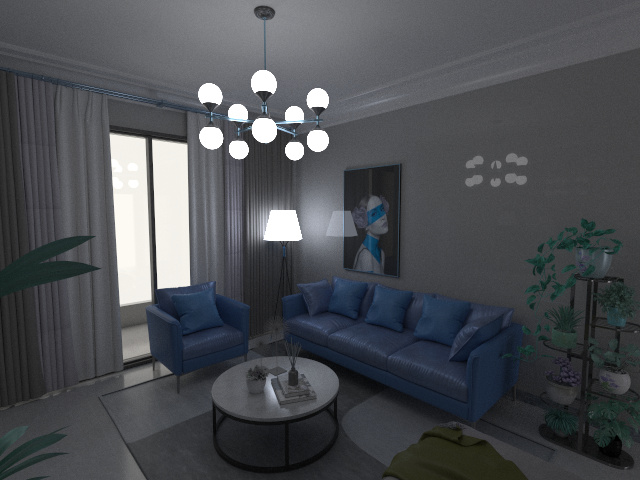} &
\includegraphics[width=0.15\linewidth]{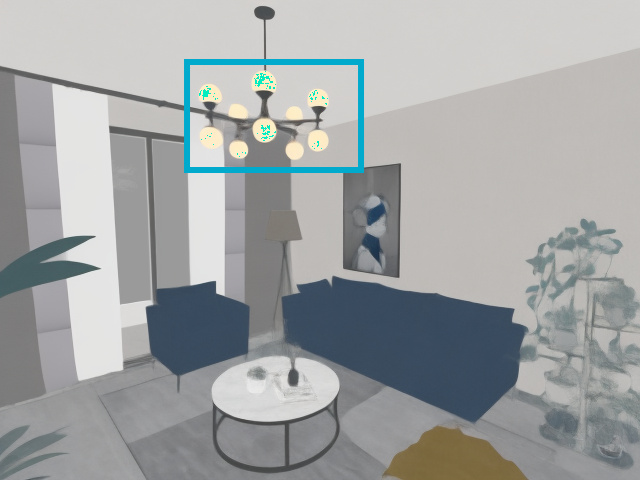} &
\includegraphics[width=0.15\linewidth]{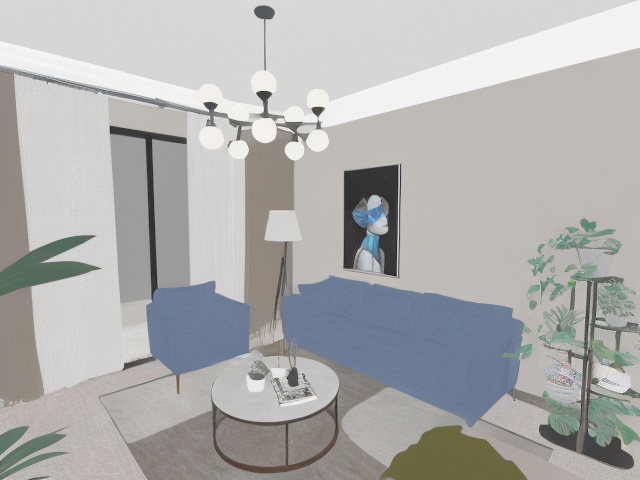} &
\includegraphics[width=0.15\linewidth]{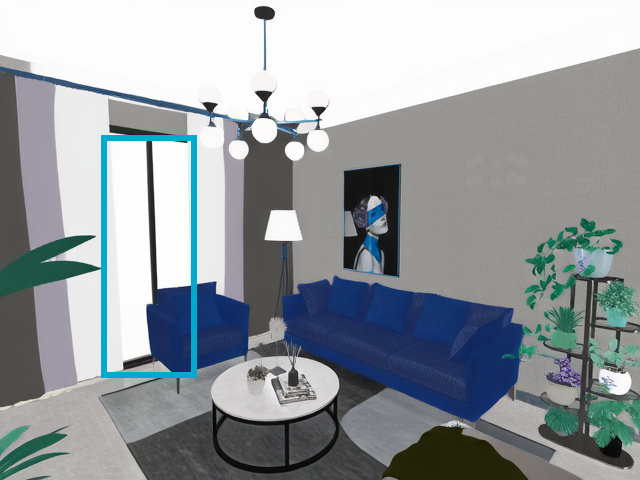} &
\includegraphics[width=0.15\linewidth]{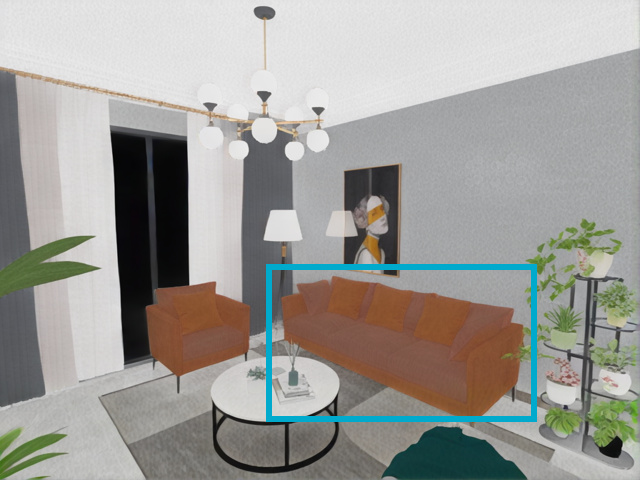} &
\includegraphics[width=0.15\linewidth]{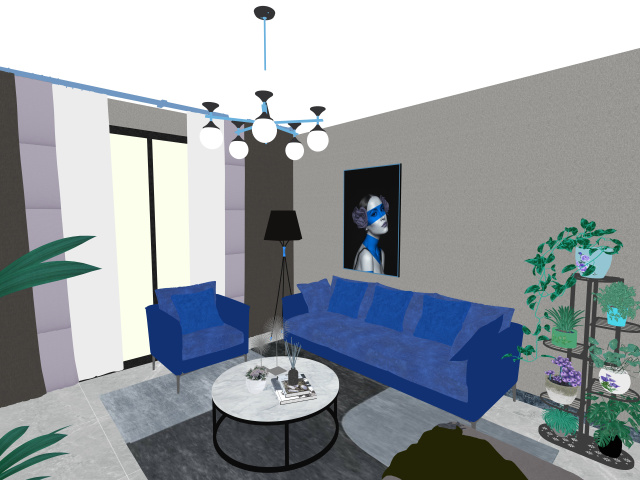} \\
\includegraphics[width=0.15\linewidth]{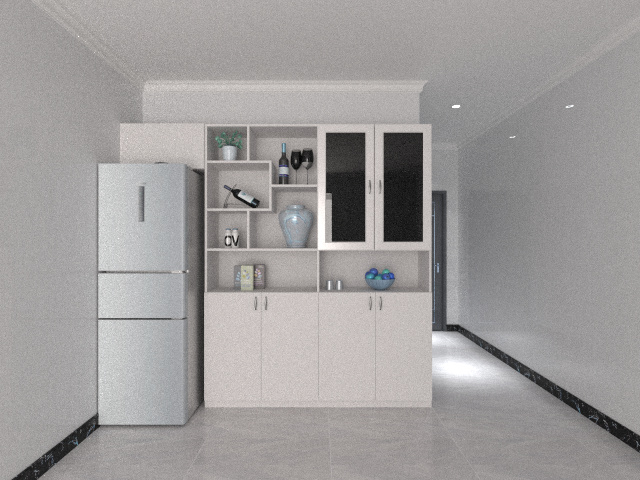} &
\includegraphics[width=0.15\linewidth]{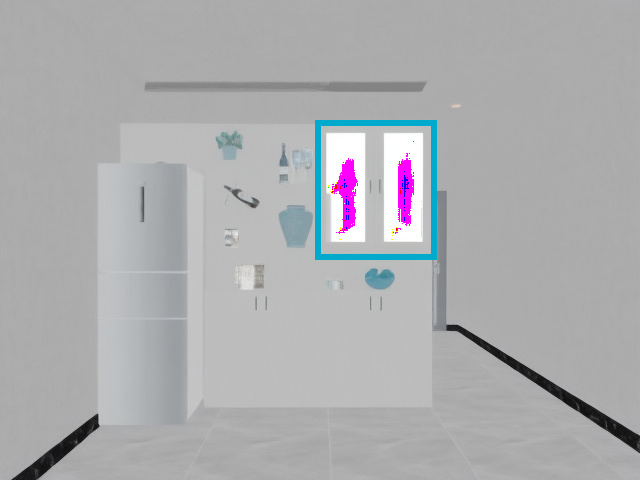} &
\includegraphics[width=0.15\linewidth]{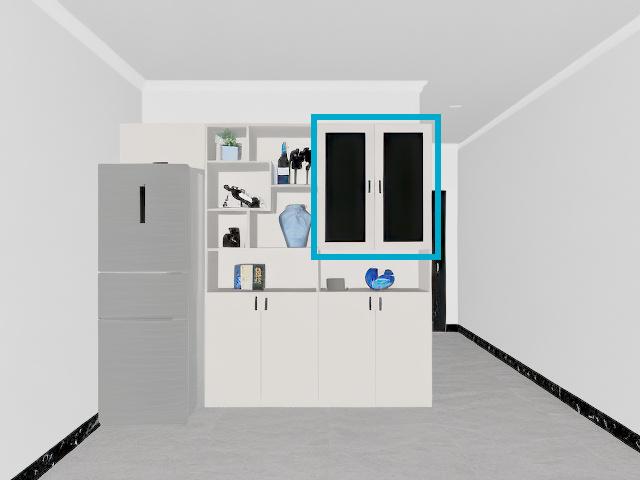} &
\includegraphics[width=0.15\linewidth]{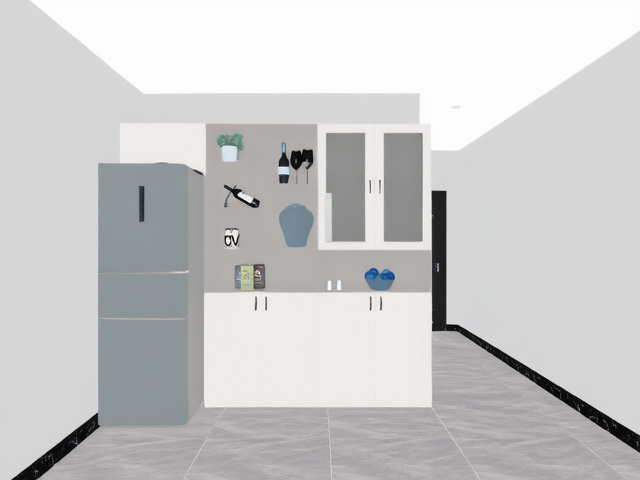} &
\includegraphics[width=0.15\linewidth]{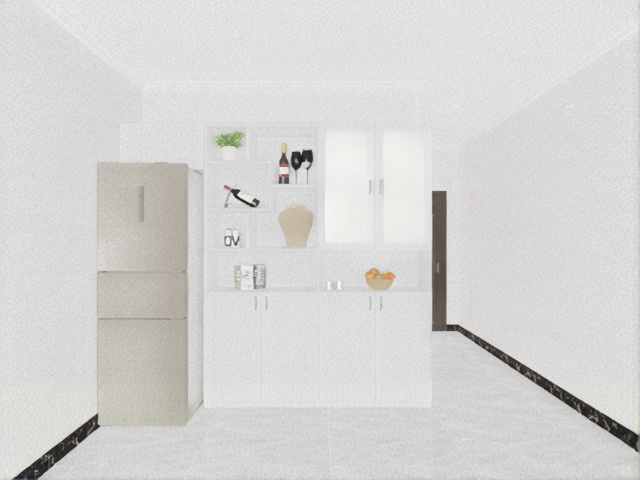} &
\includegraphics[width=0.15\linewidth]{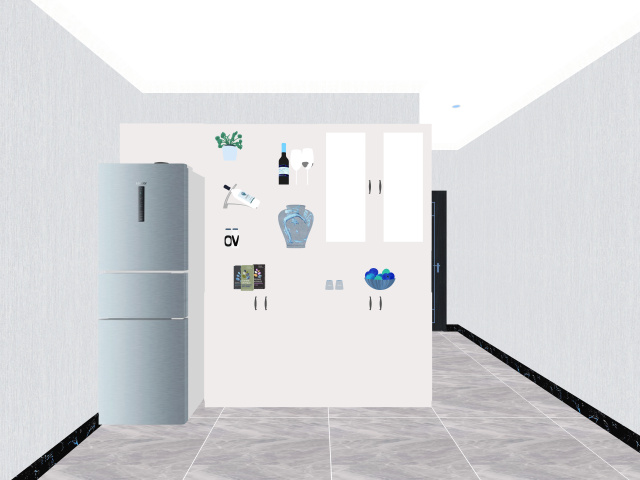} \\
\includegraphics[width=0.15\linewidth]{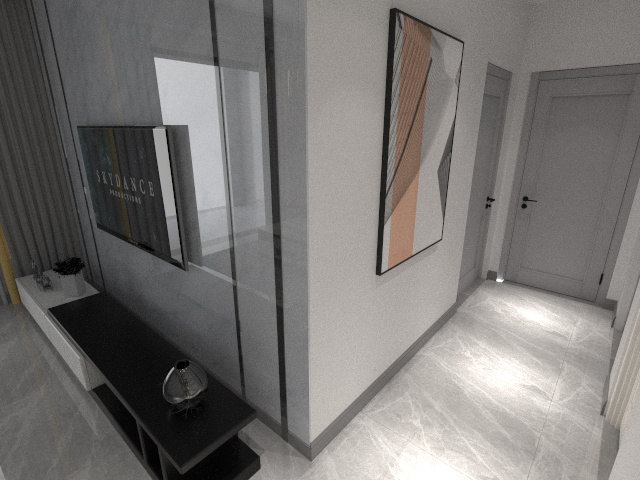} &
\includegraphics[width=0.15\linewidth]{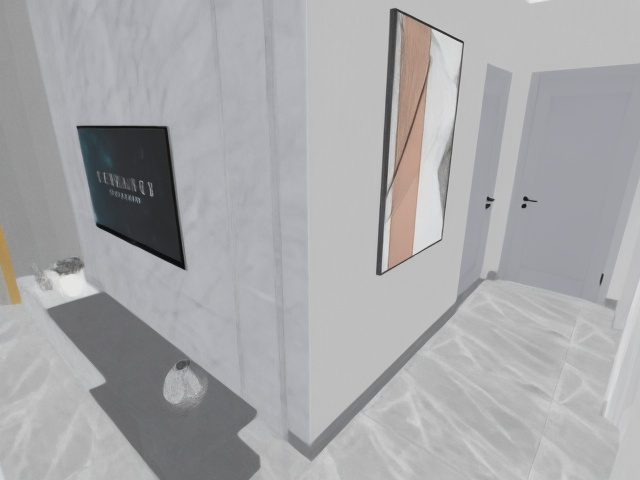} &
\includegraphics[width=0.15\linewidth]{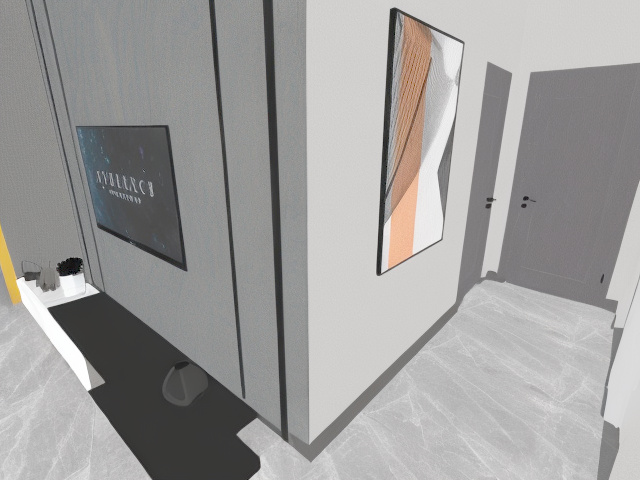} &
\includegraphics[width=0.15\linewidth]{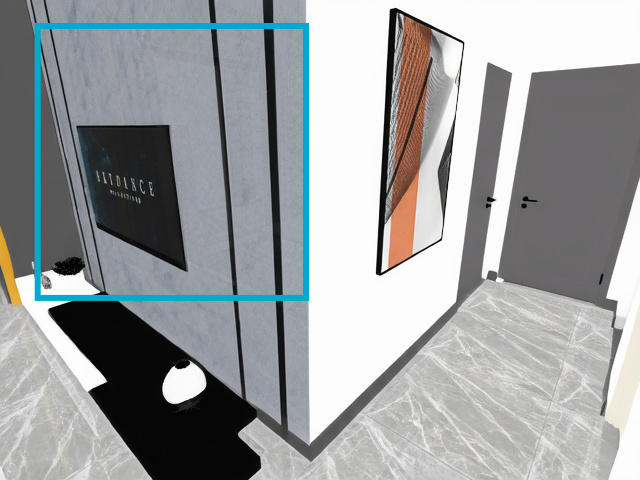} &
\includegraphics[width=0.15\linewidth]{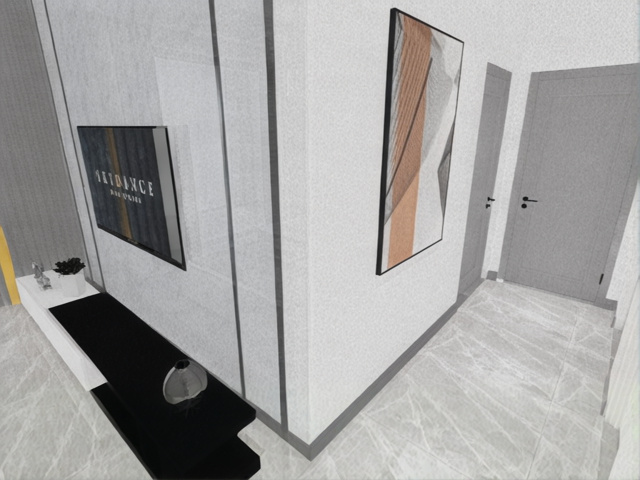} &
\includegraphics[width=0.15\linewidth]{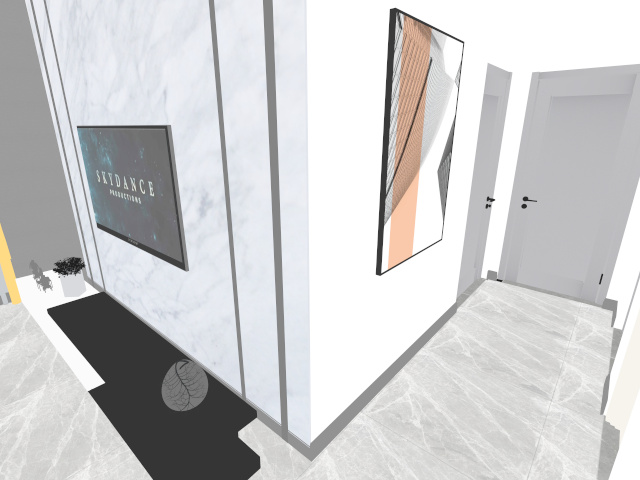} \\
\includegraphics[width=0.15\linewidth]{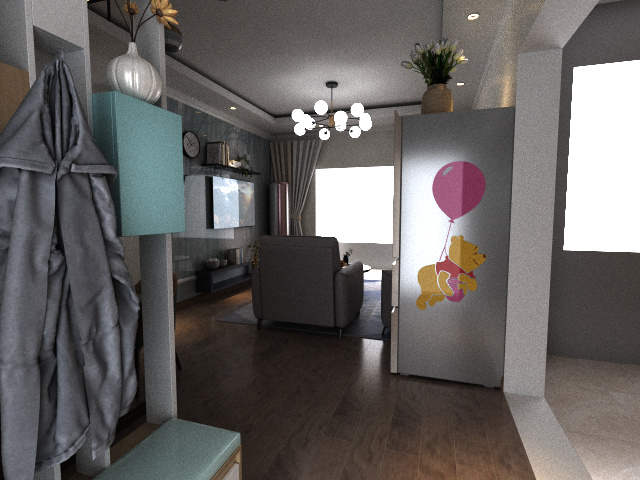} &
\includegraphics[width=0.15\linewidth]{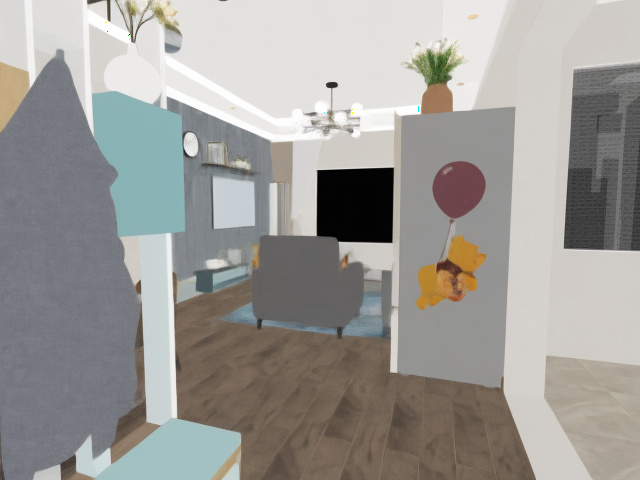} &
\includegraphics[width=0.15\linewidth]{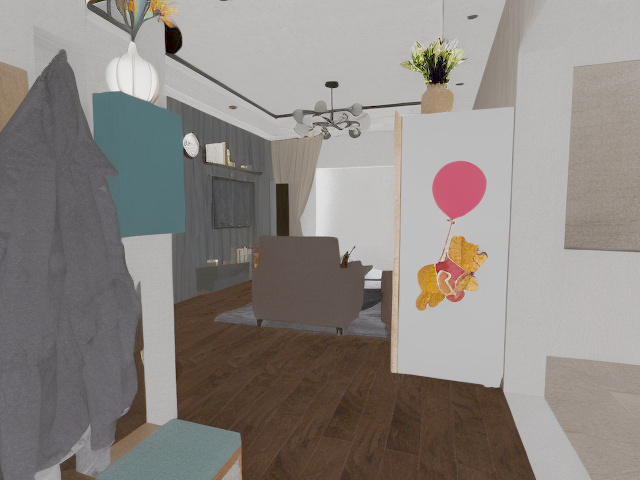} &
\includegraphics[width=0.15\linewidth]{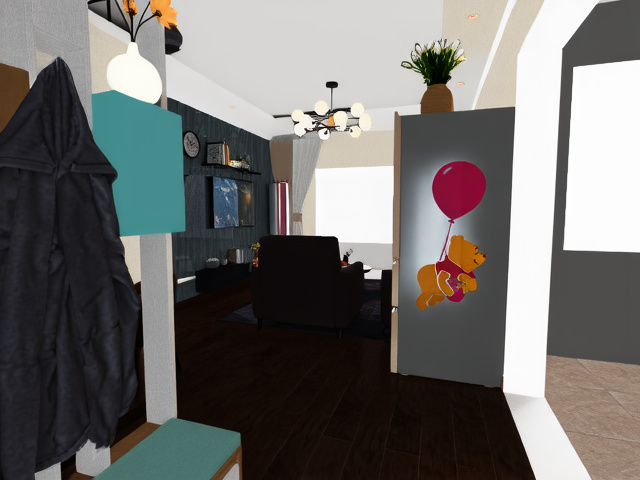} &
\includegraphics[width=0.15\linewidth]{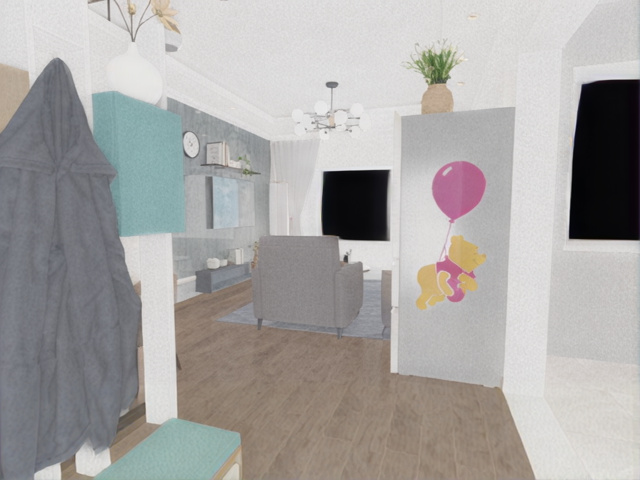} &
\includegraphics[width=0.15\linewidth]{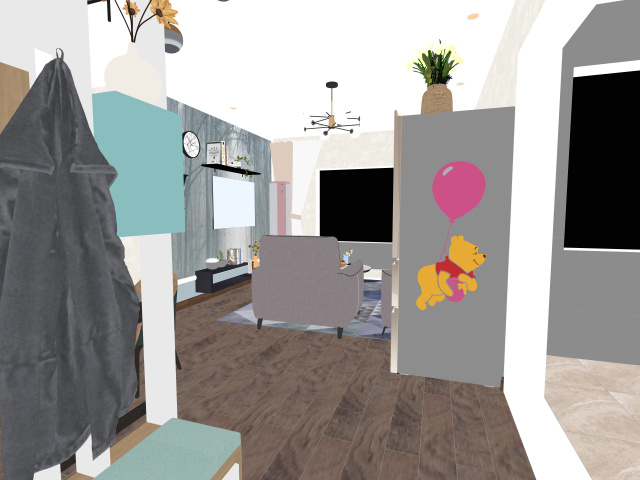} \\
\end{tabular}
\caption{Qualitative comparison on the InteriorVerse dataset. Cyan rectangles highlight the hallucinated mirror reflection in Kocsis et al. (scene 1), the near-zero reflectance collapse on reflective and transparent surfaces in RGB$\leftrightarrow$X (scenes 1 and 5), the color shift introduced by our method (scenes 1 and 3), and artifacts introduced by Kocsis et al. on the chandelier and wall elements (scenes 3 and 4).}
\label{fig:sota_iv}
\end{figure*}

The qualitative results on InteriorVerse (Figure~\ref{fig:sota_iv}) show that our method most reliably suppresses spurious reflections, such as the mirror in scene~1, that other methods leave baked into the albedo, at the cost of an occasional color shift on reflective or blue surfaces.

\begin{figure*}[tbp]
\centering
\setlength{\tabcolsep}{1pt}
\renewcommand{\arraystretch}{1}
\begin{tabular}{cccccc}
\footnotesize Input RGB & \footnotesize Zhu et al. & \footnotesize Kocsis et al. & \footnotesize RGB$\leftrightarrow$X & \footnotesize Ours & \footnotesize GT Reflectance \\
\includegraphics[width=0.15\linewidth]{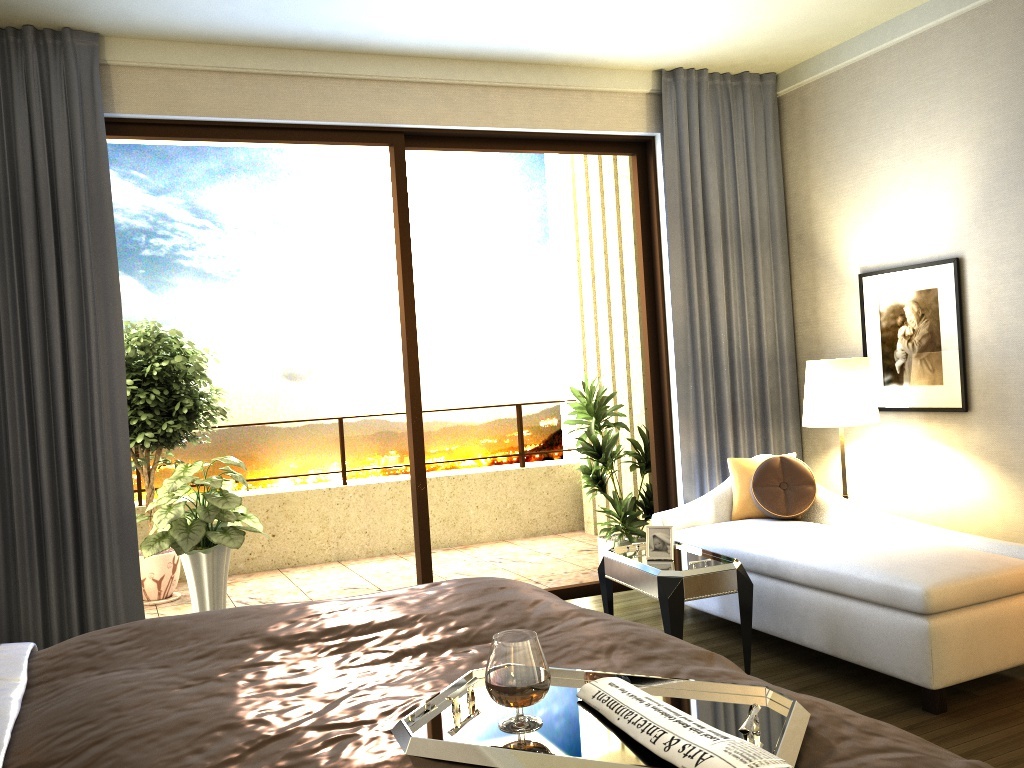} &
\includegraphics[width=0.15\linewidth]{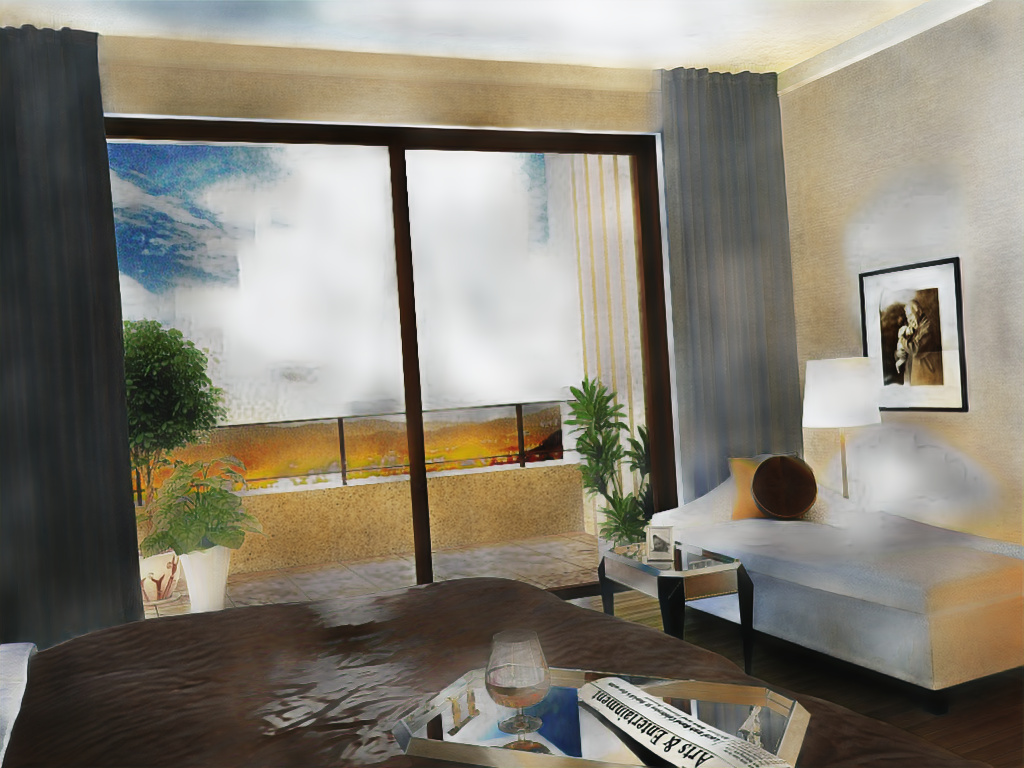} &
\includegraphics[width=0.15\linewidth]{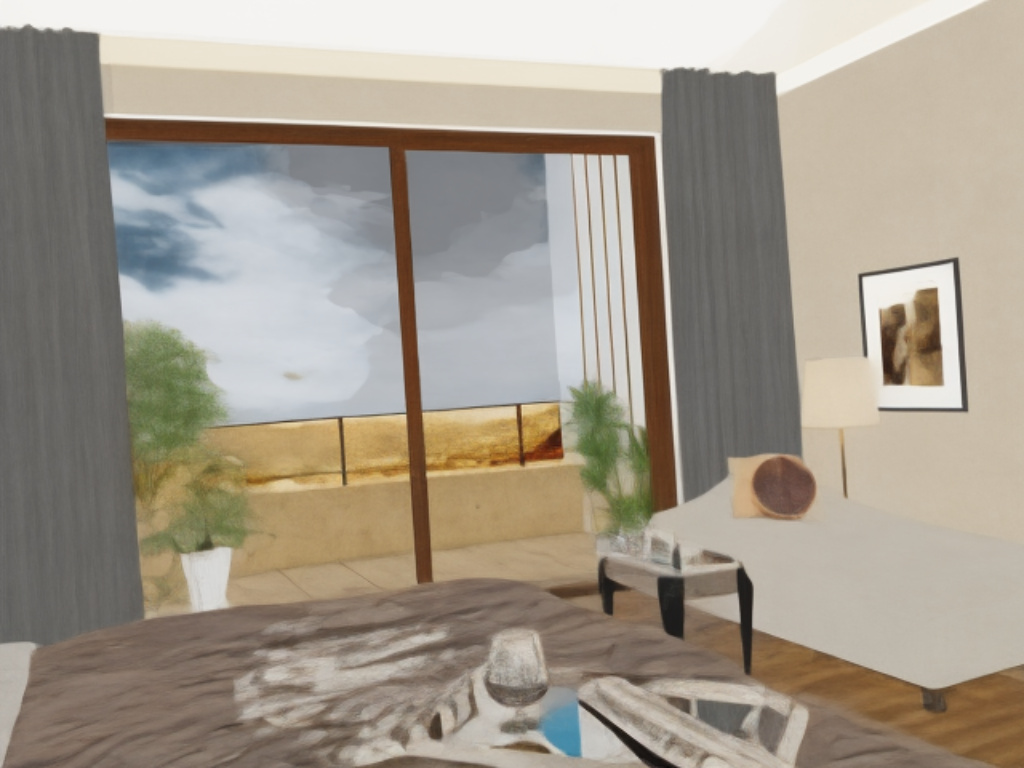} &
\includegraphics[width=0.15\linewidth]{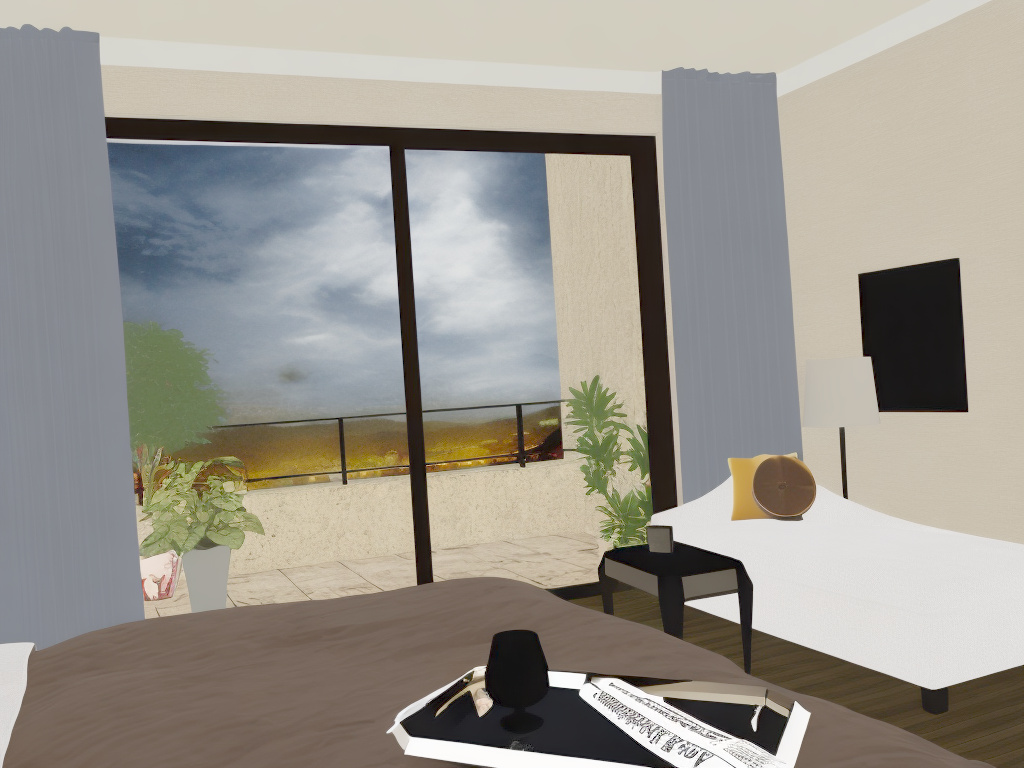} &
\includegraphics[width=0.15\linewidth]{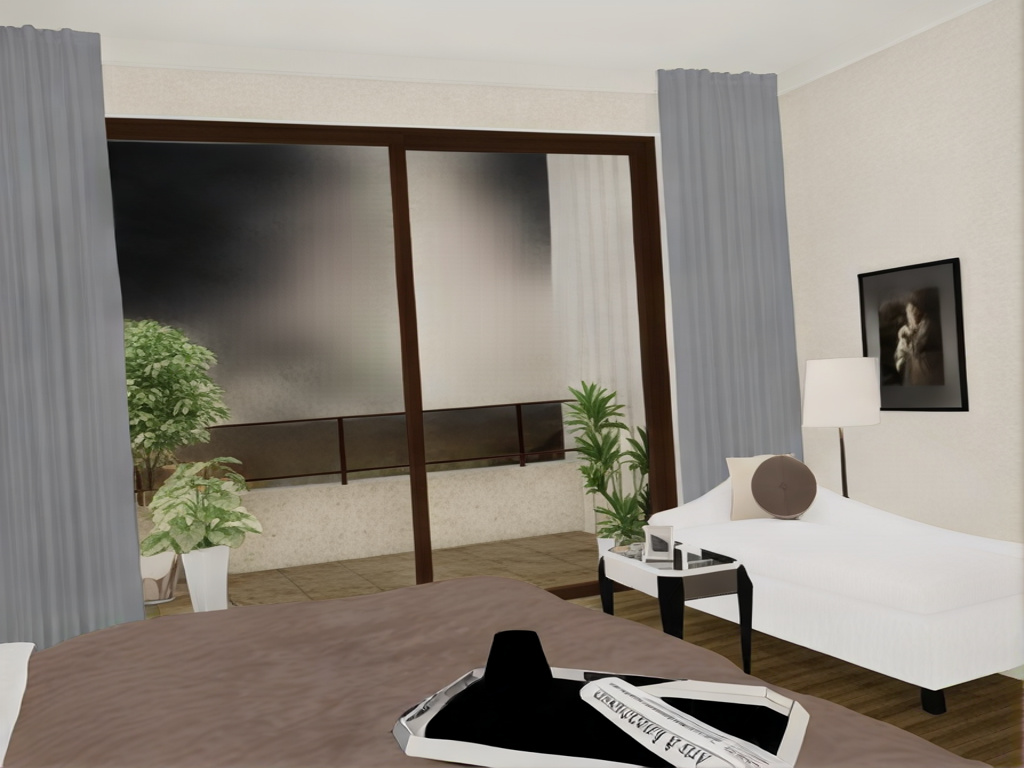} &
\includegraphics[width=0.15\linewidth]{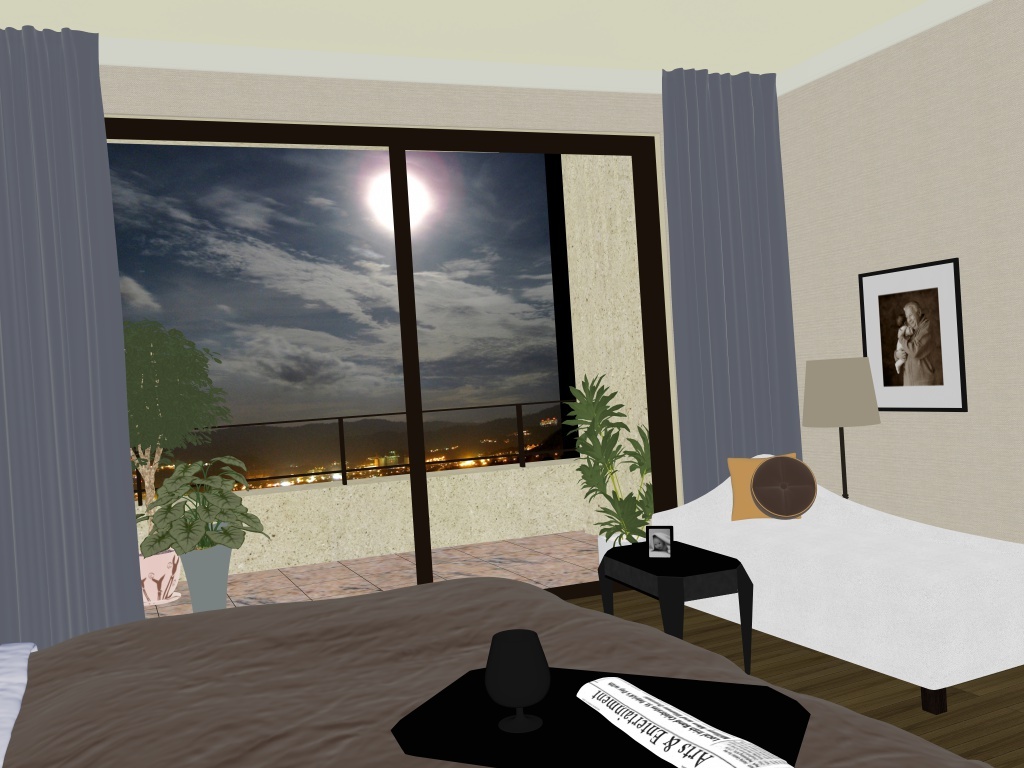} \\
\includegraphics[width=0.15\linewidth]{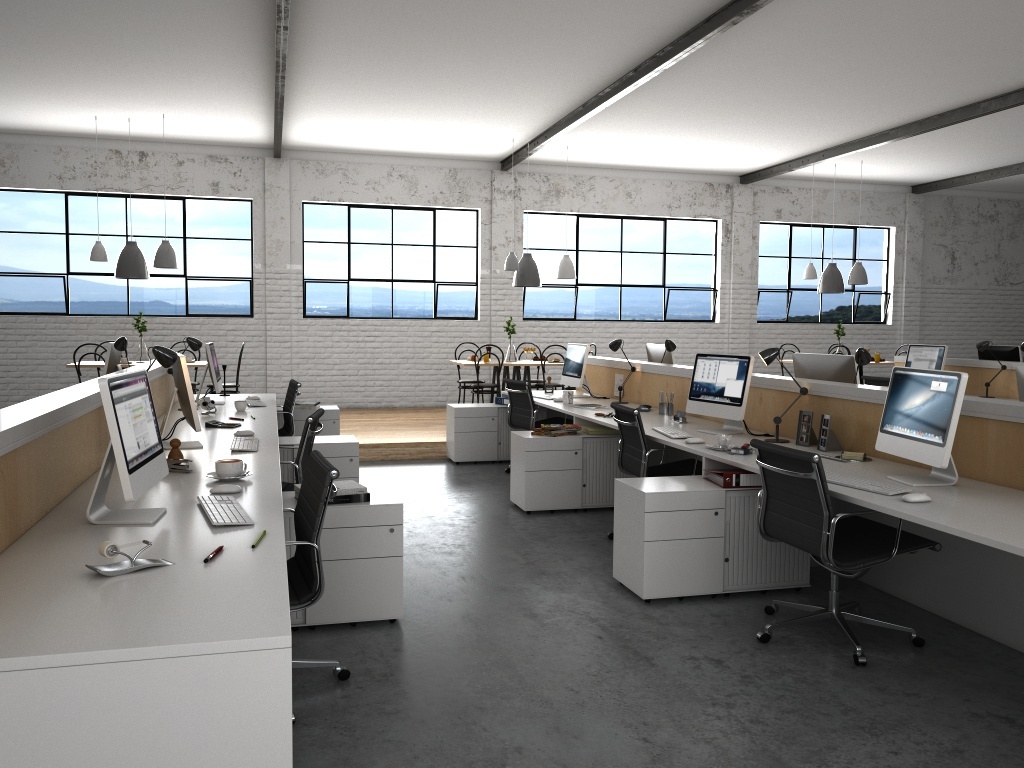} &
\includegraphics[width=0.15\linewidth]{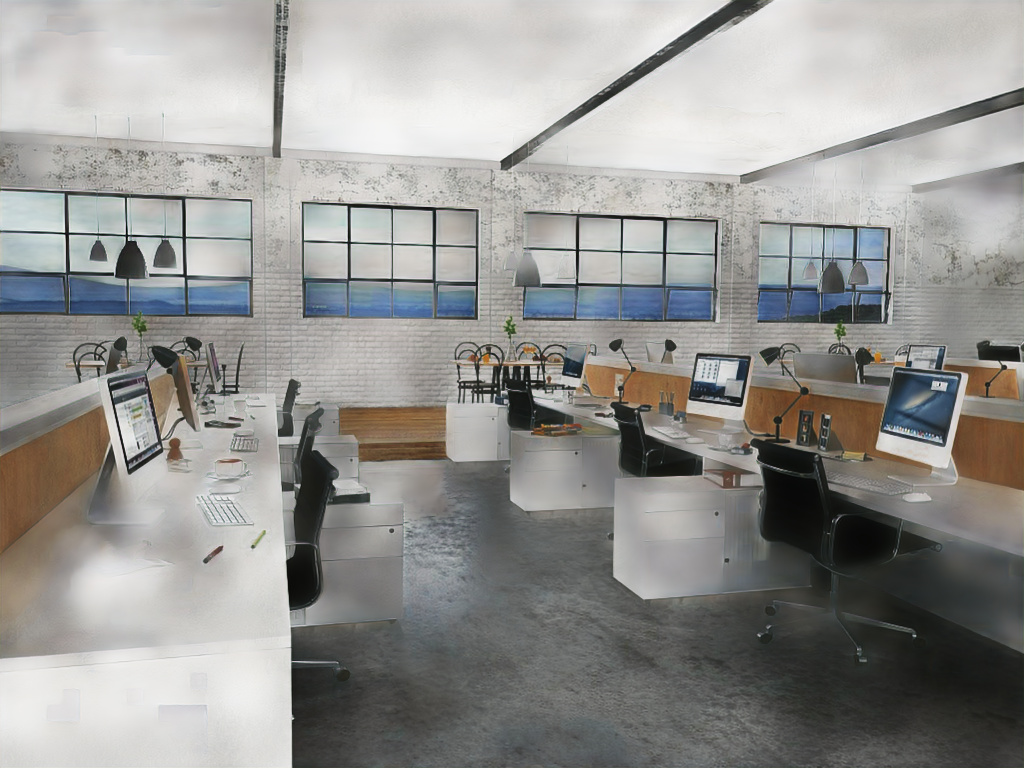} &
\includegraphics[width=0.15\linewidth]{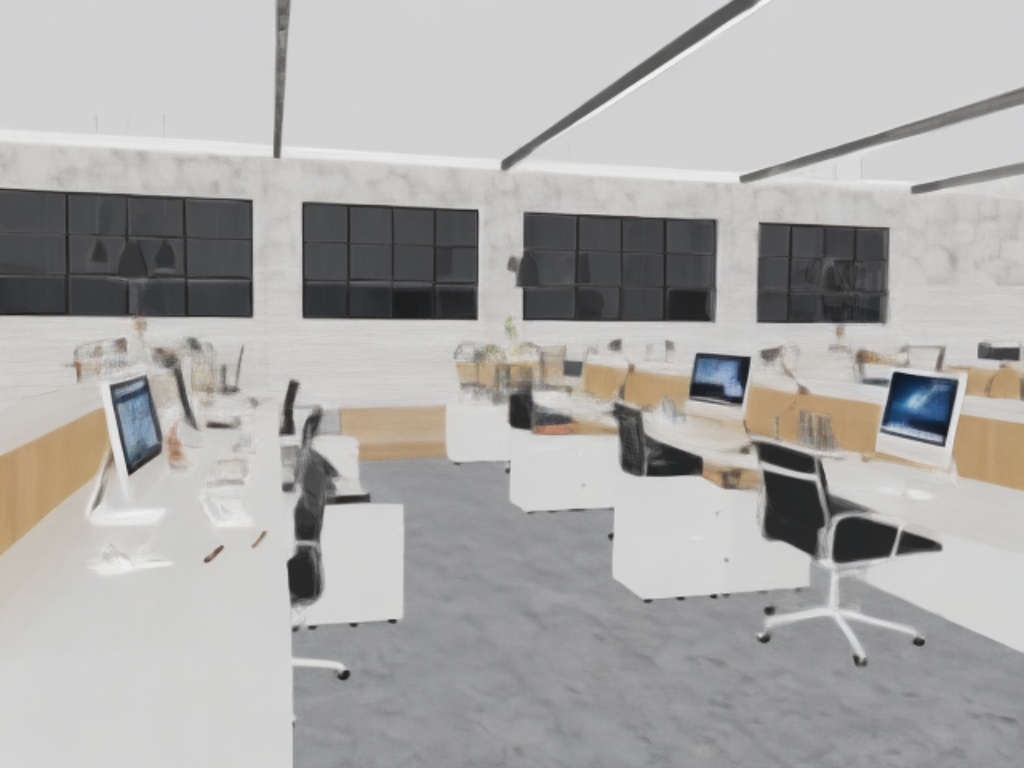} &
\includegraphics[width=0.15\linewidth]{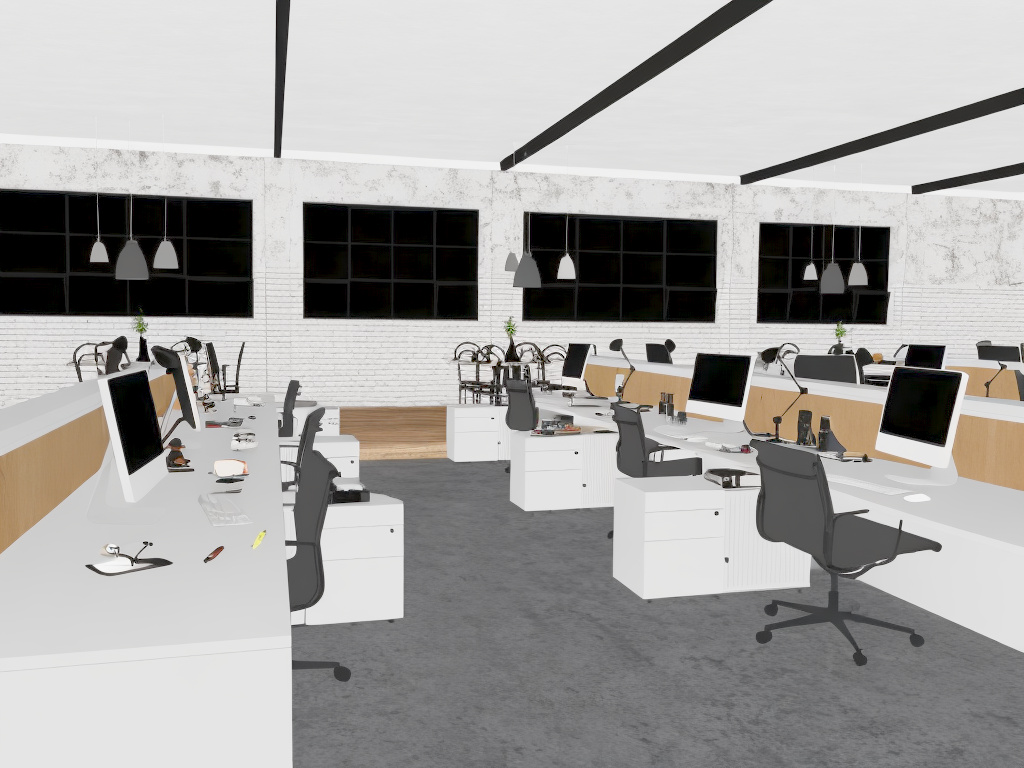} &
\includegraphics[width=0.15\linewidth]{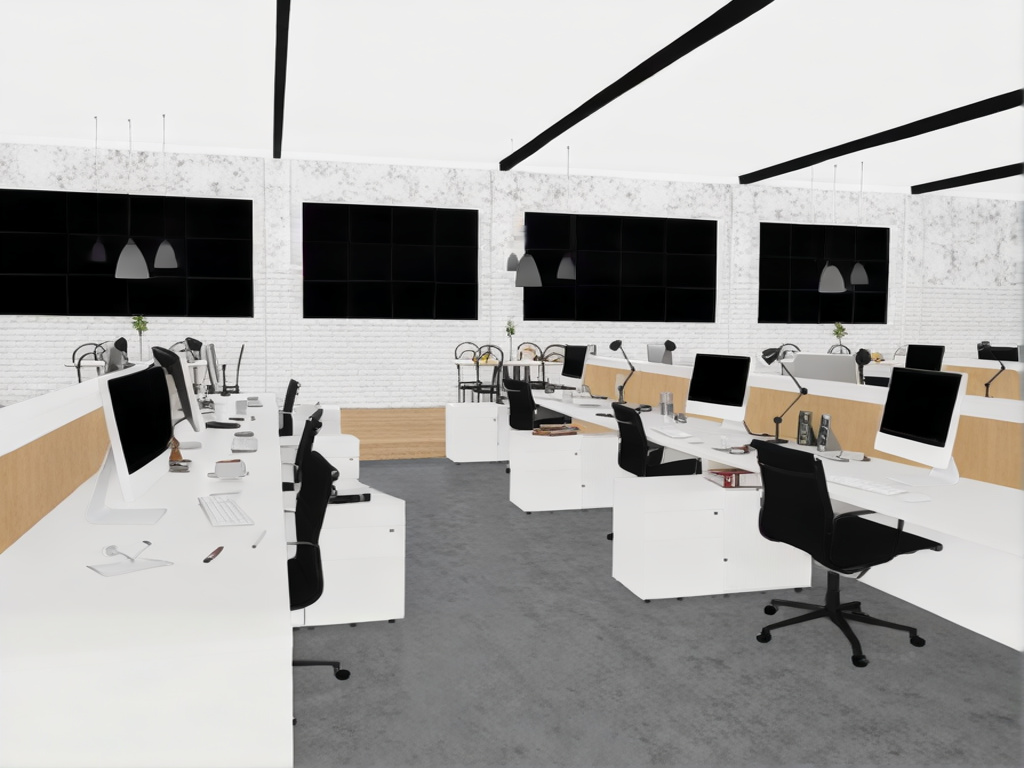} &
\includegraphics[width=0.15\linewidth]{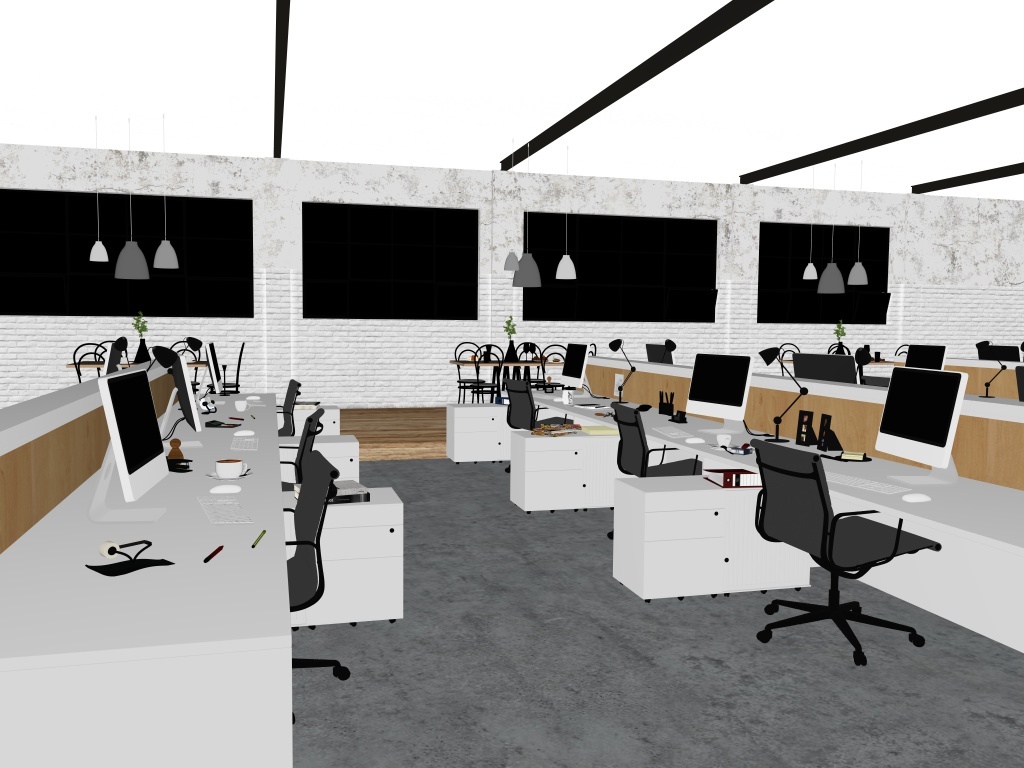} \\
\includegraphics[width=0.15\linewidth]{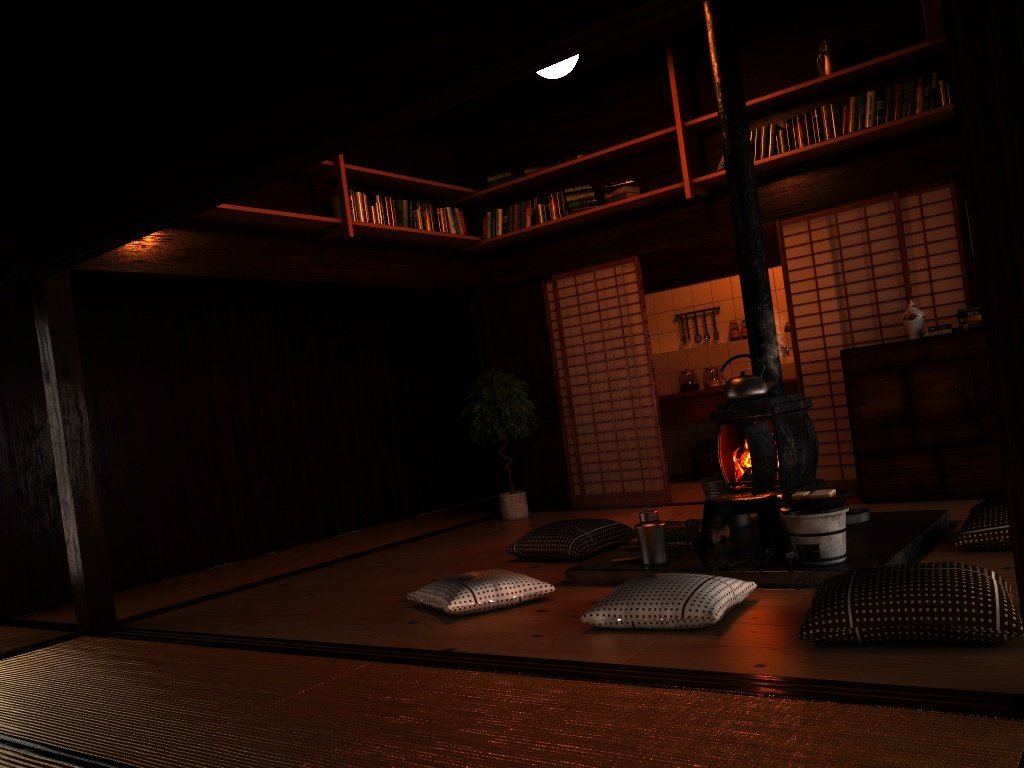} &
\includegraphics[width=0.15\linewidth]{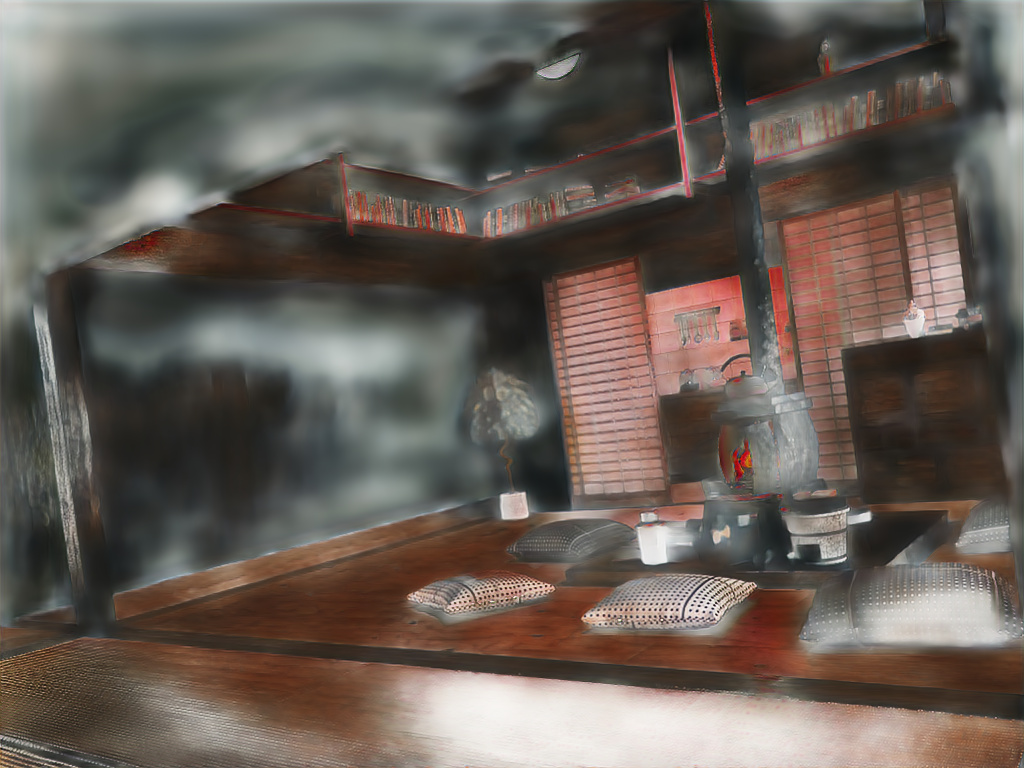} &
\includegraphics[width=0.15\linewidth]{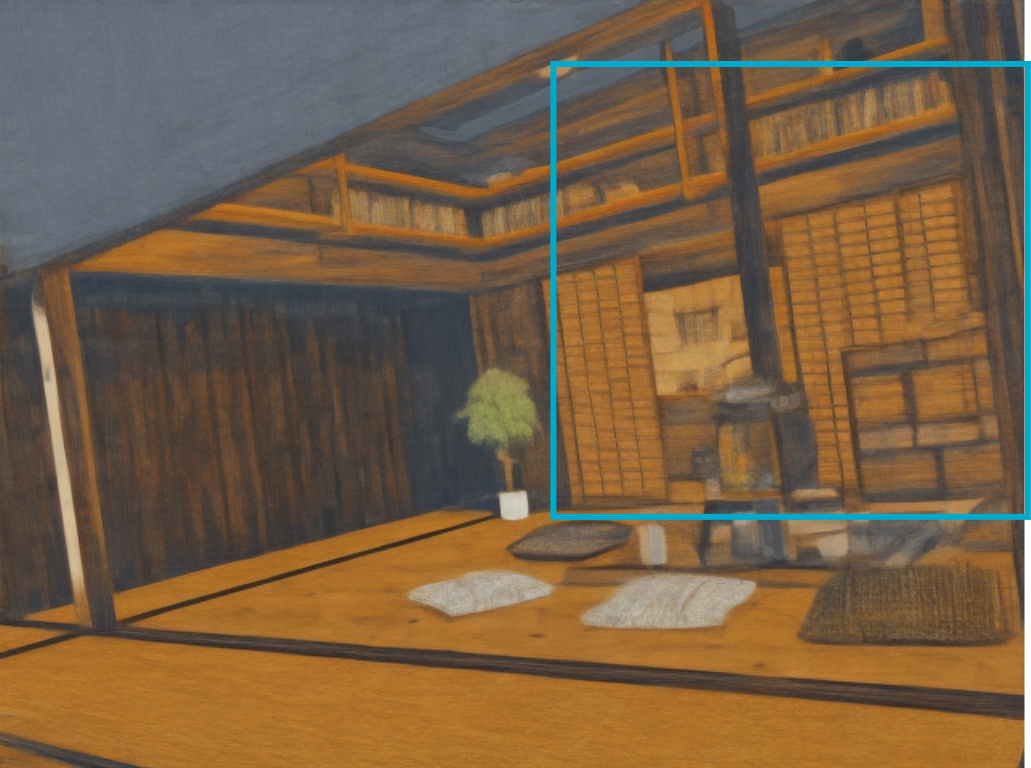} &
\includegraphics[width=0.15\linewidth]{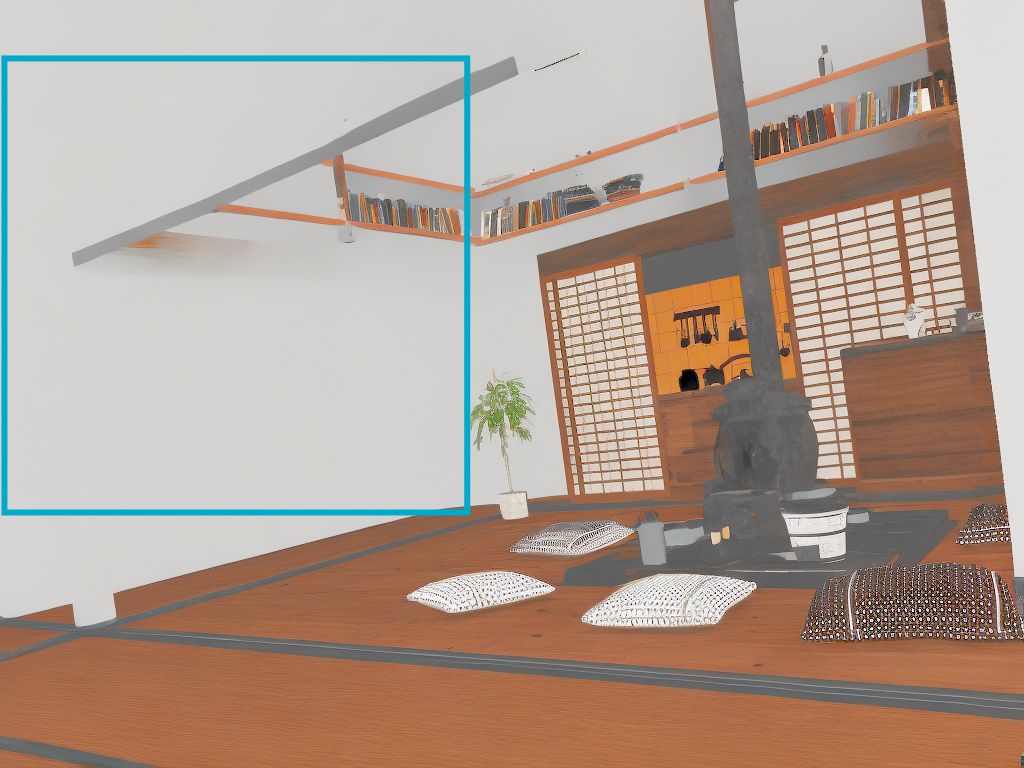} &
\includegraphics[width=0.15\linewidth]{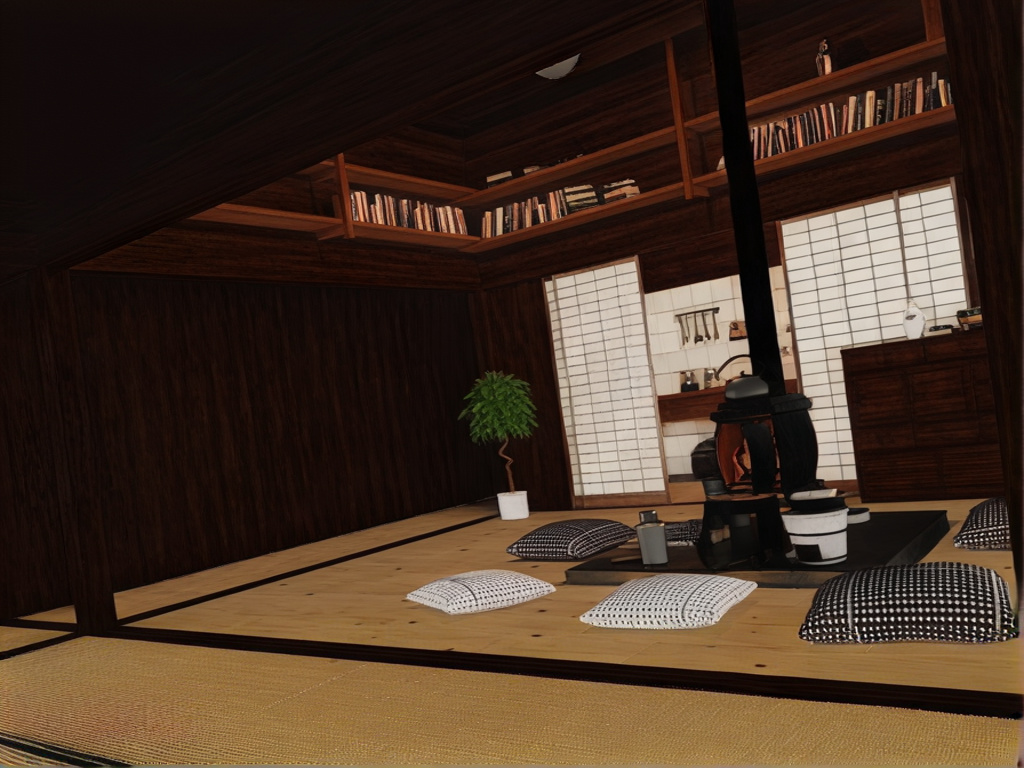} &
\includegraphics[width=0.15\linewidth]{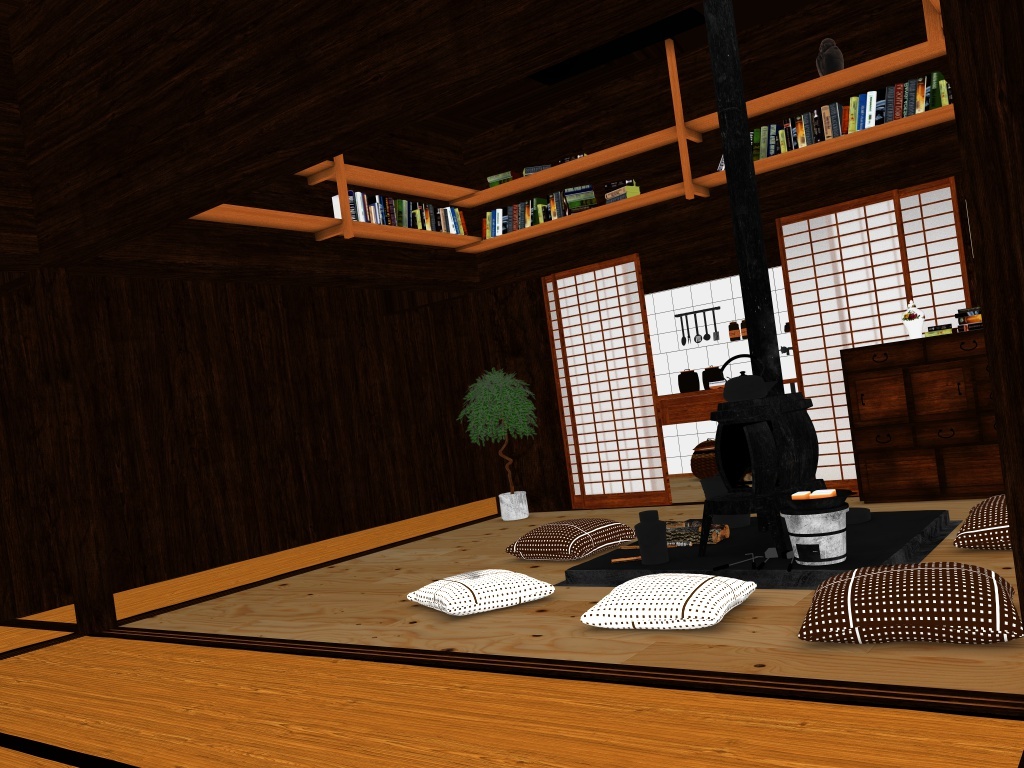} \\
\includegraphics[width=0.15\linewidth]{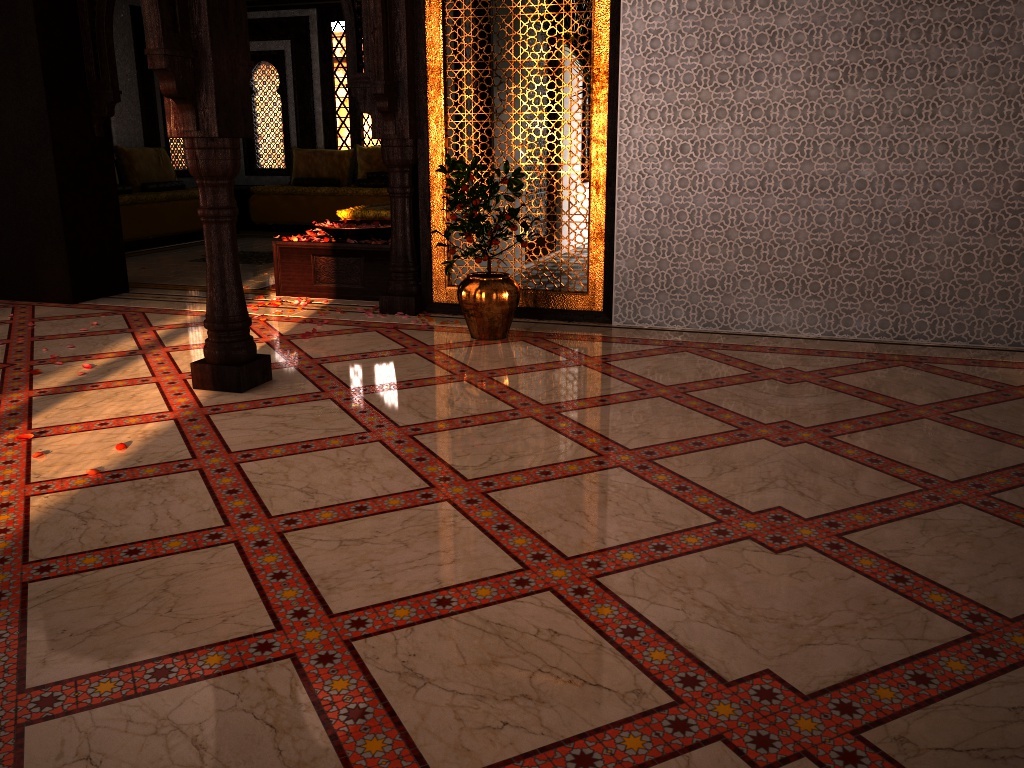} &
\includegraphics[width=0.15\linewidth]{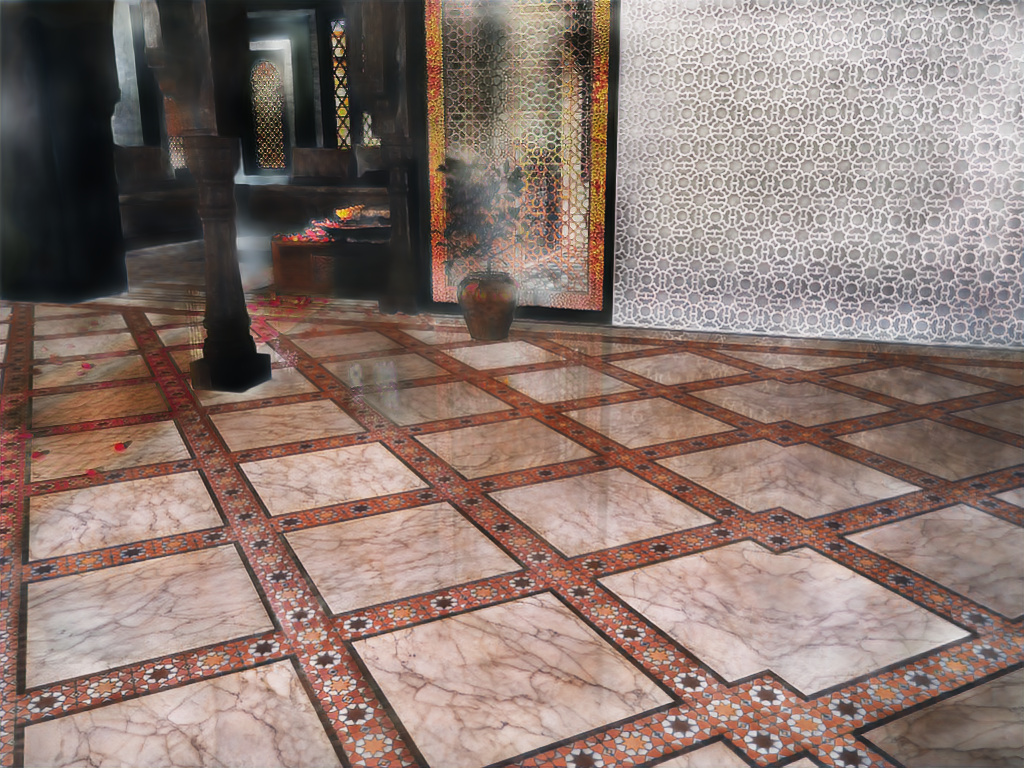} &
\includegraphics[width=0.15\linewidth]{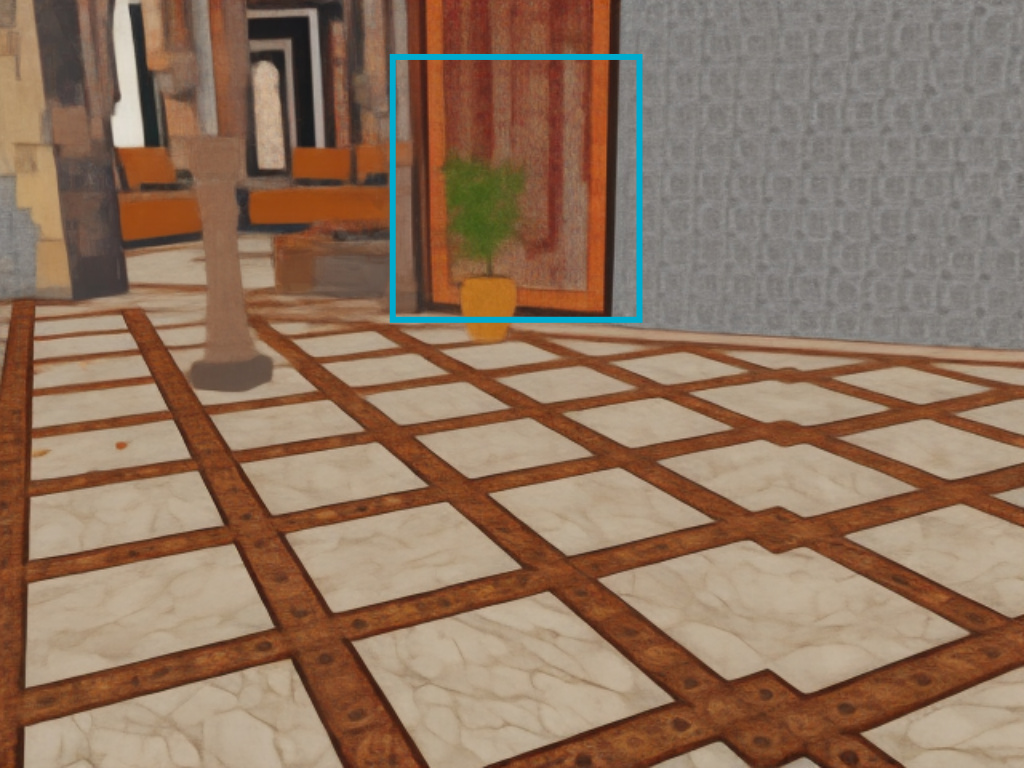} &
\includegraphics[width=0.15\linewidth]{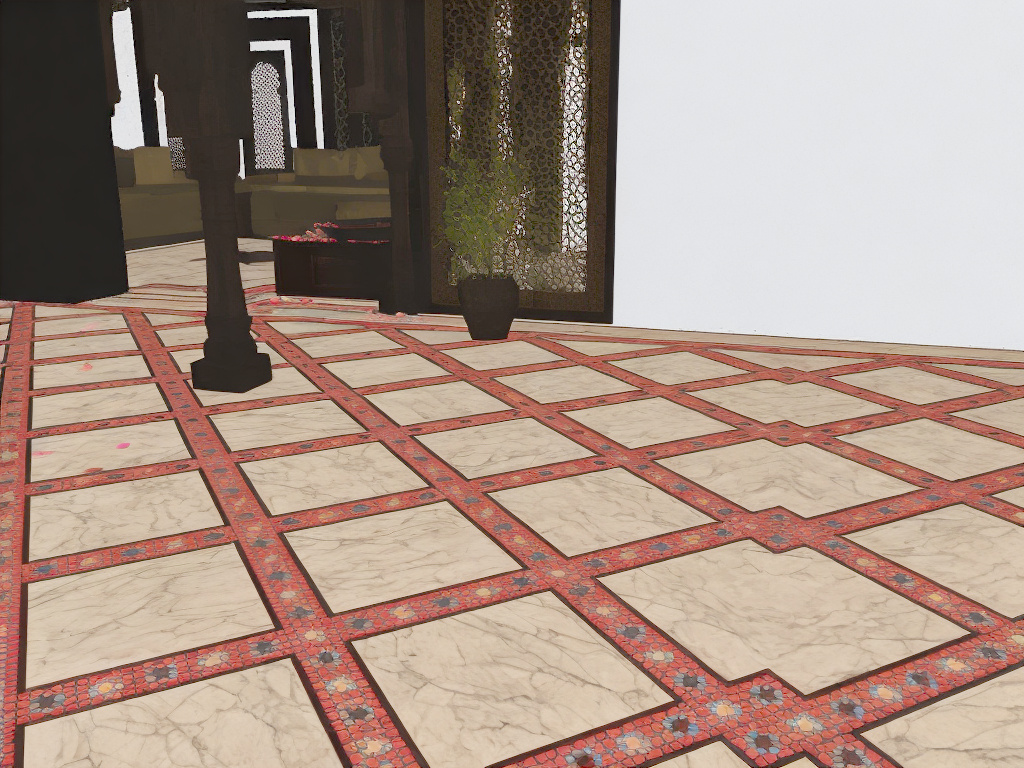} &
\includegraphics[width=0.15\linewidth]{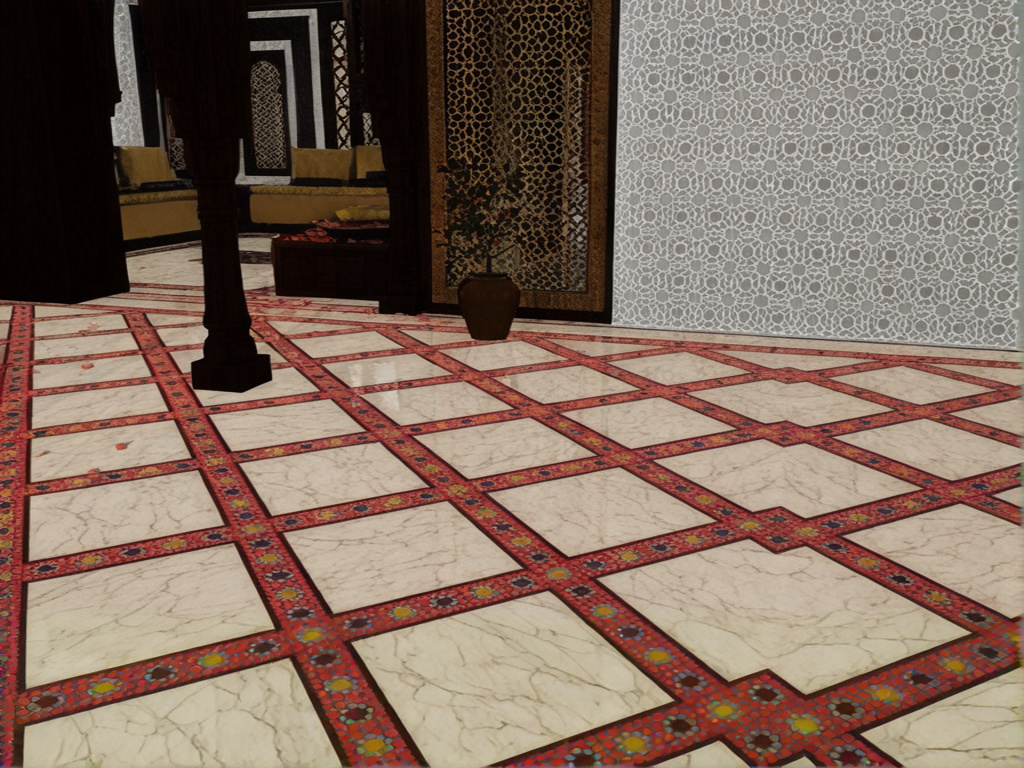} &
\includegraphics[width=0.15\linewidth]{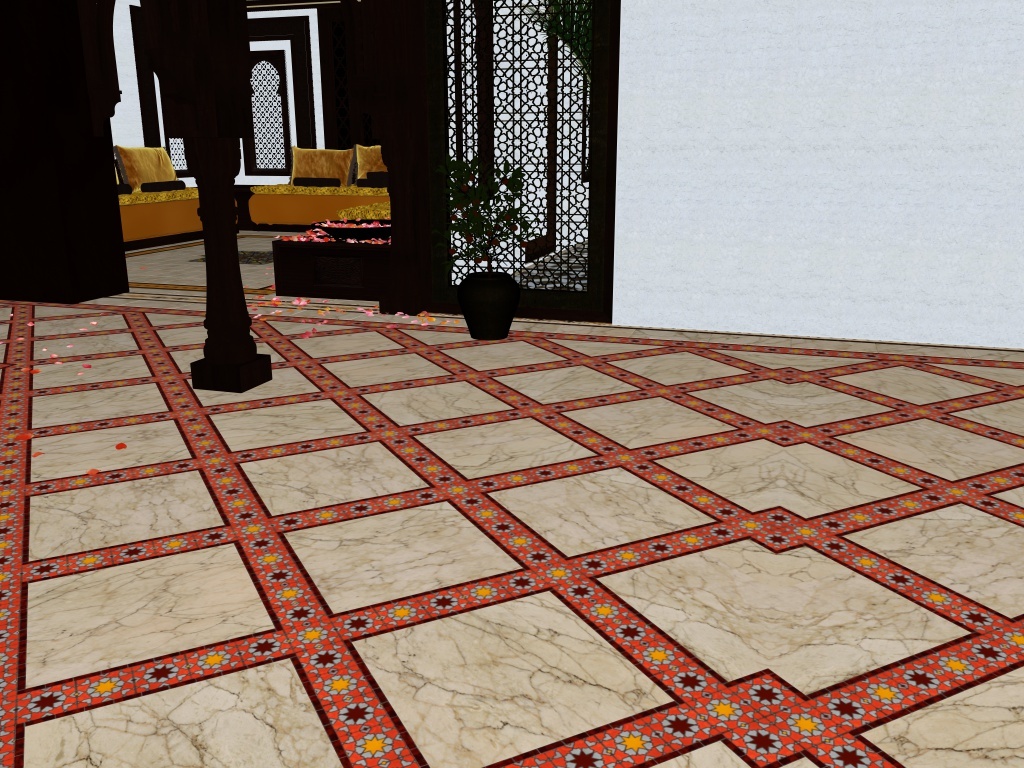} \\
\includegraphics[width=0.15\linewidth]{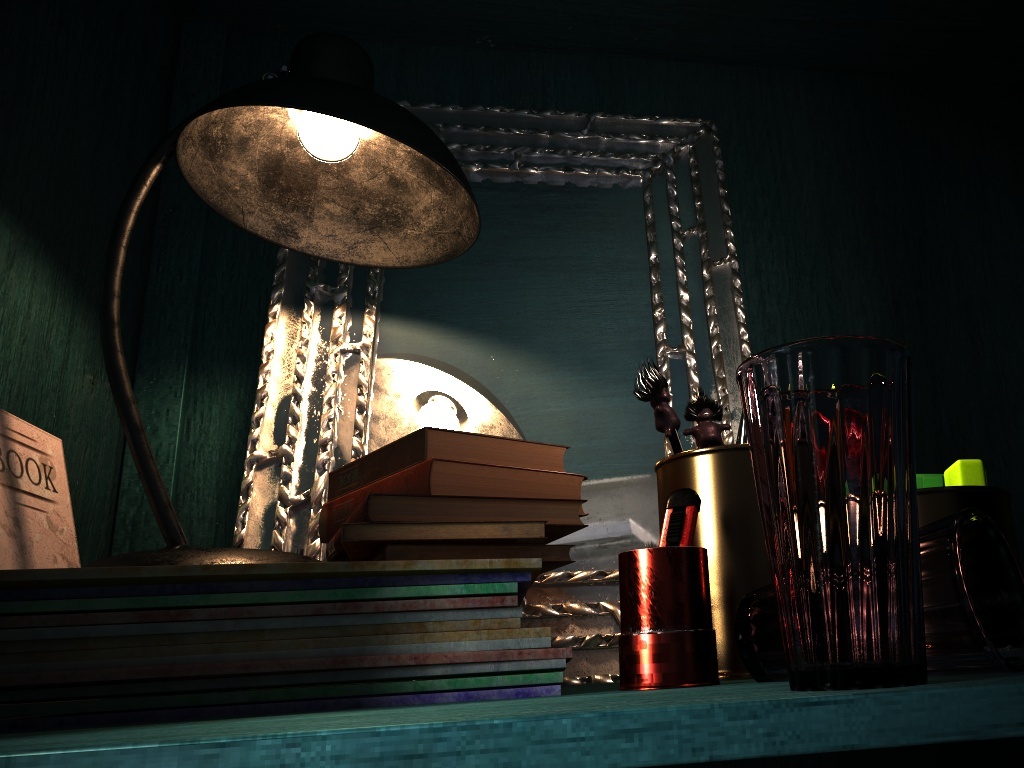} &
\includegraphics[width=0.15\linewidth]{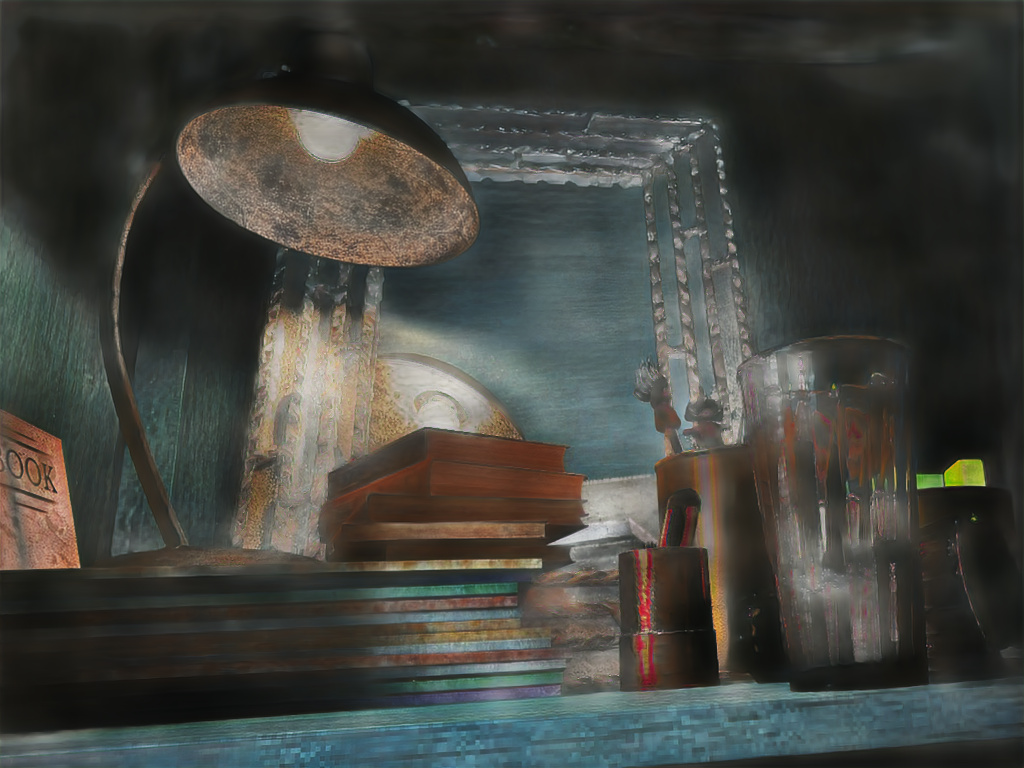} &
\includegraphics[width=0.15\linewidth]{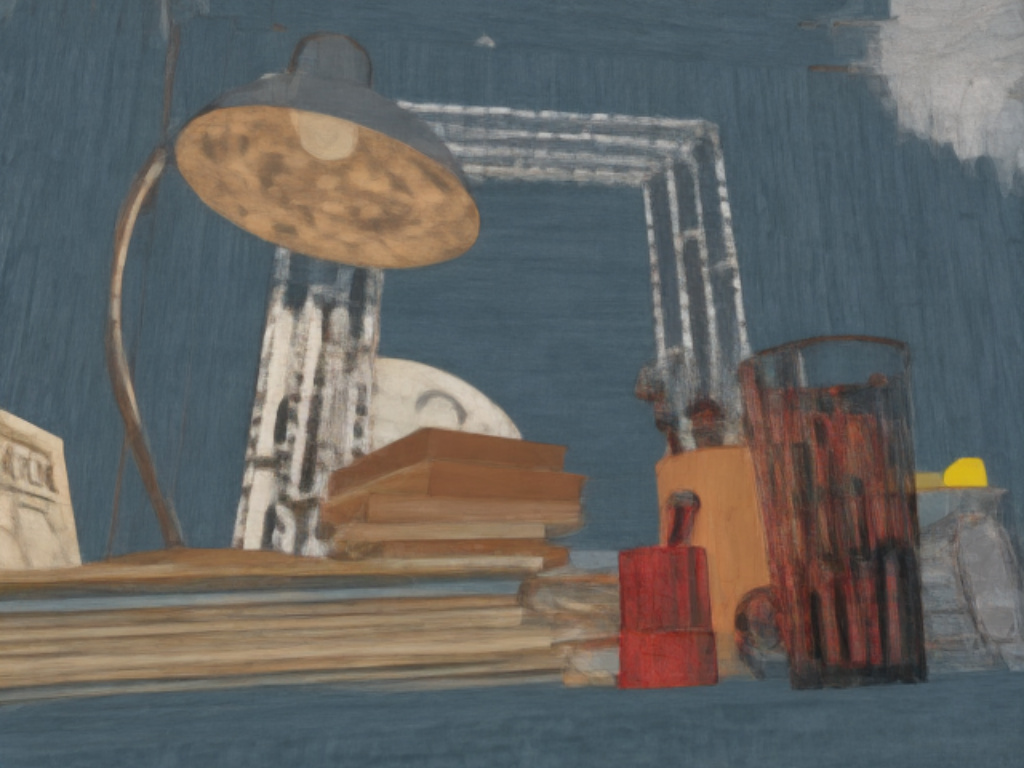} &
\includegraphics[width=0.15\linewidth]{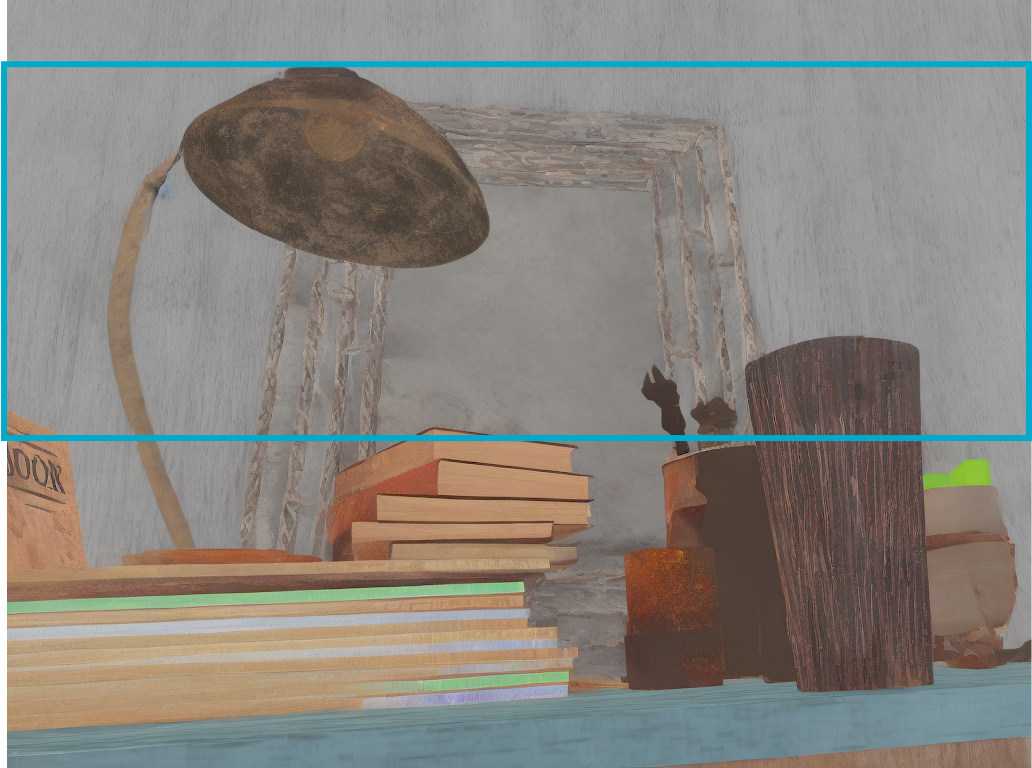} &
\includegraphics[width=0.15\linewidth]{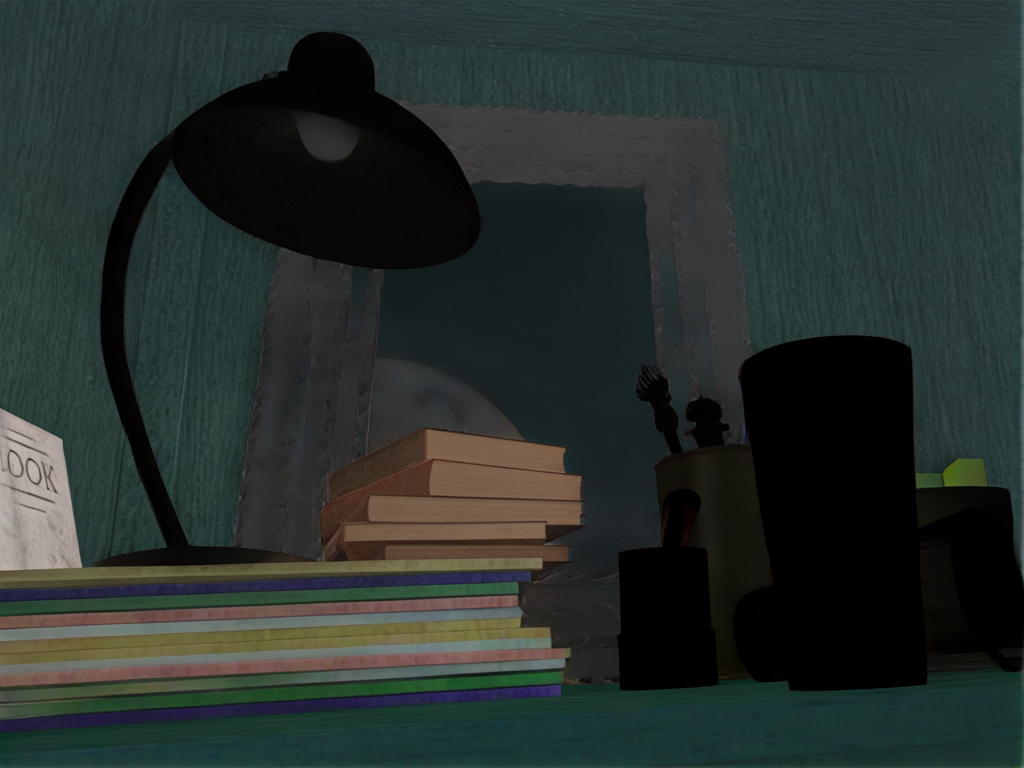} &
\includegraphics[width=0.15\linewidth]{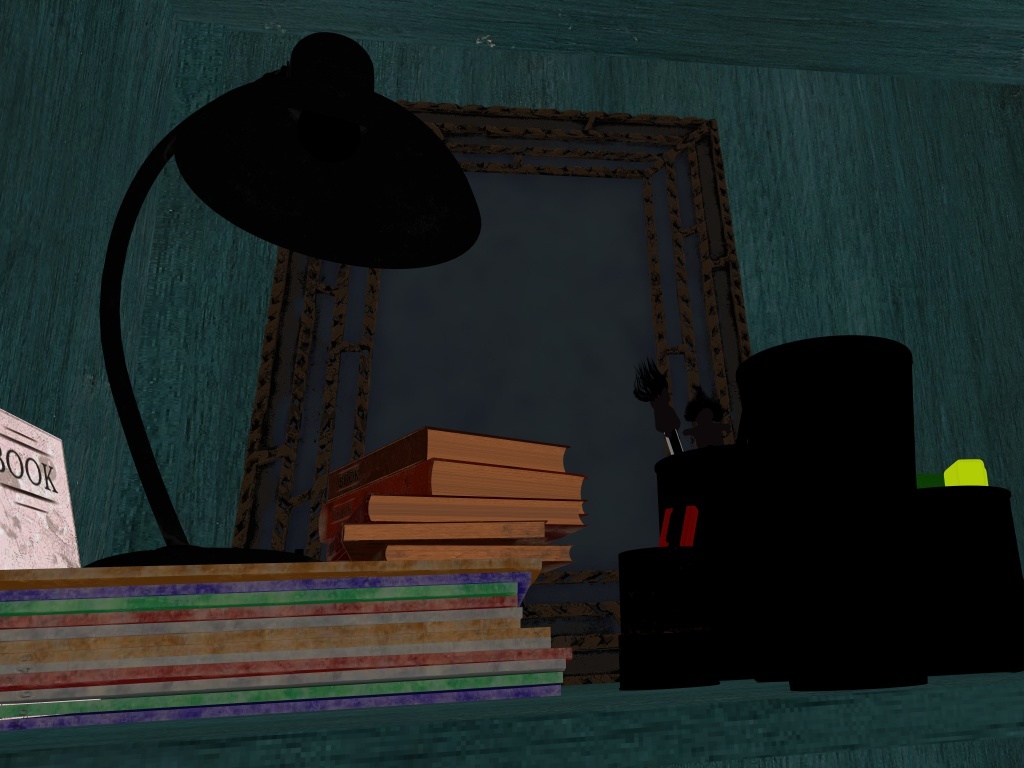} \\
\includegraphics[width=0.15\linewidth]{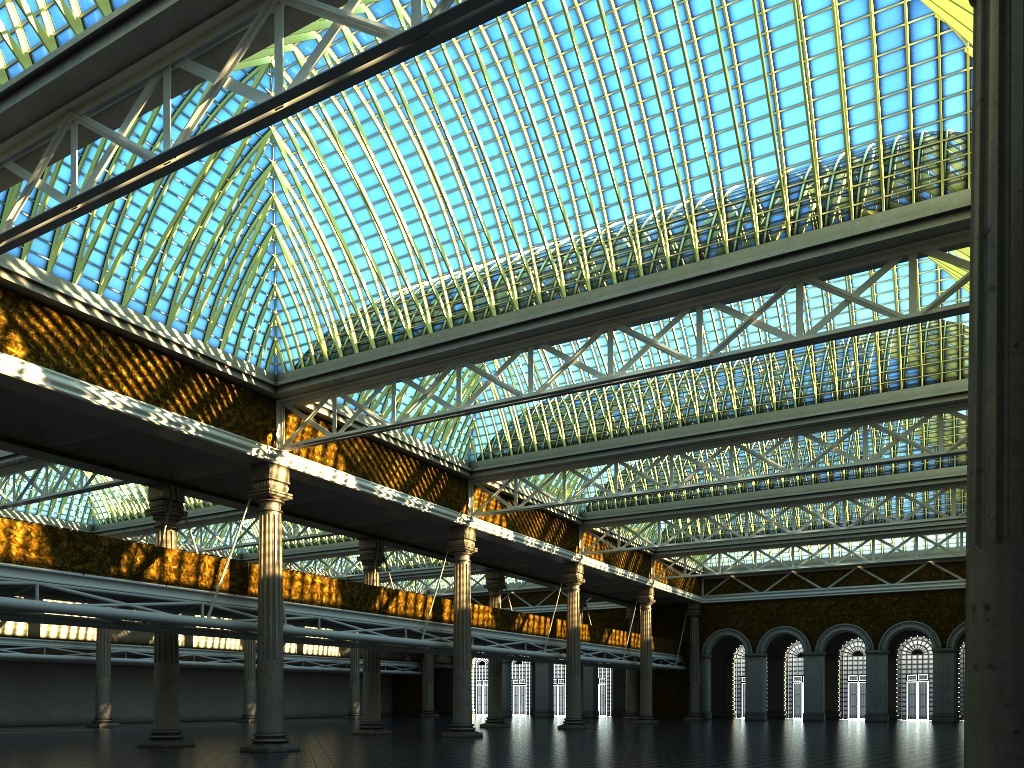} &
\includegraphics[width=0.15\linewidth]{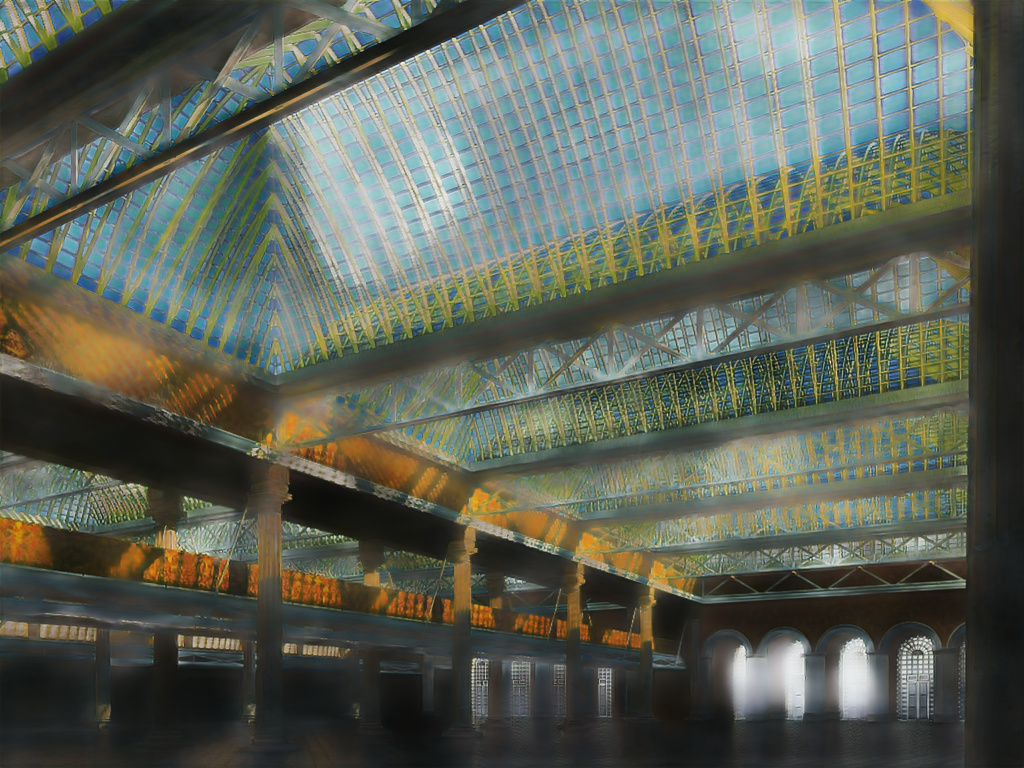} &
\includegraphics[width=0.15\linewidth]{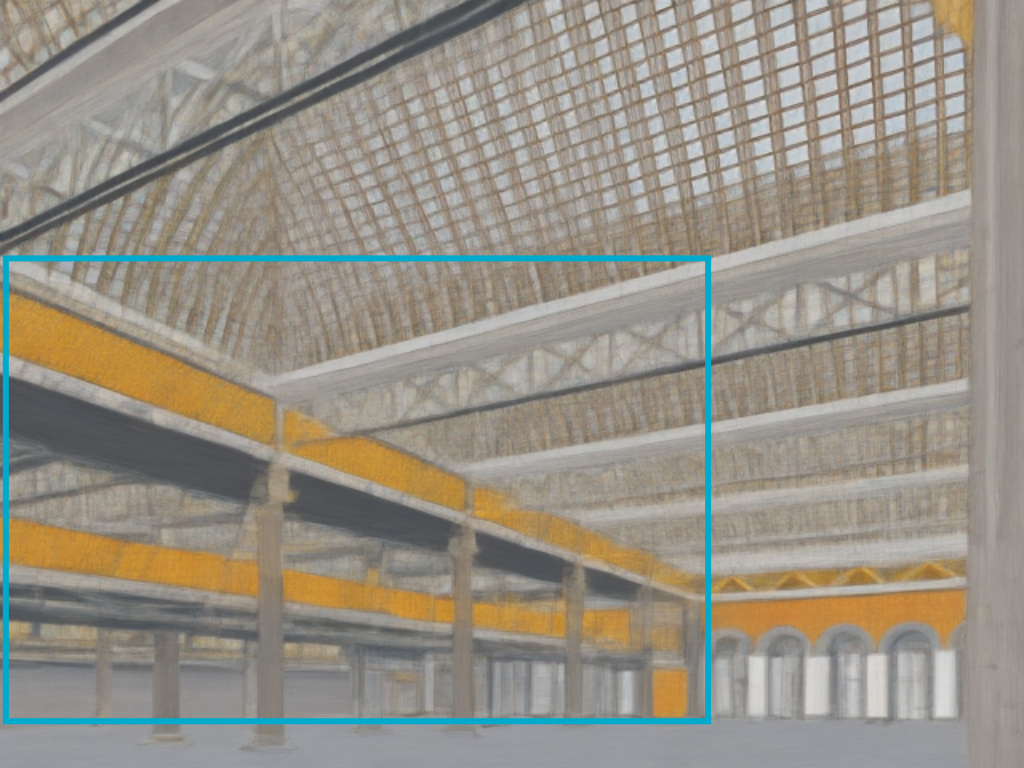} &
\includegraphics[width=0.15\linewidth]{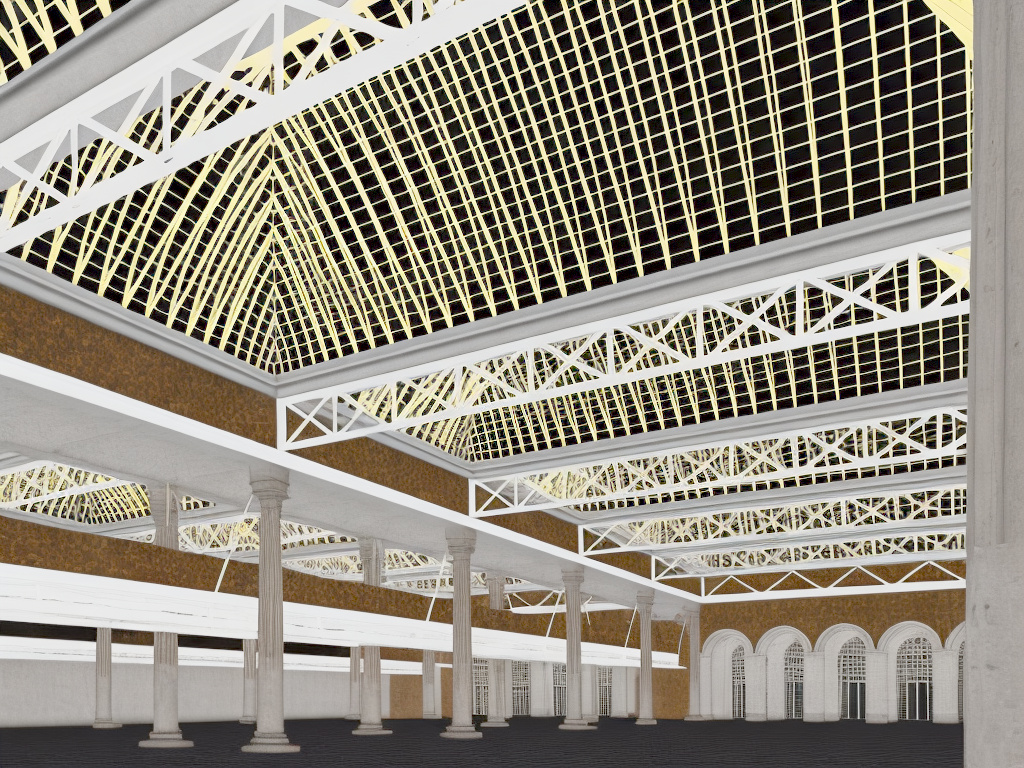} &
\includegraphics[width=0.15\linewidth]{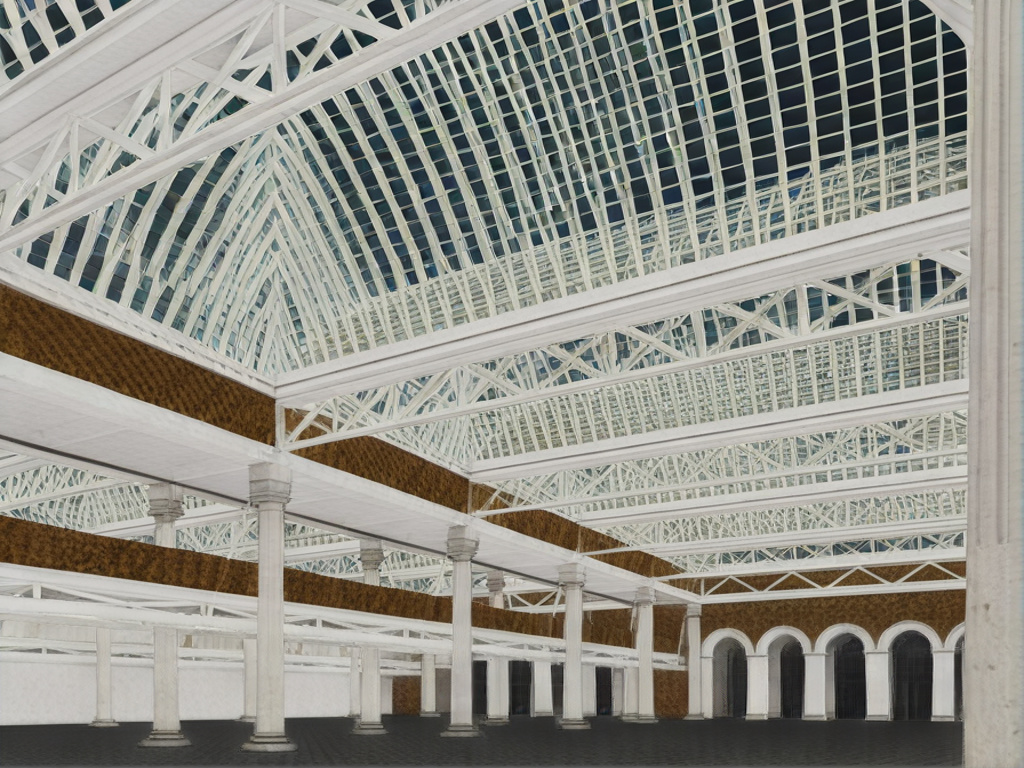} &
\includegraphics[width=0.15\linewidth]{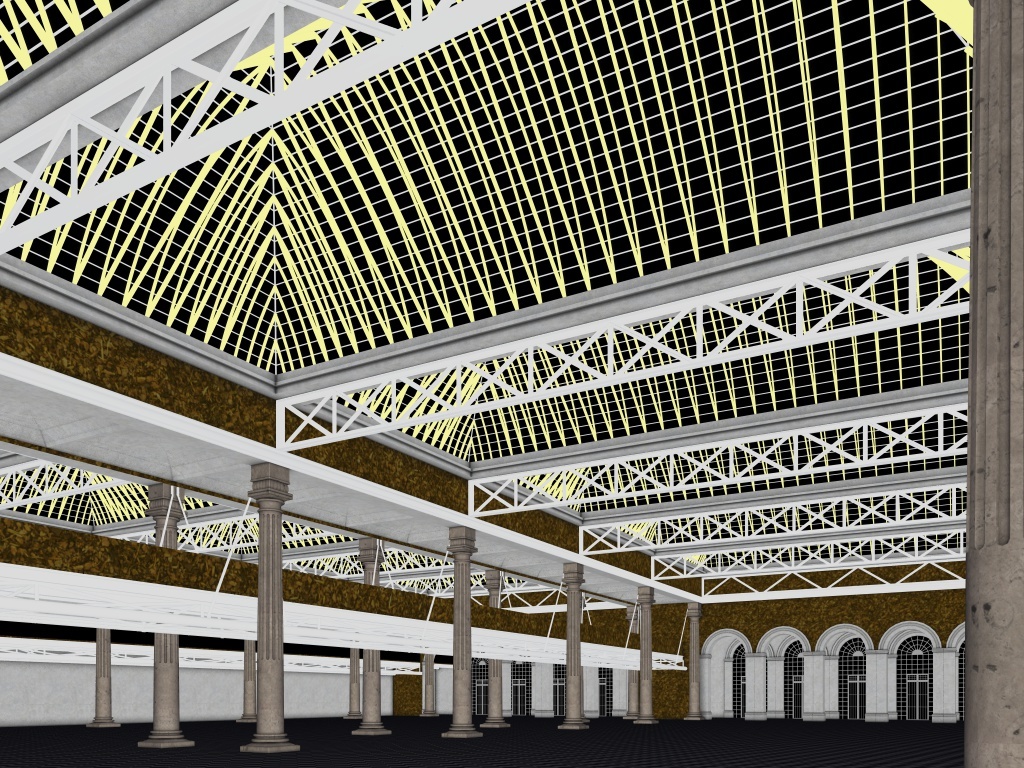} \\
\end{tabular}
\caption{Qualitative comparison on the Hypersim dataset. Cyan rectangles highlight the orange-brown color shift introduced by Kocsis et al. (scene 3), and the featureless estimates produced by RGB$\leftrightarrow$X in low-light regions (scenes 3 and 5).}
\label{fig:sota_hyper}
\end{figure*}

On Hypersim (Figure~\ref{fig:sota_hyper}), our predictions remain consistent across scenes and preserve both surface color and structural detail, while competing methods either blur fine texture or introduce a global color cast; transparent regions remain the main failure case.

\begin{figure*}[tbp]
\centering
\setlength{\tabcolsep}{1pt}
\renewcommand{\arraystretch}{1}
\begin{tabular}{ccccc}
\footnotesize Input RGB & \footnotesize Kocsis et al. & \footnotesize RGB$\leftrightarrow$X & \footnotesize IntrinsicDiffusion & \footnotesize Ours \\
\includegraphics[width=0.19\linewidth]{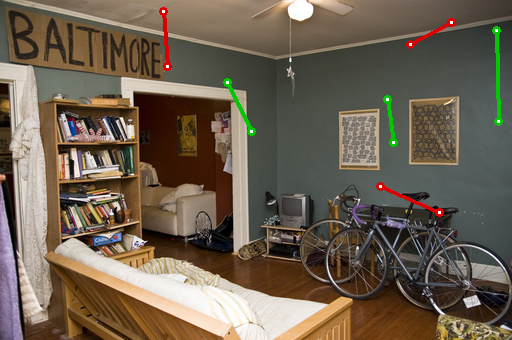} &
\includegraphics[width=0.19\linewidth]{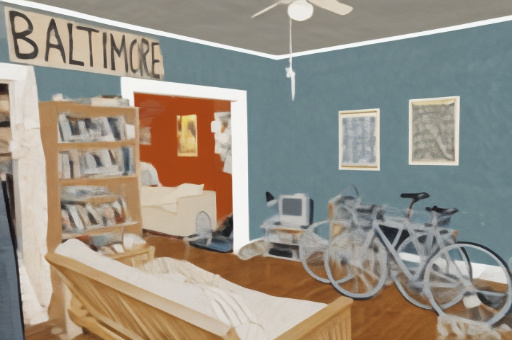} &
\includegraphics[width=0.19\linewidth]{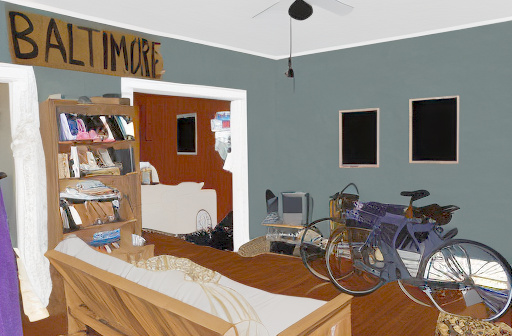} &
\includegraphics[width=0.19\linewidth]{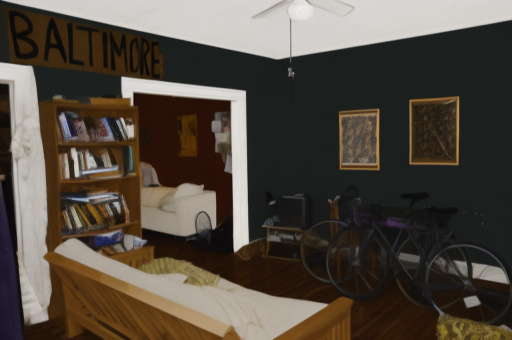} &
\includegraphics[width=0.19\linewidth]{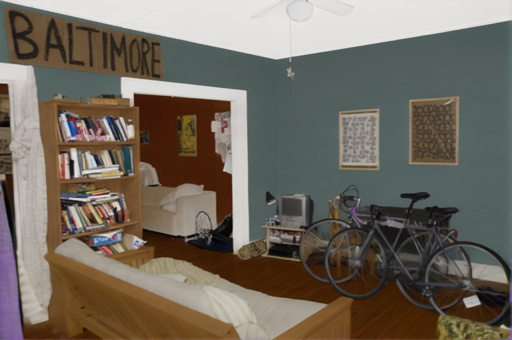} \\
\includegraphics[width=0.19\linewidth]{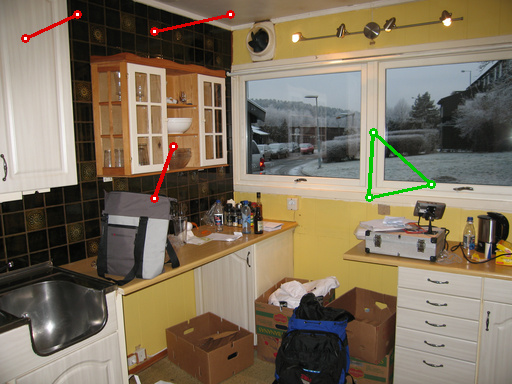} &
\includegraphics[width=0.19\linewidth]{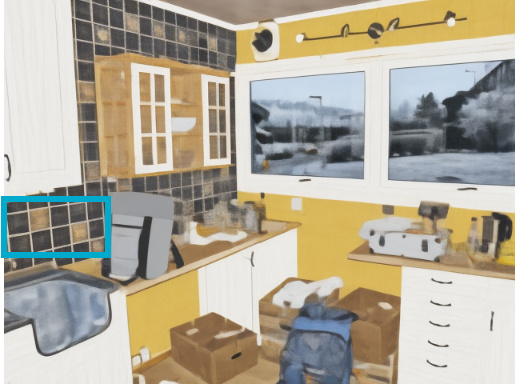} &
\includegraphics[width=0.19\linewidth]{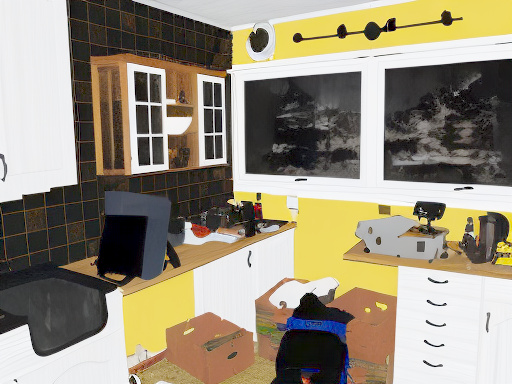} &
\includegraphics[width=0.19\linewidth]{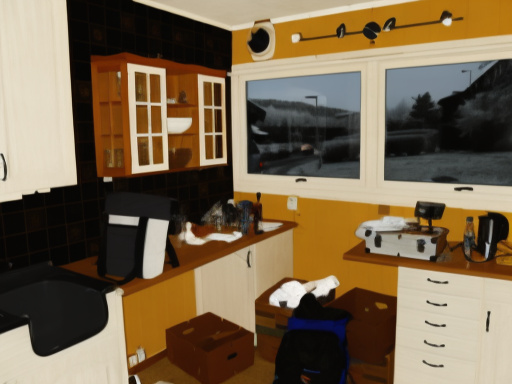} &
\includegraphics[width=0.19\linewidth]{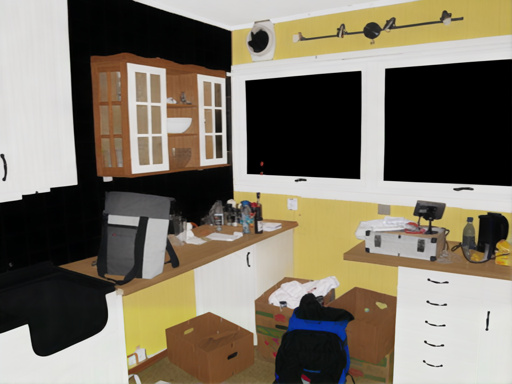} \\
\includegraphics[width=0.19\linewidth]{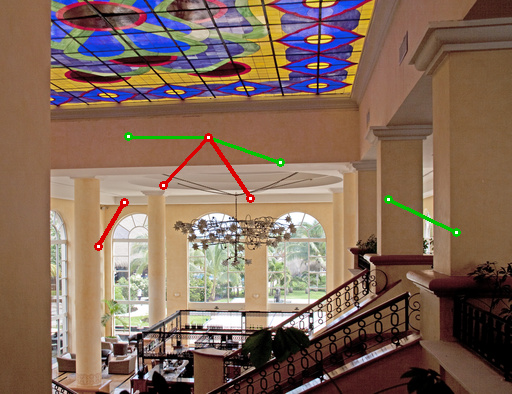} &
\includegraphics[width=0.19\linewidth]{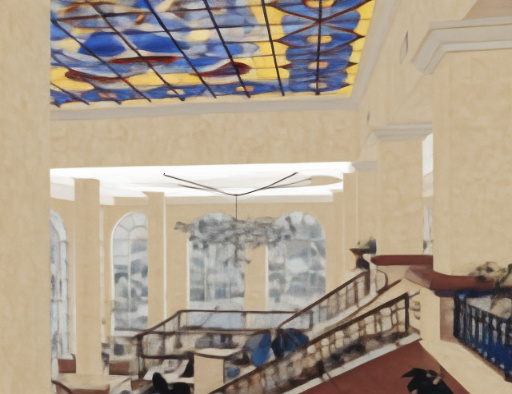} &
\includegraphics[width=0.19\linewidth]{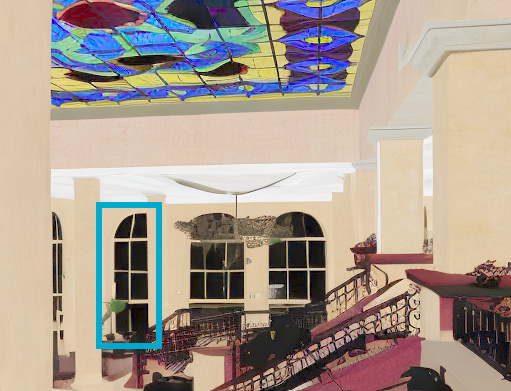} &
\includegraphics[width=0.19\linewidth]{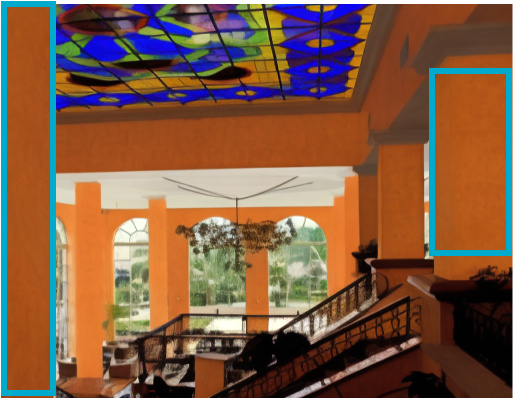} &
\includegraphics[width=0.19\linewidth]{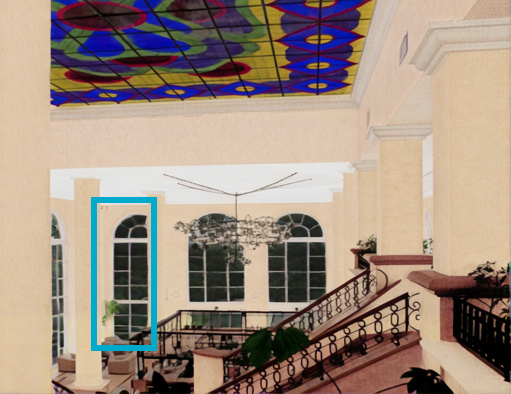} \\
\includegraphics[width=0.19\linewidth]{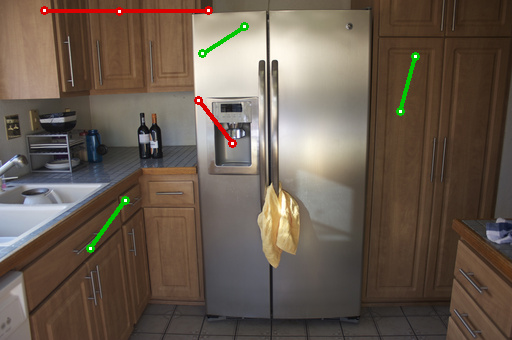} &
\includegraphics[width=0.19\linewidth]{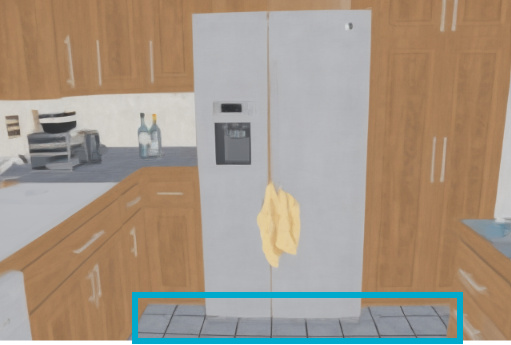} &
\includegraphics[width=0.19\linewidth]{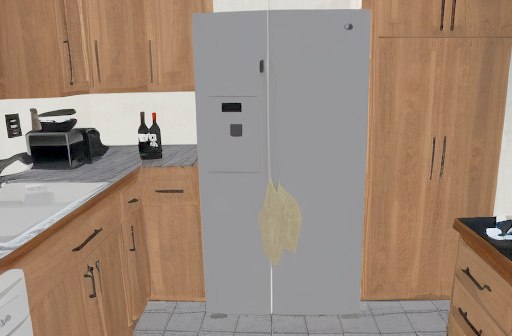} &
\includegraphics[width=0.19\linewidth]{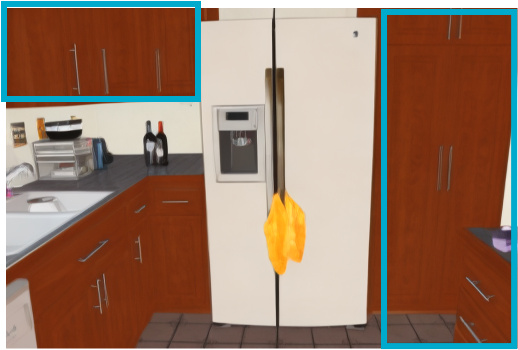} &
\includegraphics[width=0.19\linewidth]{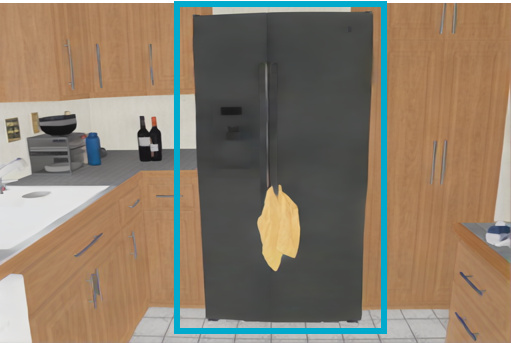} \\
\includegraphics[width=0.19\linewidth]{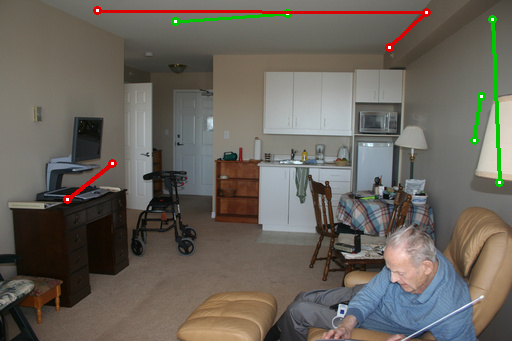} &
\includegraphics[width=0.19\linewidth]{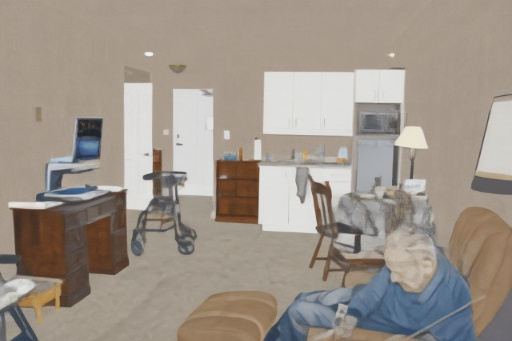} &
\includegraphics[width=0.19\linewidth]{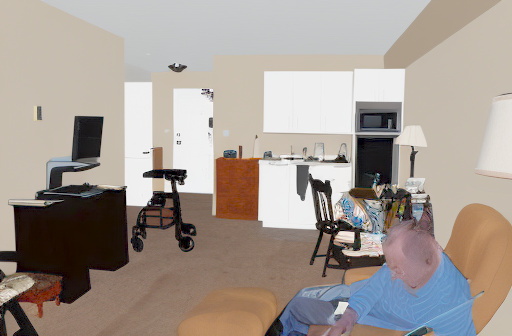} &
\includegraphics[width=0.19\linewidth]{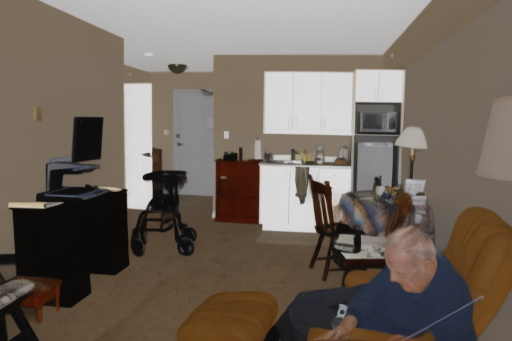} &
\includegraphics[width=0.19\linewidth]{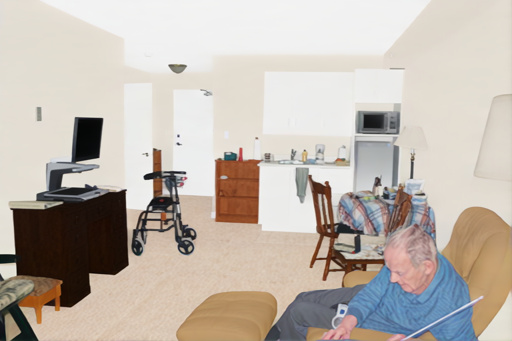} \\
\end{tabular}
\caption{Qualitative results on the IIW dataset. Green and red point pairs in the input image indicate human annotations of equal and different reflectance, respectively. Cyan rectangles highlight the orange/yellowish color shift introduced by IntrinsicDiffusion, the near-black predictions on window regions, and the reflectance collapse on the refrigerator's metallic surface.}
\label{fig:sota_iiw}
\end{figure*}

Finally, on IIW (Figure~\ref{fig:sota_iiw}), our method and RGB$\leftrightarrow$X show the most faithful color rendition on opaque surfaces among the compared methods, while both still struggle with transparent and metallic materials. Because IIW annotations are sparse and purely relative, the WHDR metric does not penalize the color shifts that we and other diffusion-based methods occasionally exhibit, so long as both compared pixels fall inside the shifted region.

Overall, our single-step LBM formulation is competitive with far more expensive multi-step diffusion pipelines and, on some benchmarks, with methods that additionally fine-tune on real-world feedback via reinforcement learning, despite training exclusively on synthetic data with a small number of transport steps. The gap that remains relative to the RL-guided variants is consistent with our central finding: it is not primarily an architectural limitation, but a consequence of relying solely on synthetic supervision, which we discuss further in Section~\ref{sec:limitations}.

\section{Limitations and Future Work}
\label{sec:limitations}
\paragraph{Model capacity and efficiency.} Our approach builds on the Stable Diffusion XL UNet backbone ($\approx$2.5B parameters), which imposes substantial memory requirements and restricts training to $256\times256$ patches; prediction quality degrades beyond roughly 2K resolution at inference. Preliminary attempts to fine-tune with LoRA~\cite{hu2022lora} to reduce this cost did not converge satisfactorily. Diffusion Transformer backbones, as used in Stable Diffusion 3, are a promising direction for improving scalability without this limitation.
\vspace{-5mm}
\paragraph{Dependence on synthetic training data.} Both training datasets introduce a domain gap that constrains real-world generalization. In Hypersim, near-zero albedo values are assigned indiscriminately to dark opaque objects and non-diffuse materials, creating supervision ambiguity, and a small subset of corrupted, textureless scenes injects further noise during training. InteriorVerse exhibits a bias toward specific indoor color-material associations, which likely explains the color shifts observed in our qualitative results (Section~\ref{sec:sota}). More broadly, this reflects the scarcity of large-scale, real-world datasets with dense intrinsic ground truth discussed in Section~\ref{sec:related}: an ideal dataset would combine dense annotations with broad scene diversity (indoor, outdoor, objects, people) under multiple illuminations and non-Lambertian effects such as interreflections and caustics.
\vspace{-5mm}
\paragraph{Evaluation inconsistencies across the literature.} As Tables~\ref{tab:sota_mit}--\ref{tab:sota_combined} illustrate, IID methods are evaluated with heterogeneous, dataset-specific metrics (WHDR for IIW; PSNR/LPIPS for InteriorVerse and Hypersim; DSSIM/LMSE/MSE for MIT and ARAP), and few methods report all five benchmarks. This fragmentation makes it difficult to identify a single best-performing method and motivates standardized, multi-dataset evaluation protocols for the field.
\vspace{-5mm}
\paragraph{Future directions.} Beyond addressing the limitations above, a particularly promising direction is reducing reliance on dense ground-truth annotations through alternative supervision signals; ReasonX~\cite{dirik2026reasonx} shows that reinforcement learning guided by perceptual feedback from multimodal large language models can substantially improve decomposition on real-world images, and being model-agnostic, it is a natural extension for our LBM-based pipeline. Extending the conditioning study in Section~\ref{par:recons_loss} to other residual, non-Lambertian effects, such as specular reflections and interreflections, is a further direction suggested by our mutual albedo-shading conditioning results.

\section{Conclusions}
\label{sec:results}

In this work, our results support the suitability of the LBM framework to deal with low-level image processing tasks compared to previous generative approaches. In particular, we focused on the IID problem that is highly ill-posed and requires solutions with physical coherence. We propose a novel model for albedo estimation that satisfies the image formation model while leveraging the flexibility of LBM to incorporate conditioning inputs, and we show that albedo and shading act as mutually informative conditioning signals for one another, whereas surface normals do not, despite their geometric expressiveness. Benchmarked against state-of-the-art IID methods on five datasets (Section~\ref{sec:sota}), our single-step model is competitive with substantially more expensive multi-step diffusion pipelines, though a gap remains relative to methods that additionally fine-tune on real-world feedback.

Finally, we want to note that the IID field appears to have reached a bottleneck, where progress has seemingly stalled. It is difficult to identify a clear state-of-the-art method, as no existing approach consistently outperforms others across all datasets, and evaluation practices remain fragmented across dataset-specific metrics (Section~\ref{sec:limitations}). This limitation is likely due to insufficient generalization capabilities, which stem from the strong dependence on large-scale synthetic datasets for training. Recently, Reason-X~\cite{dirik2026reasonx} has acknowledged and addressed this bottleneck by introducing a novel perspective based on reinforcement learning, aimed at improving the ability of trained models to generalize to real-world images. As a natural extension of our work, a key direction for future research will be to explore this approach to further enhance the performance and generalization of the proposed method.

\section*{Acknowledgments}

This work was funded by Grant PID2024-162555OB-I00; by MICIU/AEI/10.13039/501100011033, ERDF/EU and the FEDER; by the grant Càtedra ENIA UAB-Cruïlla (TSI-100929-2023-2) from the Ministry of Economic Affairs and Digital Transition of Spain. DSL also acknowledges the FPI grant from Spanish Ministry of Science and Innovation (PRE2022-101525). JVC also acknowledges the 2025 Leonardo Grant for Scientific Research and Cultural Creation from the BBVA Foundation. The BBVA Foundation accepts no responsibility for the opinions, statements and contents included in the project and/or the results thereof, which are entirely the responsibility of the authors.

\bibliographystyle{unsrt}
\bibliography{bib2CIC}

⁄\end{document}